\documentclass{clv3}

\usepackage[utf8]{inputenc}
\usepackage{amsmath}
\usepackage{tikz-dependency}
\usepackage{booktabs, multirow} 
\usepackage{soul}

\usepackage{subcaption}
\usepackage{url}
\usepackage{multirow}

\usepackage{tikz}
\usepackage{todonotes}
\usepackage{times}
\usepackage{pifont}
\usepackage{hyperref}
\usepackage{cleveref,multirow}
\usepackage{xcolor,colortbl}
\usepackage{pdflscape}

\newcommand{\thirdrev}[2]{#2}
\newcommand{\secrev}[2]{#2}

\definecolor{darkblue}{rgb}{0, 0, 0.5}

\definecolor{green}{rgb}{0.1,0.1,0.1}

\definecolor{darkgreen}{rgb}{0.4, 0.69, 0.2}
\definecolor{lightgreen}{rgb}{0.56, 0.93, 0.56}

\definecolor{darkred}{rgb}{1.0, 0.25, 0.25}
\definecolor{lightred}{rgb}{1.0, 0.72, 0.77}

\hypersetup{colorlinks=true,citecolor=darkblue, linkcolor=darkblue, urlcolor=darkblue}

\historydates{Submission received:            15 January 2021,
Revised version received:        19 August 2021,
Accepted for publication:        7 October 2021}

\dochead{}

\runningtitle{To Augment or Not to Augment?}

\runningauthor{Gözde Gül Şahin}

\begin{document}

\title{To Augment or Not to Augment? A Comparative Study on Text Augmentation Techniques for Low-Resource NLP}

\author{Gözde Gül Şahin\thanks{UKP Lab, Department of Computer Science, Technical University of Darmstadt, Darmstadt, Germany. E-mail: sahin@ukp.informatik.tu-darmstadt.de.}}
\affil{UKP Lab / TU Darmstadt}

\maketitle

\begin{abstract}

Data-hungry deep neural networks have established themselves as the defacto standard for many NLP tasks including the traditional sequence tagging ones. Despite their state-of-the-art performance on high-resource languages, they still fall behind of their statistical counter-parts in low-resource scenarios. One methodology to counter attack this problem is \textit{text augmentation}, i.e., generating new synthetic training data points from existing data. Although NLP has recently witnessed a load of textual augmentation techniques, the field still lacks a systematic performance analysis on a diverse set of languages and sequence tagging tasks. To fill this gap, we investigate three \secrev{existing}{categories of text} augmentation methodologies which perform changes on the syntax (e.g., cropping sub-sentences), token (e.g., random word insertion) and character (e.g., character swapping) levels. We systematically compare the methods on part-of-speech tagging, dependency parsing and semantic role labeling for a diverse set of language families\secrev{such as Turkic, Slavic, Mongolic and Dravidian}{} \secrev{}{using various models including the architectures that rely on pretrained multilingual contextualized language models such as \texttt{mBERT}}. \secrev{}{Augmentation most significantly improves dependency parsing, followed by part-of-speech tagging and semantic role labeling. We find the experimented techniques to be effective on morphologically rich languages in general rather than analytic languages such as Vietnamese. Our results suggest that the augmentation techniques can further improve over strong baselines based on \texttt{mBERT}, especially for dependency parsing. We identify the character-level methods as the most consistent performers, while synonym replacement and syntactic augmenters provide inconsistent improvements. Finally, we discuss that the results most heavily depend on the task, language pair \textit{(e.g., syntactic-level techniques mostly benefit higher-level tasks and morphologically richer languages)}, and the model type \textit{(e.g., token-level augmentation provide significant improvements for \texttt{BPE}, while character-level ones give generally higher scores for \texttt{char} and \texttt{mBERT} based models).}}
   
\end{abstract}

\section{Introduction}


  Recent advancements in natural language processing (NLP) field have led to models that surpass all previous results on a range of high-level downstream applications such as machine translation, text classification, dependency parsing and many more. However these models require huge number of training data points to achieve state-of-the-art scores and are known to suffer from out-of-domain problem. In other words, they are not able to correctly label or generate novel, unseen data points. In order to boost the performance of such systems in presence of low-data, the researchers have introduced various \textit{data augmentation} techniques that aim to increase the sample size and also the variation of the lexical~\cite{wei-zou-2019-eda,FadaeeBM17a,Kobayashi18,KarpukhinLEG19} or syntactic patterns~\cite{VickreyK08,SahinS18,GulordavaBGLB18}. A similar line of research introduced adversarial attack and defense mechanisms~\cite{BelinkovB18,KarpukhinLEG19} based on injecting noises with the goal of more robust NLP systems.  

  Even though low-resource languages are the perfect test bed for such augmentation techniques, a large amount of studies only \textit{simulate} a low-resource environment by sampling from a high-resource language like English~\cite{wei-zou-2019-eda,DBLP:journals/corr/abs-1905-08941}. Unfortunately, methods that perform well on English, may function poorly for many low-resource languages. This is due to English being an analytic language while most low-resource languages are synthetic. 
  Furthermore, the majority of the studies focus either on sentence classification or machine translation. Although these tasks are important, the traditional sequence tagging tasks where tokens (e.g., white-\texttt{Adj} cat-\texttt{Noun}) or the relation between tokens (e.g., white \texttt{Modifier} cat) are labeled, are still considered as primary steps to natural language understanding. These tasks generally have finer grained labels and are more sensitive to noise. In addition, previous works mostly report single best scores that can be achieved by the augmentation techniques. 


  To extend the knowledge of the NLP community on text augmentation methods, there is a need for a comprehensive study that investigates (i) different types of augmentation methods (ii) on a diverse set of low-resource languages, for (iii) a diverse set of sequence tagging tasks that require different linguistic skills \textit{(e.g., morphologic, syntactic, semantic)}. In order to address these issues, we explore a wide range of text augmentation methodologies that augments on character-level~\cite{KarpukhinLEG19}, token-level~\cite{KolomiyetsBM11,ZhangL15,wang-yang-2015-thats,wei-zou-2019-eda} and syntactic-level~\cite{SahinS18,GulordavaBGLB18}. 
  To gain insights on the capability of these methods, we experiment on sequence tagging tasks of varying difficulty: POS tagging and dependency parsing and semantic role labeling, a.k.a., shallow semantic parsing. POS tagging and dependency parsing experiments are performed on truly low-resource languages that are members of various language families: Kazakh (Turkic), Tamil (Dravidian), Buryat (Mongolic), Telugu (Dravidian), Vietnamese (Austro-Asiatic), Kurmanji (Iranian) and Belarusian (Slavic). Due to lack of annotated data, we simulate the low-resource environment for semantic role labeling (SRL) on languages from a diverse set of families, namely as Turkish (Turkic), Finnish (Uralic), Catalan (Romance), Spanish (Romance) and Czech (Slavic). 
  \secrev{}{To investigate whether the augmentations are model-agnostic and can further improve on state-of-the-art models, we experiment with distinct models that use various subword units \textit{(e.g., character, byte-pair-encoding and word piece)}, pretrained embeddings \textit{(e.g., BPE-GloVe, multilingual BERT)} and architectural designs \textit{(e.g., biaffine, transition-based parser)}. We train and test each model multiple times and report the mean, standard deviation scores along with the p-values calculated via paired t-tests.} Furthermore, we \secrev{conduct experiments with different augmentation parameters to analyze the performance range of these techniques along with}{investigate the sensitivity of augmentation techniques to parameters in a separate study, and analyze the contribution of each technique to the improvement of frequent and rare tokens and token classes}. 

  \secrev{We show that less sophisticated augmentation techniques such as injecting character-level noise, deleting random words or generating grammatical, meaningless sentences are more statistically likely to improve the performances of POS tagging and dependency parsing, especially in truly low-resource scenarios. This is, however, dependent on the language. For instance, gains are achieved for Belarusian, Tamil, Kazakh and Kurmanji POS taggers, while no consistent improvements have been found for Vietnamese, Hungarian and Telugu. We show similar augmentation patterns for dependency parsing, such as injection of character-level noise dominantly providing the best scores. One of the key differences, however, is that the syntactic augmentation almost always improves dependency parsing unlike POS tagging. Our experiments show that the success of augmentation techniques differ with the language which is especially emphasized in the semantic role labeling task. For instance, morphologically richer languages Turkish, Czech and Finnish benefit more from sophisticated augmentation techniques, while synthetic noise, especially character swapping, advance Spanish and Catalan SRL systems the most.}{Our results show that augmentation methods benefit the dependency parsing task more significantly than part-of-speech tagging and semantic role labeling---in the given order---independent from the parser type. The improvements are more reliably observed in morphologically richer languages, i.e., the results for Vietnamese (an analytic language) are varied---in some cases significantly worse than the baselines. We find that augmentation can still provide performance gains over the strong baselines built on top of pretrained contextualized language models---especially in dependency parsing. In general, character-level augmentation gives more consistent improvements for the experimented languages and tasks, whereas synonym replacement and rotating sentences mostly result with a weak increase or decrease. We find that the subword unit choice and task requirements also have a large impact on the results. For instance, we observe significant gains over the character based POS tagger, but no improvement for the BPE based one for Tamil language. In addition, we find that token-level augmentation is more effective for BPE based models, whereas character-level augmentation techniques provide significant gains for the rest. Furthermore we observe that syntactic augmentation techniques are more likely to improve the performance on morphologically richer languages as pronounced in the SRL task.}

\section{Related Work}
\label{sec:relwork}
  Augmentation techniques for NLP tasks fall into two categories: feature space augmentation (FSA) and text augmentation (TA). FSA techniques mostly focus on augmenting the continuous representation space directly inside the model, while TA processes the discrete variables such as raw or annotated text. 

  \paragraph{FSA} \citet{GuoMZ19} and \citet{Guo20} propose to generate a synthetic sample in the feature space via linearly and nonlinearly interpolating the original training samples. Although the idea originates from computer vision field~\cite{ZhangCDL18}, it has been adapted to English sentence classification tasks successfully~\cite{DBLP:journals/corr/abs-1905-08941}. More recently \citet{GuoKR20} proposed a similar technique that mixes up the input and the output sequences and showed improvements on several sequence-to-sequence tasks such as machine translation between high-resource language pairs. \secrev{}{Despite their success in text classification and sequence-to-sequence tasks, they are seldom used for sequence tagging tasks. \citet{zhang-etal-2020-seqmix} use the mixup technique~\cite{ZhangCDL18} in scope of active learning, where they augment the queries at each iteration and later classify whether the augmented query is plausible---since the resulting queries might be noisy---and report improvements for the Named Entity Recognition~(NER) and event detection task. However building a robust discriminator for more challenging tasks as dependency parsing (DP) and semantic role labeling (SRL) is a challenge on its own. \citet{ChenWTYY20} report that direct application of the mixup techniques~\cite{GuoMZ19,Guo20} for NER introduce vast amount of noise, therefore makes the learning process even more challenging. To address this, \citet{ChenWTYY20} introduce local interpolation strategies that mixes sequences \textit{close} to each other, where \textit{closeness} for tokens is defined either as (i) occurring in the same sentence (ii) occurring in the sentence within the $k_{th}$ neighborhood; and show improvements over state-of-the-art NER models. Even though NER is a sequence tagging task, it is fundamentally different from the tasks in this study: dependency parsing (DP) and semantic role labeling (SRL). The local context for the NER task is usually the full sentence, however SRL the local context is much smaller. Imagine these sentences: ``I want to visit the \textit{USA}'' and ``\textit{USA} is trying to solve the problems''. NER would label \textit{USA} as \textsc{Country} in both sentences, whereas \textit{USA} would have the labels \textsc{Arg1: visited} and \textsc{Arg0: agent/entity trying}. As can be seen, for our sequence tagging tasks, the local context is much smaller and the labels are more fine-grained than in NER. Therefore, \cite{ChenWTYY20}-(i) would have an extremely small sample space, whereas \cite{ChenWTYY20}-(ii) is likely to produce too much noise. To sum up, employing these techniques for more complicated sequence tagging tasks such as DP and SRL is not straight forward. This is due to (i) \textit{the relations between tokens} being labeled instead of the tokens themselves; and (ii) the labels being extremely fine-grained compared to classification tasks. Furthermore, FSA techniques require direct access to the neural architecture since they modify the embedding space. That means they cannot be used in combination with black-box systems which might be a drawback for NLP practitioners. Finally, unlike the aforementioned sequence labeling tasks, DP and SRL have relatively complex input layer where additional linguistic information (e.g., postags, predicate flag) is used. That also makes the direct application of FSA techniques more challenging. For these reasons, we focus on the other category of augmentation methods.}  


  \paragraph{TA} 
  \citet{VickreyK08} introduce a number of expert-designed sentence simplification rules to augment the dataset with simplified sentences, and showed improvements on English semantic role labeling. Similarly, \citet{SahinS18} propose an automated approach to generated simplified and reordered sentences using dependency trees. We refer to such methods as \textit{label-preserving} since they do not alter the semantics of the original sentence. They mostly perform on syntactically annotated data and sometimes require manually designed rules~\cite{VickreyK08}. 

  Other set of techniques~\cite{GulordavaBGLB18,FadaeeBM17a} perform lexical changes instead of syntactic restructuring. \citet{GulordavaBGLB18} generate synthetic sentences by replacing a randomly chosen token with its \textit{syntactic} equivalent. \citet{FadaeeBM17a} replaces more frequent words with the rare ones to create stronger lexical-label associations. \citet{wei-zou-2019-eda,ZhangL15} make use of semantic lexicons like WordNet to replace words with their synonyms. \citet{Kobayashi18,WuLZHH19,FadaeeBM17a} employ pretrained language models to generate a set of candidates for a token position, while \citet{anaby2020not,kumar2020,ding-etal-2020-daga} take a generative approach. Most work focus on classification and translation except from \citet{ding-etal-2020-daga} that experiment on low-resource tagging tasks. However it assumes one label per token, therefore is not directly applicable to relational tagging tasks such as dependency parsing and semantic role labeling. Additionally, \citet{wei-zou-2019-eda} propose simple techniques such as adding, removing, or replacing random words and show that such perturbations improve the performances of English sentence classification tasks. Unlike other lexical augmentation methods, these do not require syntactic annotations, semantic lexicons and large language models that make them more suitable to low-resource settings. 

  There also exists sentence-level techniques such as back-translation~\cite{sennrich-etal-2016-improving} that aim to generate a paraphrase by translating back and forth between a pair of languages. 
  However such techniques are mostly not applicable to sequential tagging tasks like semantic parsing while they cannot guarantee the token-label association. Furthermore training an intermediate translation model for paraphrasing purposes may not be possible for genuinely low-resource languages. \thirdrev{}{On a similar line \citet{abs-2104-08826} leverage a pretrained large language model to generate text samples and report improvements on English text classification tasks.}

  Another category of TA uses \textit{noising} techniques that are closely associated to adversarial attacks to text systems. \citet{BelinkovB18} attack machine translation systems by modifying the input with synthetic (e.g., swapping characters) and natural noise (e.g., injecting common spelling mistakes). Similarly \citet{KarpukhinLEG19} defend the machine translation system by training on a set of character-level synthetic noise models. \secrev{}{\citet{HanZJT20} introduce an adversarial attack strategy for structured prediction tasks including dependency parsing. They first train a seq2seq generator via reinforcement learning where they design a special reward function that evaluates whether the generated sequence could trigger a wrong output. The evaluation is carried out by two other reference parsers---i.e., if both parsers are \textit{tricked} then it is likely that the victim parser would also be misled. \citet{HanZJT20} then use adversarial training as defense mechanism and show improvements. On the same line of research, \citet{ZhengZZHCH20} propose replacing a token with a ``similar'' token, where similarity is defined as having a similar log likelihood of being generated by pretrained BERT~\cite{DevlinCLT19} given the context and having the same POS tag (in the black-box setting). The tokens are chosen in a way that the parser's error rate is maximized. The attack is called a ``sentence-level attack'' when the chosen tokens' positions are irrelevant. On a ``phrase-level attack'', the authors first choose two subtrees and then maximize the error rate on the target subtree by modifying the tokens in the source subtree. Even though the adversarial example generation techniques~\cite{ZhengZZHCH20,HanZJT20} could be used to augment data in theory, the requirements such as a separate seq2seq generator, a BERT based scorer~\cite{ZhangKWWA20}, reference parsers that are of certain quality, external POS taggers and high quality pretrained BERT~\cite{DevlinCLT19} models, make them challenging to apply on low-resource languages. Besides, most of the aforementioned adversarial attacks are optimized to trigger an undesired change in the output with minimal modifications, while data augmentation is only concerned about increasing the generalization capacity of the model.}

\paragraph{Surveys}  

\thirdrev{}{Due to the growing number of data augmentation techniques and interest in them, a couple of survey studies have been recently published~\cite{HedderichLASK21,FengGWCVMH21,ChenJE20}. \citet{HedderichLASK21} give a comprehensive overview of recent methods that are proposed to tackle low-resource scenarios. The authors overview a range of techniques both for low-resource languages and domains, discussing their limitations, requirements and outcomes. Apart from data augmentation methods, the authors discuss more general approaches such as cross-lingual projection, transfer learning, pretraining (of large multilingual models) and meta-learning, which are not in scope of this work. \citet{FengGWCVMH21} provide a more focused survey, zooming in data augmentation techniques rather than other common approaches such as transfer learning. They categorize data augmentations into three categories as (i) rule-based, (ii) example interpolation and (iii) model-based techniques. \textit{Rule-based techniques} are referred to as easy-to-compute methods such as EDA~\cite{wei-zou-2019-eda} and dependency tree morphing~\cite{SahinS18} which are also covered in this study. 
\textit{Example interpolation} techniques are used to define feature-space augmentation (FSA) type of methods that we have discussed earlier in this section. \textit{Model-based techniques} refer to augmentation techniques that rely on large models trained on large texts (e.g., backtranslation and synonym replacement using BERT like models) which either don't preserve the labels or require a strong pretrained model---which is mostly not available for low-resource languages. Both \citet{FengGWCVMH21} and \citet{HedderichLASK21} provide a taxonomy and a structured bird's eye view on the topic without any empirical investigation as in this study. The closest work to ours is by \citet{vaniaKSL19}, that explore two of the augmentation techniques along with other approaches~(e.g., transfer learning) for low-resource dependency parsing, which is limited in numbers of tasks, languages and augmentation techniques. Finally, \citet{ChenJE20} provide an empirical survey covering a wide range of NLP tasks and augmentation approaches where most of them exist for English only. Hence, the experiments are performed \textit{only on English}, providing no insights for low-resource languages or sequence tagging tasks. Unlike previous literature, our work (i) focuses on sequential tagging tasks that require various linguistic skills: part-of-speech tagging, dependency parsing and semantic role labeling; (ii) experiments on multiple low-resource languages from a diverse set of language families; (iii) compares and analyzes a rich compilation of augmentation techniques that are suitable for low-resource languages and the focused tasks. 
} 


\section{Augmentation Techniques}
\label{sec:augmain}
\thirdrev{}{
As discussed in Section~\ref{sec:relwork}, we categorize the augmentation techniques as \textit{textual} (TA) and \textit{feature-space} augmentation (FSA). In this paper, we focus on \textbf{textual augmentation} techniques rather than FSA for several reasons. First of all, existing FSA techniques have only been implemented and tested for simple sequence tagging tasks where the task is to choose the best tag for the token from a quite limited number of labels. However in our sequence tagging tasks, the labels are quite \textbf{fine-grained}, and the \textbf{relation between tokens} are labeled rather than the tokens themselves. Second, FSA techniques require direct access to the neural architecture since they modify the embedding space, which means they cannot be used in combination with black-box systems. Finally, for some of the sequence tagging tasks, namely as dependency parsing and semantic role labeling, the models might have relatively complex input layers (e.g., incorporate additional features such as postags, predicate flag), which makes the direct application of FSA techniques more challenging.
}

\thirdrev{}{
Textual augmentation techniques, i.e., augmentation techniques that alter the input text, can be applied on many different levels such as character, token, sentence or document. Since we focus on sentence-level tagging tasks, document level augmentation techniques are simply ignored in this study. Furthermore, the focus of this paper is to investigate techniques that are \textit{suitable for low-resource languages} and are able to \textit{preserve the task labels} up to some degree. For instance sentence level augmentation techniques like backtranslation are not suitable since they would not preserve the token labels. Similarly, genuinely low-resource languages do not have associated strong pretrained language models due to lack of raw data. Therefore sophisticated techniques that make use of such models are also left out in this paper. 
}

\setlength{\tabcolsep}{2.4pt}
\begin{table*}[!ht]
    \centering
    \scalebox{.8}{
    \footnotesize
    
    \begin{tabular}{l|ccc}
    \toprule
    \textbf{\textsc{DA} Method} & Level & Task & Reason \\ 
    
    \midrule  
    \textsc{Synonym Replacement} \cite{wang-yang-2015-thats} &  Input  & Agnostic & -  \\
    \textsc{Random Deletion} \cite{wei-zou-2019-eda} & Input  & Agnostic & -  \\
    \textsc{Random Swap} \cite{wei-zou-2019-eda} & Input  & Agnostic & -  \\
    \textsc{DTreeMorph}  \cite{SahinS18} & Input  & Agnostic & - \\
    \textsc{Synthetic Noise}  \cite{KarpukhinLEG19} & Input  & Agnostic & - \\
    \textsc{Nonce} \cite{GulordavaBGLB18} & Input & Agnostic & - \\ 
    
    \midrule

    
    \textsc{Backtranslation} \cite{sennrich-etal-2016-improving} & Input & Agnostic & labels not preserved  \\
    \textsc{UBT \& TBT} \cite{vaibhav-etal-2019-improving} & Input & Agnostic & labels not preserved \\
    \textsc{Data Diversification} \cite{nguyen2020data} & Input & Agnostic & labels not preserved \\
    \textsc{SCPN} \cite{wieting-gimpel-2017-revisiting} &  Input  & Agnostic  & labels not preserved  \\
    \textsc{Semantic Text Exchange} \cite{feng-etal-2019-keep} & Input & Agnostic & labels not preserved \\
    \textsc{XLDA} \cite{singh2019xlda} & Input & Agnostic & labels not preserved  \\
    \textsc{LAMBADA} \cite{anaby2020not} & Input & classification & labels not preserved \\    
    
    \midrule
    \textsc{ContextualAug} \cite{Kobayashi18} & Input  & Agnostic & requires strong pretrained model \\
    \textsc{Soft Contextual DA} \cite{gao-etal-2019-soft} & Emb/Hidden & Agnostic & requires strong pretrained model \\
    \textsc{Slot-Sub-LM} \cite{louvan2020simple} &  Input & slot filling & requires strong pretrained model \\
    %
    %
    \midrule
    \textsc{WN-Hypers}  \cite{feng2020genaug} & Input  & Agnostic & requires WordNet \\
    \textsc{UEdin-MS} (DA part) \cite{grundkiewicz-etal-2019-neural} & Input & Agnostic & requires spell checker \\
    
    \midrule
    \textsc{SeqMixUp}  \cite{GuoKR20} & Input & seq2seq  & not suitable \\   
    \textsc{Emix}  \cite{jindal-etal-2020-augmenting} & Emb/Hidden  & classification & not suitable \\
    \textsc{SpeechMix}  \cite{jindalspeechmix} & Emb/Hidden  & Speech/Audio & not suitable \\
    \textsc{MixText}  \cite{chen2020mixtext} & Emb/Hidden  & classification & not suitable \\
    \textsc{SwitchOut}  \cite{wang2018switchout} & Input  & machine translation & not suitable \\   
    \textsc{SignedGraph}  \cite{chen-etal-2020-finding} & Input  & paraphrase & not suitable \\
    \textsc{DAGA}  \cite{ding-etal-2020-daga} & Input+Label  & sequence tagging & not suitable \\
    \textsc{SeqMix} \cite{zhang-etal-2020-seqmix} & Input+Label & active sequence labeling & not suitable \\
    \textsc{GECA} \cite{andreas-2020-good} & Input  & Agnostic & not suitable\\
    \bottomrule
    \end{tabular}
    }
    \caption{ Comparison of selected \textsc{DA} methods adapted from \citet{FengGWCVMH21}. \emph{Level} denotes the depth at which data is modified by the DA. \emph{Task} refers to whether the DA method can be applied to different tasks (a.k.a., task-agnostic), or specifically designed for a task. \emph{Reason} column provides the reason why the method is not included in this paper---``-'' if included.}
    \label{table:comparison}
\end{table*}

\thirdrev{}{To provide more details, the overview of the DA techniques investigated in \citet{FengGWCVMH21} is given in Table~\ref{table:comparison} to justify our selection of techniques. Here, the first category refers to the techniques that are included in this study. They modify the data on the \textbf{input} level, are task agnostic, can preserve the token and relation labels---hence suitable for our sequence tagging tasks) and do not require large amounts of text or models. The methods in the second category~\cite{sennrich-etal-2016-improving,vaibhav-etal-2019-improving,nguyen2020data,wieting-gimpel-2017-revisiting,feng-etal-2019-keep,singh2019xlda,anaby2020not} are not able to preserve the labels. For instance, Semantic Text Exchange (STE)~\cite{feng-etal-2019-keep} aims to replace an entity in a given sentence while modifying the rest of the sentence accordingly. Given the input \textit{``great food , large portions ! my family and i really enjoyed our saturday morning breakfast''} and the entity to be replaced as \textit{pizza}, STE generates a new sentence \textit{``80\% great pizza , chewy crust ! nice ambiance and i really enjoyed it .''}. Since the generated sentence is both syntactically and semantically different from the original sentence, such techniques cannot be used for any of our tasks. Furthermore most of these techniques are tested on English language only and require large amounts of data to generate meaningful paraphrases. The next category of techniques~\cite{Kobayashi18,gao-etal-2019-soft,louvan2020simple} 
benefit from large pretrained language models to replace a token/phrase. Even though such models might exist for some of the low-resource languages, the quality of the models are low due to insufficient training data. Therefore using these models introduces more noise than expected. The next category contains techniques~\cite{feng2020genaug,grundkiewicz-etal-2019-neural} that require external tools/lexicons such as WordNet and spell checkers which are only available for high resource languages. The final category consists of the techniques that are tuned for a specific NLP task. Therefore they can only be used for the specific task, and not for the tasks we focus on in this study. One exception is GECA~\cite{andreas-2020-good} that can be used for parsing low-resource languages. One issue with GECA is that the boundary of extracted fragments can exceed the constituency span, hence there is no guarantee that the fragment would be a subtree.}

This section includes a detailed discussion on three main categories of \thirdrev{}{textual} augmentation techniques that are used in our experiments, namely as \textit{syntactic}, \textit{token-level} and \textit{character-level}. 
A summary of the techniques under each category and their suitability to our downstream tasks can be found in Table~\ref{tab:aug_example}. Finally it discusses the parameters associated with each technique in Sec.~\ref{ssec:params}.

\begin{table*}[!htp]\centering
    \scalebox{.9}{
    
    \begin{tabular}{l*{6}{c}}\toprule
    & &\textbf{POS} &\textbf{DEP} &\textbf{SRL} &\textbf{Generated} \\
    \midrule
    \multirow{6}{*}{\rotatebox[origin=c]{90}{\textbf{Char}}} &&&&&\\
    &\textbf{Char Insert (CI)} &\texttt{x} &\texttt{x} &\texttt{x} &I wrrotle him a legtter \\
    &\textbf{Char Substitute (CSU)} &\texttt{x} &\texttt{x} &\texttt{x} &I wyote him a lettep \\
    &\textbf{Char Swap (CSW)} &\texttt{x} &\texttt{x} &\texttt{x} &I wtore him a lteter \\
    &\textbf{Char Delete (CD)} &\texttt{x} &\texttt{x} &\texttt{x} &I wote him a leter \\
    &&&&&\\
    \midrule
    \multirow{6}{*}{\rotatebox[origin=c]{90}{\textbf{Token}}} &&&&&\\
    &\textbf{Synonym Replacement (SR)} &\texttt{x} &\texttt{x} &\texttt{x} &I wrote him a message \\
    &\textbf{RW Delete (RWD)} &\texttt{x} &\texttt{o} &\texttt{o} &I him a letter \\
    &\textbf{RW Swap (RWS)} &\texttt{x} &\texttt{o} &\texttt{o} &I him wrote a letter \\
    &\textbf{RW Insert (RWI)} &\texttt{o} &\texttt{o} &\texttt{o} &I wrote him a her letter \\
    &&&&&\\
    \midrule
    \multirow{5}{*}{\rotatebox[origin=c]{90}{\textbf{Syntactic}}} &&&&&\\
    &\textbf{Crop} &\texttt{x} &\texttt{x} &\texttt{x} &I wrote a letter \\
    &\textbf{Rotate} &\texttt{x} &\texttt{x} &\texttt{x} & Him a letter I wrote \\
    &\textbf{Nonce} &\texttt{x} &\texttt{x} &\texttt{o} &I wrote him a flower \\
    &&&&&\\
    \bottomrule
    \end{tabular}
    }
    \caption{ Comparison of augmentation methods by means of their task viability. \texttt{x}: Can be used for the task; \texttt{o}: Can not be used for the task. Augmentations are generated on: ``I wrote him a letter''}
    \label{tab:aug_example}
  \end{table*}

\subsection{Character-Level Augmentation}
\label{ssec:char_aug_tec}


    The idea of adding synthetic noise to text applications is not new, however have mostly been used for adversarial attacks or to develop more robust models~\cite{BelinkovB18,KarpukhinLEG19}. Previous work by \citet{KarpukhinLEG19} introduce four types of synthetic noise on orthographic level: character deletion (\textsc{CD}), insertion (\textsc{CI}), substitution (\textsc{CSU}) and swapping (\textsc{CSW}). Additionally, they introduce a mixture of all noise types by sampling from a distribution of 60\% clean (no noise) and 10\% from each type of noise, which we refer to as \textit{Character All} (\textsc{CA}). They show that adding synthetic noise to training data improve the performance on test data with natural noise, i.e., text with real-world spelling mistakes, while not hurting the performance on clean data. The authors experiment on neural machine translation where the source languages are German, French, Czech and the target language is English. We hypothesize that adding right amount of synthetic noise might as well improve the performance on low-resource languages for our set of downstream tasks. 
    For \textsc{CI}, we first build a character vocabulary out of the most commonly used characters in the training set. We do not add noise to one letter words and do not apply \textsc{CSW} to first and the last characters of the token.

    The advantages of character-level synthetic noise are two-fold: First, the output of the augmentation mostly preserves the original syntactic and semantic labels. This is because the resulting tokens are mostly out of vocabulary words that are quite close to the original word---like a spelling mistake. Second, they are trivial to generate, not requiring any external resources like large language models or syntactic annotations. Finally, they are only constrained by the number of characters, that results with the ability of generating huge numbers of augmented sentences, which can be an advantage for most downstream tasks.  

\subsection{Token-Level Augmentation}
\label{ssec:tok_aug_tec}

    This category includes methods that perform token-level changes such as adding, replacing or removing certain tokens. While some preserve the syntax or semantics of the original sentence, majority does not.  

    \paragraph{Synonym Replacement} As one of the earliest techniques~\cite{KolomiyetsBM11,ZhangL15,wang-yang-2015-thats,wei-zou-2019-eda}, it aims to replace words with their synonyms. A lexicon containing synonymity like WordNet, or a thesaurus is generally required to retrieve the synonyms. Since most languages do not have such a resource, some researchers~\cite{wang-yang-2015-thats} exploit special pretrained word embeddings and use k-nearest neighbors (by means of cosine similarity) of the queried word as the replacement. As discussed in Sec.~\ref{sec:relwork}, more recent studies~\cite{Kobayashi18,WuLZHH19,FadaeeBM17a,anaby2020not,kumar2020} employ contextualized language models such as bi-directional LSTMs or BERT~\cite{DevlinCLT19} to find related words such that the class of the sentence is still preserved. However these methods require strong pretrained language models which are generally not available for truly low resource languages. Furthermore, these methods are mostly applied to sentence classification tasks, where the labels are considered coarse-grained compared to our downstream tasks. 
    Considering these, we use a simplified approach similar to \citet{wang-yang-2015-thats} and query the randomly chosen token on non-contextualized pretrained embeddings. Most languages we experiment on are morphologically productive, having out-of-vocabulary words problem. To circumvent this issue we use the subword-level fastText embeddings~\cite{grave2018learning}. 


    \paragraph{Random Word} Similar to character-level noise, one can inject a higher level of noise by randomly inserting (\textsc{RI}), deleting (\textsc{RD}) or swapping (\textsc{RS}) tokens. The EDA framework by~\citet{wei-zou-2019-eda} show the efficiency of these techniques---despite their simplicity---on multiple text classification benchmarks. Following \citet{wei-zou-2019-eda}, we experiment with all techniques except from \textsc{RI}. Since our downstream tasks require contextual annotation of tokens, we can not insert a random word without annotation. Similar to character-level methods, they are easy to apply and can produce extensive amount of synthetic sentences as they are only constrained with the number of tokens. One disadvantage is the inability to preserve the syntactic and semantic labels. For instance deleting a word may yield an ungrammatical sentence without a valid dependency tree. Therefore they are not eligible for two of our tasks: dependency parsing and semantic role labeling. 

    \begin{figure*}[!ht]
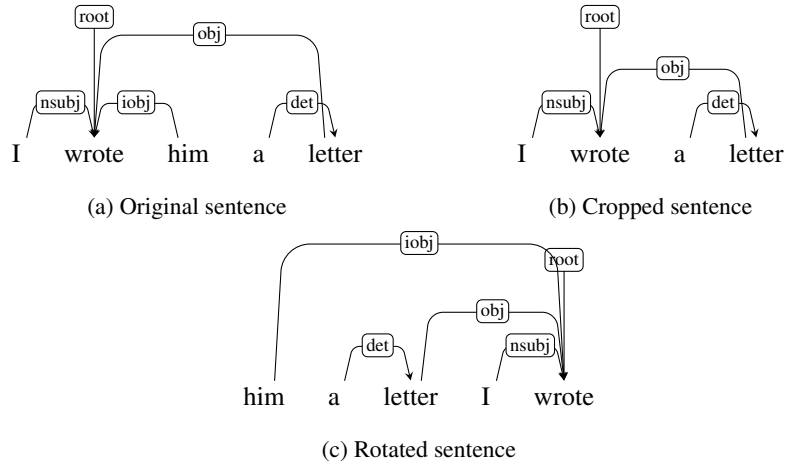

      \centering 
      \begin{subfigure}[b]{0.45\textwidth}
      \centering
        \begin{dependency}
            \begin{deptext}[column sep=1.2em]
                I \& wrote \& him \& a \& letter \\
            \end{deptext}
            \deproot{2}{root}
            \depedge{1}{2}{nsubj}
            \depedge{3}{2}{iobj}
            \depedge{4}{5}{det}
            \depedge{5}{2}{obj}
        \end{dependency}
        \caption{Original sentence}
      \end{subfigure}
      \begin{subfigure}[b]{0.45\textwidth}
      \centering 
        \begin{dependency}
            \begin{deptext}[column sep=1.2em]
                I \& wrote \& a \& letter \\
            \end{deptext}
            \deproot{2}{root}
            \depedge{1}{2}{nsubj}
            \depedge{3}{4}{det}
            \depedge{4}{2}{obj}
        \end{dependency}
        \caption{Cropped sentence}
      \end{subfigure}
      \begin{subfigure}[b]{0.45\textwidth}
      \centering 
        \begin{dependency}
            \begin{deptext}[column sep=1.2em]
                him \& a \& letter \& I \& wrote \\
            \end{deptext}
            \deproot{5}{root}
            \depedge{4}{5}{nsubj}
            \depedge{1}{5}{iobj}
            \depedge{2}{3}{det}
            \depedge{3}{5}{obj}
        \end{dependency}
        \caption{Rotated sentence}
      \end{subfigure}
      \caption{ Demonstration of the syntactic augmentation techniques \textit{crop} and \textit{rotate}.}  
    \label{fig:syntactic_demo}
    \end{figure*}

    \subsection{Syntactic Augmentation}
    \label{ssec:syn_aug_tec}

    This category consists of more sophisticated methods that benefit from syntactic properties to generate new sentences. The main disadvantages of this category are (i) the need for syntactic annotation, and (ii) being more constrained, i.e., not being able to generate as many data points as previous categories. 
          
    \paragraph{Nonce} This technique, introduced by \citet{GulordavaBGLB18}, aims to produce dummy sentences by replacing some of the words in the original sentence, such that the produced sentence is syntactic equivalent of the original while not preserving the original semantics. In more details, randomly chosen content words, i.e., noun, adjective and verb, are replaced with words that have the same part of speech tag, morphological tags and dependency label. Given the sentence \textit{``Her sibling bought a cake''}, the following sentences can be generated: \textit{``\textbf{My} sibling \textbf{saw} a cake''} or \textit{``His \textbf{motorbike} bought a \textbf{island}''}. As can be seen, the generated sentences mostly do not make any sense, however syntactically equivalent. 
    For this reason, it can only be employed for syntactic tasks and not for SRL. 

    \paragraph{Crop} This augmentation algorithm by~\citet{SahinS18}, morphs the dependency trees to generate new sentences. 
    The main idea is to remove some of the dependency links to create simpler, shorter but still (mostly) meaningful sentences. In order to do so, \citet{SahinS18} defines a list of dependency labels which is referred to as Label Of Interest (LOI) that attaches subjects and direct, indirect objects to the predicates. Then a simpler sentence is created by remaining only one LOI at a time. 
    Following \citet{SahinS18}, we \textit{only} consider the dependency relations that attach subjects and objects to predicates as LOIs; and ignore other adverbial dependency links involving predicates. We keep the augmentation non-recursive, i.e., we only reorder the first level of flexible subtrees, hence the flexible chunks inside these subtrees are kept fixed. The idea is demonstrated in Figure~\ref{fig:syntactic_demo}. 

    \paragraph{Rotate} It uses a similar idea to \textit{image rotation}, where the sentence is \textit{rotated} around the \texttt{root} node of the dependency tree. The method uses the same LOIs as cropping and creates flexible chunks that can be reordered in the sentence. A demonstration of rotation is shown in Figure~\ref{fig:syntactic_demo}. 
    Both the cropping and rotation operations may cause semantic shifts and ill-formed sentences, mostly depending on the morphological properties of the languages. For instance, languages that use case marking to mark subjects and objects can still generate valid sentence forms via rotation operation simply because the word order is more flexible~\cite{wordOrder}. However, most valid sentences may still sound strange to a native speaker, since there is still a \textit{preferred order} and the generated sentence may have an infrequent word order. Furthermore, for languages with a strict word order like English, rotation may introduce more noise than desired. Finally, the augmented sentences may still be beneficial for learning, since they would provide the model with more examples of variations in semantic argument structure and in the order of phrases. 

\subsection{Parameters}
\label{ssec:params}

    All of the aforementioned augmentatiom methods are \textit{parametric}. As we show later in Sec.~\ref{sec:results}, the choice of parameters may have a substantial impact on the success of the methods. The parameters and their value range are defined as below:

    \paragraph{Ratio} This parameter is used to calculate the number of new sentences to be generated from one original sentence. To ensure that more sentences are generated from longer sentences, it is simply calculated as $ratio*|sentence|$, i.e., for a sentence of length 10, only 1 extra sentence will be produced with $ratio=0.1$. It is only used for orthographic methods since syntactic methods are constrained in other ways (e.g., number of tokens that share morphological features and dependency labels). The values we experiment with are as follows: $[0.1, 0.2, 0.3, 0.4]$.

    \paragraph{Probability} This determines the probability of a token-level augmentation, i.e., the ratio of the augmented tokens to sentence length. For word insertion, deletion and all character-based methods (\textsc{CI}, \textsc{CD}, \textsc{CSU}, \textsc{CSW}, \textsc{CA}), it can be interpreted as the ratio of tokens that undergo a change to the number of total tokens; while for word swapping operation, it refers to the ratio of swapped token pairs to total number of tokens. Moreover, it determines the number of characters to which the augmentation are applied in case of character-level augmentation. Similar to ratio, we experiment with the value range: $[0.1, 0.2, 0.3, 0.4]$. For the syntactic methods \textsc{crop} and \textsc{rotate}, probability is associated with the operation dynamics, i.e., when the operation probability is set to 0.2, the expected number of additional sentences produced via cropping would be $\#LOI/5$, where $\#LOI$ indirectly refers to the number of valid subtrees for the operation. Since the number of additional sentences are already constrained with $\#LOI$, we also experiment with higher probabilities for crop and rotate. The range is then $[0.1, 0.3, 0.5, 0.7, 1.0]$. 

    \paragraph{Max Sentences} In case of considerably long sentences or substantial number of available candidates, the number of augmented sentences can grow rapidly. This sometimes causes an undesired level of noise in the training data. To avoid this, the number of maximum sentences, a.k.a., sentence threshold parameter, is commonly used. Following \citet{GulordavaBGLB18,vaniaKSL19} we experiment with $5$ and $10$ as the threshold values.

\section{Experimental Setup}
\label{ssec:exper_set}

  First we provide detailed information on the downstream tasks and the languages we have experimented on. Next we discuss the datasets that are used in our experiments. 

  \subsection{Tasks}
  \label{ssec:tasks}
  
We focus on three sequence tagging tasks that are central to NLP: POS-tagging (POS), dependency parsing (DEP) and semantic role labeling (SRL). POS and DEP require processing at the morphological and syntactic level and are considered as crucial steps toward understanding natural language. Furthermore the performance on POS and DEP tasks are likely to repeat for other tasks that require syntactic and morphological information such as data-to-text generation. On the other hand, SRL is more useful to gain insights on the augmentation performance where preserving the semantics of the original sentence is essential. Therefore it can serve as a proxy for higher-level tasks that necessitate semantic knowledge such as question answering or natural language inference. 


\secrev{For POS and SRL, we use subword level bi-directional LSTM based models described in~\cite{SteedmanS18} and \cite{SahinS18}; and for dependency parsing, we use the transition-based Uppsala parser v2.3~\cite{delhoneux17arc,KiperwasserG16}. For all tasks, all tokens are lowercased and sentences longer than 200 tokens are omitted from the training set.}
{The aim of this study is to analyze the performance of various augmentation techniques for low-resource languages in a systematic way, rather than implementing state-of-the art sequence taggers which would generally require additional resources and engineering effort. Furthermore, this study involves a large number of experiments comparing eleven augmentation techniques across three tasks and several languages in a multi-run setup. For these reasons, we have chosen models that are not heavily engineered and are modular, but also with proven competence on diverse set of languages. For all tasks, we use neural models that operate on subword-level rather than words. This is necessary to tackle the out of vocabulary problem which is inevitable for languages with productive morphology, especially in low resource scenarios. Furthermore we do not use any additional features (e.g., POS tags, pretrained word embeddings), external resources or ensemble of multiple models. Considering these facts, for our POS tagging experiment, we use the subword-level sequence tagging model~\cite{HeinzerlingS19multi} that is both modular and provides results which  are on par with state-of-the-art on many languages. The model employs an autoregressive architecture (e.g., RNN or bi-LSTM) for random or non-contextual subwords; and uses fine-tuning paradigm for large pretrained language models. For our dependency parsing experiments, we use the transition-based Uppsala parser v2.3~\cite{delhoneux17arc,KiperwasserG16} which achieved the second best average LAS performance~\footnote{The best performing model~\cite{RosaM18} is a cross-lingual model that trains a single model together using multiple treebanks, which was not applicable to our scenario.} on low-resource languages in CoNLL 2018 shared task~\cite{ZemanHPPSGNP18}. 
Additionally, we experiment with a biaffine dependency parser built on top of large pretrained contextualized word representations~\cite{GlavasV21}, since fine-tuning such models on downstream tasks has recently achieved state-of-the-art results on many tasks and languages. Finally, we use a character-level bi-directional LSTM model~\cite{SahinS18} to conduct SRL experiments\footnote{To the best of our knowledge, there does not exist an open-source dependency-based SRL model built on top of large contextualized language models with proven competence on diverse set of languages, and implementing such a model is out of scope of this paper.}. More details on the models and their configuration are given in the following subsections.}   

\subsubsection{POS Tagging} The goal is to associate each token with its corresponding lexical class, i.e., syntactic label (e.g., \textit{noun, adjective, verb}). Although it is a token level task, disambiguation using contextual knowledge is mostly necessary as one token may belong to multiple classes (e.g., to fly \textit{(Verb)} or a fly \textit{(Noun)}). For languages with rich morphology, it is generally referred to as morphological disambiguation while the correct morphological analysis---including the POS tag---is chosen among multiple analyses. For analytic languages like English, it is mostly performed as the first step in the traditional NLP pipeline. For this task, we inherit the universal POS tag set that are shared among languages and defined within the scope of Universal Dependencies project~\cite{ud26}. 


\paragraph{Subword Units} \secrev{}{We experiment with three subword units: characters, BPEs and Word Pieces. Characters and character n-grams have been one of the most popular subword units~\cite{LingDBTFAML15}, since they (i) don't require any preprocessing, (ii) are language-agnostic and (iii) computationally cheap due to the small vocabulary size. Byte Pair Encoding~(BPE)~\cite{SennrichHB16a} is a simple segmentation algorithm that learns a subword vocabulary by clustering the frequent character pairs together for a predefined number of times. The algorithm only requires raw text---hence language-agnostic, and computationally simple. Furthermore, it has been shown to improve the performance on various NLP tasks, especially machine translation. \citet{Heinzerling018} trained BPE embeddings using the GloVe~\cite{PenningtonSM14} word embedding objective and made it available for 275 languages with multiple vocabulary sizes. However neither randomly initialized character embeddings nor gloVe-BPE embeddings are aware of context. Therefore, we also experiment with contextualized embeddings (i.e., different embeddings for the same subword depending on the context) that operate on Word Pieces. For this study we choose BERT~\cite{DevlinCLT19}: a Transformer~\cite{VaswaniSPUJGKP17} based contextualized language model that recently led to state-of-the-art results for many languages and tasks. We use the publicly available multilingual BERT~(mBERT)\footnote{https://github.com/google-research/bert/blob/f39e881/multilingual.md} that has been trained on the top 100 languages with the largest Wikipedia using a shared vocabulary across languages. Some of the low-resource languages we use in our experiments are not part of mBERT's languages. 
} 

\paragraph{The model} \secrev{}{
The overall architecture of the sequence tagging model used in this study~\cite{HeinzerlingS19multi} is given in Fig.~\ref{fig:postagsota}. Even though the modular architecture enables combining different subwords, we experiment with the units separately for the sake of measuring the individual effect. 
\begin{figure}
\centering
    \includegraphics[width=0.65\textwidth]{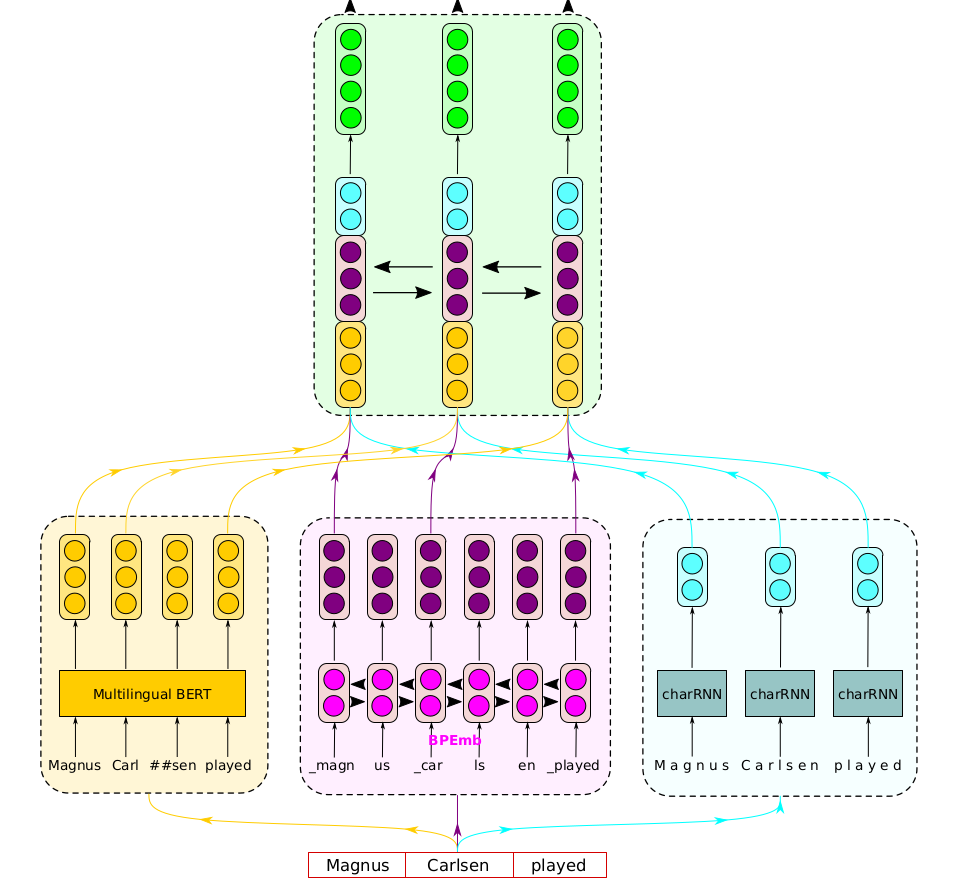}
  \caption{ General architecture of the sequence tagger taken from \citet{HeinzerlingS19multi}.}
  \label{fig:postagsota}
\end{figure}
For the character and BPE-level model, first each word is split into a sequence of subwords produced by the $\rho$ function. For characters, each subword unit is randomly initialized, whereas, for BPE, pretrained embeddings are looked up.  
\begin{gather} 
   \rho(w) = {sub_0,sub_1,..,sub_n}
\end{gather}
For character and BPE-level models, the sequence is encoded via an RNN and a bi-LSTM network respectively. For BERT-based model, encoder is simply the pretrained Transformer that is fine-tuned during training, therefore no additional encoding is performed. For the character-level model, the \textit{final} hidden states are used to represent the token, while for BPE and mBERT, only the states of the first subword at each token are used. 
\begin{gather} 
   \vec{hs_f}, \vec{hs_b} = \text{Encoder}(\rho(w)) \\
   \vec{w} = [\vec{hs_f};\vec{hs_b}]
\label{eq:comp}
\end{gather}
Next, the token embeddings, $\vec{w}$, are passed onto another bi-LSTM layer, where $i$ denotes the index of the word.  
\begin{gather} 
   \vec{h_{f}}, \vec{h_{b}} = \text{bi-LSTM}(\vec{w_{i}})
 \end{gather}
Next, the concatenated hidden states from forward, $h_{f}$, and backward directions, $h_{b}$, are fed to a classification layer to project the feature space onto the label space. Finally, the probability distribution of each POS tag is calculated via a softmax function. The label with the highest probability is assigned to the input token as follows, where $S$ represents the sentence and $\vec{L}$ denotes the POS tag sequence. 
\begin{gather} 
    \vec{p(L_i|S)} = softmax(W_{l}\cdot[\vec{h_{f}};\vec{h_{b}}]+\vec{b_{l}})
\end{gather}
}\secrev{We used uniform initialization for all network components. The model is optimized via stochastic gradient descent with initial learning rate as 1. Gradients above $2$ are clipped. We trained the models for 50 epochs, and decrease the learning rate by half when the results on development set do not improve after 3 epochs. The embedding size for characters as well as hidden size of intermediate layers are chosen as 128. Batch size is set to 8 due to small training set. Number of bi-LSTM layers of all bi-LSTM components are left as a parameter and the values 1 and 2 are used in the experiments. Accuracy is used to evaluate the POS tagging results.}{BERT is fine-tuned during training, i.e., the model's weights are updated by backpropogating through all layers. For each language, we use the best vocabulary size reported in previous work~\cite{HeinzerlingS19multi}. For each subword-level model, we use the parameters from the original work~\cite{HeinzerlingS19multi}.}

\subsubsection{Dependency Parsing} It aims to provide a viable structural or grammatical analysis of a sentence by finding the links between the tokens. It assumes that the dependent word is linked to its parent, a.k.a head word, with one of the dependency relations such as \textit{modifier, subject} or \textit{object}. The resulting grammatical analysis is called a dependency graph, shown in Fig.~\ref{fig:syntactic_demo}. 
We use the universal dependency label sets defined by Universal Dependencies project~\cite{ud26} and report the Labeled Attachment Score (LAS) as the performance measure. \secrev{}{For the experiments, we use two different models: uuparser~\cite{KiperwasserG16,delhoneux17arc} and the biaffine parser finetuned on large contextualized language models~\cite{GlavasV21}.} 

\paragraph{uuparser} \secrev{}{The parser is based on the transition-based \citet{KiperwasserG16} that uses bi-LSTMs to create features. Here $c$ refers to character and $e$ refers to embedding. First, a character based representation is generated via bi-LSTMs. Next, the non-contextual representation for a token is created by concatenating an embedding for the token itself, which we initialize randomly. 
\begin{gather} 
   \vec{e_c} = \text{bi-LSTM}({c_0,c_1,..,c_n}) \\
   \vec{w} = [\vec{e_w};\vec{e_c}]
\end{gather}
Then we create a context-aware representation for the token at index $i$ using an additional bi-LSTM layer:
\begin{gather} 
   \vec{h_{i}} = \text{bi-LSTM}(\vec{w_{i}})
   \label{eq:dep_rep}
 \end{gather}}
We use the arc-hybrid transition-based parser that is later extended with partially dynamic oracle and \textsc{Swap} transition to be able to create non-projective dependency graphs. A typical transition-based parser consists of so-called configurations and a set of transitions, a.k.a., actions, which can be applied to a configuration to create a new configuration. Parsing starts with a predefined initial configuration. At each iteration, a classifier chooses the best transition given the features extracted from the configuration, and updated the configuration. This step is repeated until a predefined terminal configuration. The arc-hybrid parser used in this work, defines configuration as a stack, a buffer and a set of dependency arcs. \secrev{}{Then the feature for the configuration is created by concatenating the representations of a fixed number of tokens calculated via Eq.~\ref{eq:dep_rep} from the top of the stack and the first element of the buffer.} Finally, a scoring function that is implemented as a multi-layer perceptron, assigns scores to transitions and dependency labels, given the extracted feature. The transition and the label with the highest score is then used to create the next configuration. \secrev{We used the default settings of the uuparser~\cite{KiperwasserG16,delhoneux17arc}\footnote{Available at \url{https://github.com/UppsalaNLP/uuparser}}. Similar to POS tagging, embedding sizes are set to 128, and the training is performed for 100 epochs. We report the Labeled Attachment Score (LAS) as the performance measure.}{We train separate models for each treebank, using \textit{only} randomly initialized character and word features.} 

\paragraph{biaffine}\secrev{}{The parser consists of an attention layer on top of the outputs from a transformer-based language model, as shown in Fig.~\ref{fig:biaffine}. If a token consists of multiple subword segments, one token representation is created by averaging the transformer outputs for each segment, denoted with $\mathbf{X}$. $\mathbf{X} \in \mathbb{R}^{N \times H}$ is then used as the representation of syntactic dependents, where $N$ and $H$ refer to the number of tokens in the sentence and the transformer hidden state size. To represent the \texttt{root} node, the transformer representation for the \texttt{[CLS]}: $\mathbf{x}_\mathit{CLS}$, a.k.a., the sentence start token, is used. To represent the dependent heads, $\mathbf{X}' = [\mathbf{x}_\mathit{CLS}; \mathbf{X}] \in \mathbb{R}^{(N+1) \times H}$ is defined. 
\begin{figure}
\centering
    \includegraphics[width=0.5\textwidth]{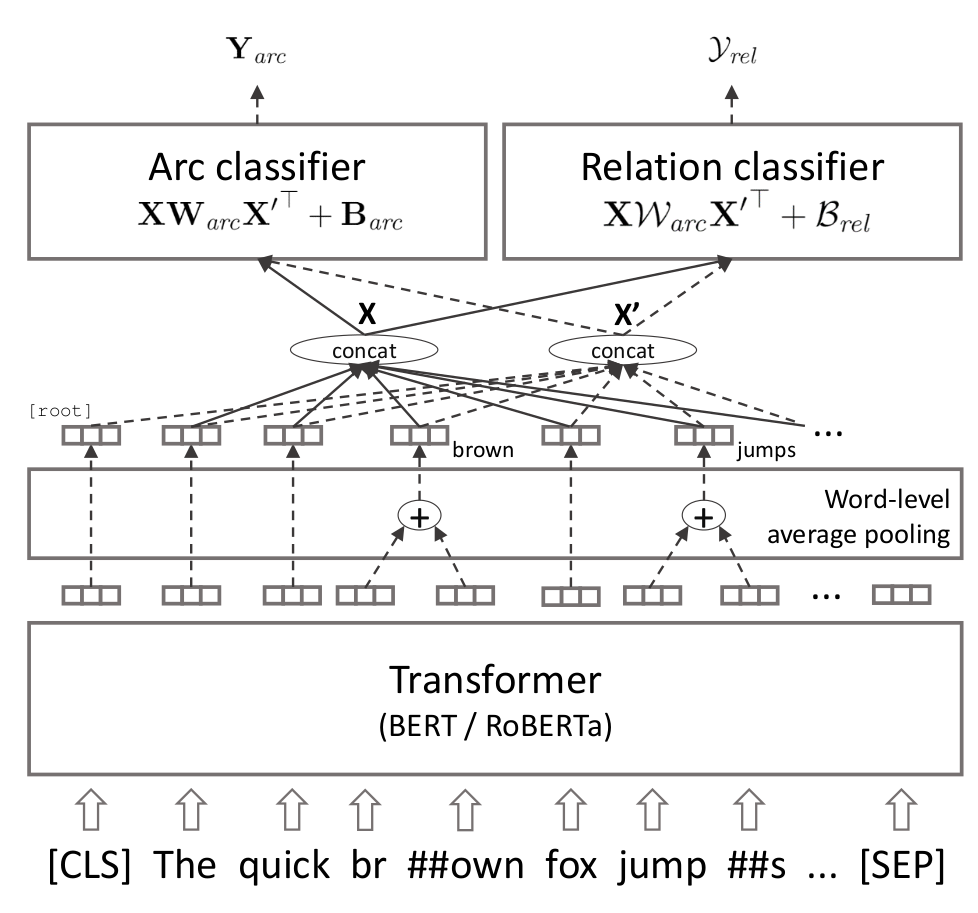}
  \caption{ Transformer based architecture of the biaffine parser taken from \citet{GlavasV21}.}
  \label{fig:biaffine}
\end{figure}
The arc and relation scores are then calculated as biaffine products as following:
\vspace{-1em}
{
\small
\begin{equation*}
    \mathbf{Y}_{\mathit{arc}} = \mathbf{X}\mathbf{W}_\mathit{arc}\mathbf{X'}^{\top} + \mathbf{B}_\mathit{arc};\hspace{0.5em}
    \mathbf{Y}_{\mathit{rel}} = \mathbf{X}\mathbf{W}_\mathit{rel}\mathbf{X'}^{\top} + \mathbf{B}_\mathit{rel}
\end{equation*}}
\noindent where $\mathbf{W}$ and $\mathbf{B}$ refer to the weight and bias parameters for the arc and relation classifiers. The correct dependency head is selected simply as the row corresponding to the maximum score in $\mathbf{Y}_{\mathit{arc}}$. The arc and relation classification losses are defined as cross-entropy losses respectively over the sentence tokens and gold arcs.
}

\subsubsection{Semantic Role Labeling} Semantic Role Labeling (SRL), a.k.a. \textit{shallow semantic parsing}, is defined as analyzing a sentence by means of predicates and the arguments attached to them. A wide range of semantic formalisms and annotation schemes exist however the main idea is labeling the arguments according to their relation to the predicate.
\begin{quote} 
  $[$I$]$\textsubscript{A0: buyer} $[$bought$]$\textsubscript{buy.01: purchase} $[$a new headphone$]$\textsubscript{A1: thing bought} from $[$Amazon$]$\textsubscript{A2: seller}
\end{quote} 
The example given above shows a labeled sentence with English Proposition Bank~\cite{martha2005proposition} semantic roles, where \textit{buy.01} denotes the first sense of the verb ``buy'', and \textit{A0, A1} and \textit{A2} are the numbered arguments defined by the predicate's semantic frame. For this study, we perform dependency-based SRL, which means that only the head word of the phrase (e.g., \textit{headphone} instead of \textit{a new headphone}) will be detected as an argument and will be labeled as \textit{A1}. To evaluate SRL results, we used the official CoNLL-09 evaluation script on the official test split. The script calculates the macro-average F1 scores for the semantic roles available in the data. 

\paragraph{The model} \secrev{}{Similar to previous models, each token is segmented into subwords. For SRL, we only use characters as the subword unit, since it provided competitive results for many languages~\cite{SahinS18}.
\begin{gather} 
   \rho(w) = {sub_0,sub_1,..,sub_n}
\end{gather}
Afterwards, following \citet{LingDBTFAML15}, a weighted composition of the hidden states: $hs_f$ from forward and $hs_b$ from backward direction are calculated and used as the token embedding as given in Eq.~\ref{eq:comp}.
\begin{gather} 
   \vec{hs_f}, \vec{hs_b} = \text{bi-LSTM}({sub_0,sub_1,..,sub_n}) \\
   \vec{w} = W_f \cdot \vec{hs_f} + W_b \cdot \vec{hs_b} + b
\label{eq:comp}
\end{gather}
In order to mark the predicate of interest, we concatenate a predicate flag $pf_i$ to $\vec{w}$ calculated as in Eq.~\ref{eq:comp}. It is simply defined as $1$ for the predicate of interest and $0$ for the rest. Next, $\vec{x_{i}}$s, are passed onto another bi-LSTM layer, where $i$ denotes the index of the word. 
\begin{gather} 
	\vec{x_{i}} = [\vec{w};pf_i] \\
	\vec{h_{f}}, \vec{h_{b}} = \text{bi-LSTM}(\vec{x_{i}})
\end{gather}
The probability distribution of each semantic role label is finally calculated via a softmax function given the final bidirectional hidden states: $h_{f}$ and $h_{b}$, and the label with the highest probability is assigned to the input token.
\begin{gather} 
\vec{p(A_i|S,P)} = softmax(W_{l}\cdot[\vec{h_{f}};\vec{h_{b}}]+\vec{b_{l}})
\end{gather}
We use the default model parameters reported in \citet{SahinS18}.
}

\begin{gather} 
   \lambda y \ \lambda x \ add (x,y) \\
   \lambda y \ \lambda x \ add (x,y) (shopping list) \\
   \lambda x \ add (x,shoppinglist) (ingredient)
\end{gather}

  \subsection{Languages}
  \label{ssec:lang}
\begin{figure*}
\centering
    \includegraphics[width=1.0\textwidth]{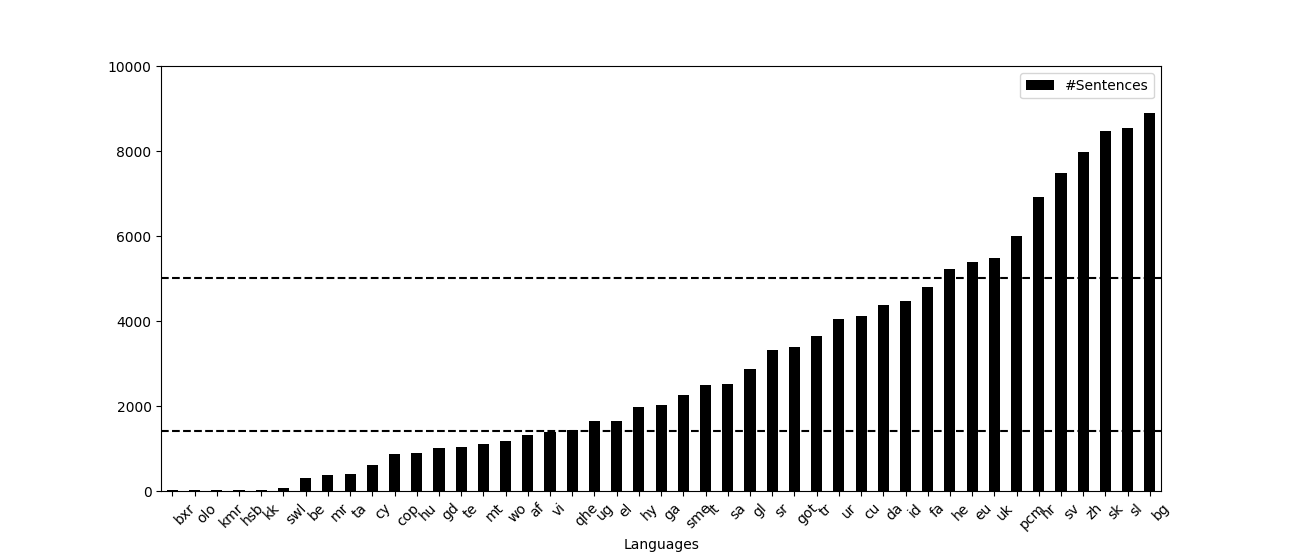}
  \caption{ \#Training sentences per UD treebank v2.6. Languages are represented with ISO codes. First line marking the first quartile and second marking the median. Languages with more than 10K training sentences are not shown.}
  \label{fig:langs}
\end{figure*}

  We set the definition of \textit{low-resource language} based on the number of training sentences available in UD v2.6~\cite{ud26}. First we calculate the quartiles via ordering the treebanks with respect to their training data size. We then define the languages under the first quartile as low-resource languages as shown in Fig.~\ref{fig:langs}. According to this definition, the list of low-resource languages are as follows: Kazakh, Tamil, Welsh, Wolof, Upper Serbian, Buryat, Swedish sign language, Coptic, Gaelish, Marathi, Telugu, Vietnamese, Kurmanji, Livvi, Belarusian, Maltese, Hungarian and Afrikaans.

Finally we choose a subset of languages that are from diverse language families: Kazakh (Turkic), Tamil (Dravidian), Buryat (Mongolic), Telugu (Dravidian), Vietnamese (Austro-Asiatic), Kurmanji (Indo-European (IE), Iranian) and Belarusian (IE, Slavic). Although this list of languages can be experimented on for the tasks of POS tagging and dependency parsing, there is no data available for any of these languages for semantic role labeling. Semantically annotated datasets are only available for a handful of languages such as Turkish (Turkic), Finnish (Uralic), Catalan (IE, Romance), Spanish (IE, Romance) and Czech (IE, Slavic). Therefore, we simulate a low-resource environment by sampling for SRL. \secrev{}{We provide a summary of languages in Table~\ref{tab:language_table} and discuss the relevant properties of each language family below.}  
\paragraph{Indo-European (IE)} We have five representatives for this language family: Kurmanji, Belarusian, Catalan, Spanish and Czech. The representative languages are from various branches: Iranian, Slavic and Romance. From the typological perspective, all languages have fusional characteristics. In other words, morphemes (prefixes, suffixes) are used to convey some linguistic information such as gender, however one morpheme can be used to mark multiple properties. Slavic languages are known to have around 7 distinct case markers, while others are not as rich. The number and type of case markers available in Slavic helps to relax the word order.  
\paragraph{Uralic and Turkic} The languages from both families are known to be agglutinative, meaning that there is one-to-one mapping between morpheme and meaning. All representative languages, namely as Kazakh (Turkic), Turkish (Turkic), Hungarian (Uralic) and Finnish (Uralic), attach morphemes to words extensively. These languages have a high morpheme to word ratio, and a comprehensive case marking system. 
\paragraph{Dravidian} Two languages, namely as Tamil and Telugu, are from Dravidian language family but in different branches. Similar to Uralic and Turkic, Dravidian languages are also agglutinative, and have extensive grammatical case marking (e.g., Tamil defines 8 distinct markers). Unlike other language families, it is not based on Latin alphabet. 
\paragraph{Austro-Asiatic} Vietnamese is the only representative of this family. Unlike previous ones, Austro-Asiatic languages are analytic. For instance, Vietnamese does not use any morphological marking for tense, gender, number or case; i.e., has a low morphological complexity. 
\paragraph{Mongolic}: As being a language from Mongolic family, Buryat is an agglutinative language with eight grammatical cases. Modern Buryat uses an extended Cyrillic alphabet.  

\secrev{}{
\begin{table*}[!htp]\centering
    \scalebox{.9}{
    
    \begin{tabular}{lll}\toprule
    \textbf{Language} & \textbf{Family} & \textbf{Typology} \\
    \midrule
    Belarusian & Indo-European (IE), Slavic &  Fusional \\
    Buryat & Mongolic & Agglutinative \\
    Catalan & Indo-European (IE), Romance & Fusional\\
    Czech & Indo-European (IE), Slavic & Fusional\\
    Finnish & Uralic & Agglutinative \\
    Kazakh & Turkic & Agglutinative \\
    Kurmanji & Indo-European (IE), Iranian & Fusional \\
    Spanish & Indo-European (IE), Romance & Fusional \\
    Tamil & Dravidian & Agglutinative \\
    Telugu & Dravidian & Agglutinative \\
    Turkish & Turkic & Agglutinative \\
    Vietnamese & Austro-Asiatic & Analytic \\ 
    \bottomrule
    \end{tabular}
    }
    \caption{ The list of languages together with their corresponding language and typological families}
    \label{tab:language_table}
  \end{table*}
  }

  \subsection{Datasets}
  \label{ssec:datasets}
  We use the Universal Dependencies v2.6 treebanks~\cite{ud26} for POS and DEP. Some of the languages such as Kazakh, Kurmanji and Buryat, do not have any development data. For those languages, we randomly sample 25\% of the training data to create a development set. For the SRL task, we use the datasets distributed by Linguistic Data Consortium (LDC) for Catalan (CAT) and Spanish (SPA)~\footnote{Catalog numbers are as follows: LDC2012T03 and LDC2012T04}. In addition, we use dependency-based annotated SRL resources released for Finnish (FIN)~\cite{Haverinen2015} and Turkish (TUR)~\cite{csahinannotation,csahin2016verb,sulubacak2018implementing,sulubacak2016imst}. All proposition banks are derived from a language-specific dependency treebank, and contain semantic role annotations for verbal predicates. We provide the basic dataset statistics for each language in Table~\ref{tab:stats}. Since SRL resources are not available for truly low-resource languages, we sample a small training dataset from the original ones, shown with \#sampled in Table~\ref{tab:stats}. Each SRL dataset employs a language-specific dependency and semantic role annotation scheme. Therefore, we needed to perform language specific preprocessing for cropping and rotation augmentation techniques given in Appendix.


\begin{table}
  \centering
    \scalebox{1.0}
    { 
      
      \begin{tabular}{llrrr}  
       &  & \textbf{\#training} & \textbf{\#dev} & \textbf{\#test} \\
      \hline 
      \multirow{5}{*}{SRL} & Czech & \#sampled & 5228 & 4213 \\ 
      & Catalan & \#sampled & 1724 & 1862 \\ 
      & Spanish & \#sampled & 1655 & 1725 \\          
      & Turkish & \#sampled & 844 & 842 \\
      & Finnish & \#sampled & 716 & 648  \\ 
      \hline
      \multirow{8}{*}{POS \& DEP} & Vietnamese & 1400 & 800 & 800 \\ 
      & Telugu & 1051 & 131 & 146  \\
      & Tamil & 400 & 80 & 120 \\     
      & Belarusian & 319 & 65 & 253 \\ 
      & Kazakh* & 23 & 8 & 1047 \\
      & Kurmanji* & 15 & 5 & 734 \\ 
      & Buryat* & 14 & 5 & 908 \\
      \end{tabular}
     }
  \caption{ Dataset statistics for all tasks, languages and splits. \#sampled: Ranges between 250-1000. *: No original development set.}
  \label{tab:stats}
\end{table}

\section{Experiments and Results}
\label{sec:results}

\secrev{}{We perform experiments on three downstream tasks: POS, DEP and SRL using the models described in Sec.~\ref{ssec:tasks}. We run each experiment 10 times using the same set of random seeds for the model that operates on the original (unaugmented) and augmented datasets. We compare augmentation techniques to their corresponding unaugmented baselines using a paired t-test and report the p-values. We use \colorbox{darkgreen}{dark green} for $p \leq 0.05$ and \colorbox{lightgreen}{light green} for $ 0.05 < p \leq 0.1$ to highlight the cases where the improvements over the baselines are statistically significant. Similarly we use \colorbox{darkred}{dark red} and \colorbox{lightred}{light red} to denote significantly lower scores.}
\thirdrev{}{Additionally, we use the sign (**) for $p \leq 0.05$ and (*) for $ 0.05 < p \leq 0.1$ to highlight the cases where the improvements over the baselines are statistically significant, and similarly the symbol ($\dagger\dagger$) and ($\dagger$) to denote significantly lower scores. Color-blind friendly results are additionally given in Appendices.}

\subsection{POS Tagging}
\label{ssec:pos_res_sec}

  \secrev{}{POS Tagging is considered as one of the most fundamental NLP tasks, such that it has been the initial step in the traditional NLP pipeline for quite a long time. Even in the era of end-to-end models, it has been shown to improve the performance of the downstream tasks of higher complexity when employed as a feature. 
  Taking the importance of POS tagging on higher-level downstream tasks into account, we conduct experiments with different subword units and training regimes (see Sec.~\ref{ssec:tasks} for details) to examine (i) whether the behavior of the augmentation methods are model-agnostic, and (ii) whether the improvements are relevant for state-of-the-art models which are fine-tuned on large pretrained multilingual contextualized language models. The results for the character~(\texttt{char}), BPEs~(\texttt{BPE}) and multilingual BERT~(\texttt{mBERT}) are given in Table~\ref{tab:postag_res_multi}. The languages that lack the corresponding pretrained model (e.g., Kurmanji and \texttt{BPE}) are not shown in the table.}

  \secrev{}{\paragraph{\texttt{char}} Except from Telugu, at least one of the techniques has significantly improved over the baselines for all other languages. The token-level methods, \textit{RWD} and \textit{RWS} have increased the scores of Kazakh and Tamil significantly, while providing a slight increase on Belarusian and Kurmanji. On the other hand, \textit{SR}, results in a mixture of increase and decrease in the scores. For instance, Buryat and Kazakh POS taggers benefit significantly from \textit{SR}, while the opposite pattern is observed for Belarusian, Kurmanji and Telugu. Unlike token-level methods, the character-level ones: \textit{CI}, \textit{CSU}, \textit{CSW}, \textit{CD} and \textit{CA}, seem to bring more consistent improvements. In particular, Belarusian, Kazakh and Tamil POS taggers are significantly improved by most of the character-level techniques, while Vietnamese tagger only benefited from \textit{CSU} and \textit{CA} and slightly hurt by \textit{CD}. In majority of the cases, syntactic methods either slightly decrease the performance or have not improved significantly. One exception is the \textit{Nonce} technique, which led to a significant improvement for Kazakh, slight improvements for Buryat and Tamil.}

  \secrev{}{\paragraph{\texttt{BPE}} In general, \texttt{BPE} scores are slightly lower than \texttt{char}, which can be considered as more room for improvement. Similar to \texttt{char}, we observe improvements over the baselines by at least one technique with the exception of Tamil. On the contrary of \texttt{char}, token-level techniques \textit{RWD} and \textit{RWS} significantly increase the scores for Belarusian, Buryat, Kazakh, Telugu and Vietnamese. Similar to \texttt{char}, \textit{SR} yields significantly higher scores for Buryat and Kazakh, while significantly reducing the performance for Belarusian and Telugu. While bringing consistent and significant improvements for Buryat and Kazakh, not many character-level methods lead to higher scores for the other languages. Finally, except the case for Buryat and Kazakh, the improvements from syntactic techniques are not significant, while the performance drop might be severe in cases like Vietnamese.}

  \secrev{}{\paragraph{\texttt{mBERT}} As expected, the mean baseline scores are the highest for most languages, however the augmentation methods still provide significant gains in some cases. The most significant improvements are achieved in Belarusian by \textit{CSU} and in Kazakh by \textit{RWS}, \textit{CSW} and \textit{Nonce}. Unlike \texttt{char} and \texttt{BPE}, we do not observe a distinct increase or decrease pattern in certain groups of techniques, except from \textit{CA} that achieved significantly higher scores for all languages. However by manual comparison of the improved/worsened results, the pattern is found to be closer to \texttt{char} than to \texttt{BPE}.}

  \paragraph{Summary and Discussion} \secrev{We found that augmentation boosts the performance of Belarus, Tamil, Kazakh and Kurmanji POS taggers, while benefiting Buryat POS tagging substantially when the right configuration is set. On the other hand, we haven't observed any significant gains for Vietnamese and Telugu which are first three largest treebanks. Although in majority of the cases augmentation techniques performed either \textit{statistically} on par or better, there were a few cases where the accuracy deteriorated: \textit{SR} for Tamil and Telugu; and \textit{crop} and \textit{rotate} for Tamil and Vietnamese. We believe that this is due to: (i) pretrained embeddings do not guarantee a syntactic equivalent of a token, and (ii) quality of embeddings also suffer from low-resources, i.e., small wikipedia.}{To summarize, we observe significant improvements over the corresponding baselines by a group of augmentation techniques on certain languages for \textit{all} experimented models. The token-level methods provide more significant improvements for \texttt{BPE} models, whereas character-level methods increase the \texttt{char} models' performances the most. Except from \textit{Nonce} and a few exceptional cases, the syntactic-level augmentation techniques either deteriorate the performance or do not lead to significant gains. Even though \texttt{mBERT} baselines are quite strong, it is still possible to achieve significantly better scores---especially for Kazakh, and \textit{CA} leads to consistent improvements across experimented languages. On the other hand, we haven't observed any significant gains by any augmentation for Telugu-\texttt{char} and Tamil-\texttt{BPE}. We also note the following decreasing patterns across models: Belarusian-\textit{SR}, Telugu-\textit{SR}, Telugu-\textit{crop}, Vietnamese-\textit{crop} and Telugu-\textit{rotate}. 
  We believe the inconsistency of \textit{SR} is due to: (i) pretrained embeddings do not guarantee a syntactic equivalent of a token, and (ii) quality of embeddings also suffer from low-resources, i.e., small Wikipedia.} 

  Furthermore, the linguistically motivated syntactic techniques such as crop and \textit{rotate} are found to be less effective than the straight forward techniques that rely on introducing noise to the system, namely as \textit{RWD}, \textit{RWS}, \textit{CI}, \textit{CSU}, \textit{CSW}, \textit{CD} and \textit{CA}. One important difference between these two categories is the \textit{number of augmented data} that can be generated via both techniques. This number is limited to linguistic factors for syntactic techniques, such as the number of subtrees or the number of tokens that share the same morphological features and dependency labels. On the other hand, the number of sentences with noise addition is only constrained by the parameters. Hence, substantially larger amounts of augmented data can be generated via simple techniques than the more sophisticated ones. 

  Additionally, we believe that these additional amount of noisy sentences provide informative signals to the POS tagging network. As shown previously~\cite{TenneyDP19}, low-level features like part of speech tags, are mostly encoded at the initial layers of the network. As layers are added to the network, more sophisticated features that require understanding of the interaction between tokens are more likely to be captured. This is connected to the nature of the unsophisticated augmentation techniques that treat tokens as isolated items, rather than considering the relation amongst other tokens like the syntactic methods. As a result, easier techniques yield stronger associations at the initial layers, while more sophisticated techniques are more likely to strengthen the intermediate layers. In addition to easier techniques, the syntactic method \textit{Nonce} also advanced the scores of most POS Taggers. Although it is considered one of the sophisticated methods, it targets tokens instead of the modifying the sentence structure.

  \begin{landscape}
    \topskip0pt
    \vspace*{\fill}
    \begin{table}[htb!]
      \centering
      \setlength{\tabcolsep}{2pt}
      \scalebox{0.68}{
      
      \begin{tabular}{clllllllllllll}
        \toprule
        & & &\multicolumn{3}{c}{\textbf{Token-Level}} &\multicolumn{5}{c}{\textbf{Character-Level}} &\multicolumn{3}{c}{\textbf{Syntactic}} \\
        \midrule
        & &\textbf{Org} &\textbf{RWD} &\textbf{RWS} &\textbf{SR} &\textbf{CI} &\textbf{CSU} &\textbf{CSW} &\textbf{CD} &\textbf{CA} &\textbf{Crop} &\textbf{Rotate} &\textbf{Nonce} \\
        \toprule
       
        \multirow{7}{*}{\rotatebox[origin=c]{90}{\texttt{char}}} & \textbf{be} & 90.29 $\pm$ 1.23 & 90.41 $\pm$ 1.68 & \cellcolor{lightgreen}91.38 $\pm$ 1.36 *& \cellcolor{lightred}88.57 $\pm$ 2.48 $\dagger$& \cellcolor{darkgreen}91.00 $\pm$ 0.43 **& \cellcolor{darkgreen}91.66 $\pm$ 0.39 **& 90.24 $\pm$ 1.17 & \cellcolor{darkgreen}91.60 $\pm$ 0.79 **& \cellcolor{darkgreen}91.50 $\pm$ 0.63 **& \cellcolor{lightred}88.75 $\pm$ 2.30 $\dagger$& 90.06 $\pm$ 1.12 & 89.68 $\pm$ 1.26 \\
        
        & \textbf{bxr} & 39.29 $\pm$ 4.16 & 43.77 $\pm$ 6.37 & 41.48 $\pm$ 6.13 & \cellcolor{darkgreen}46.44 $\pm$ 3.16 **& \cellcolor{lightgreen}41.36 $\pm$ 5.60 *& 41.75 $\pm$ 5.60 & 41.15 $\pm$ 6.19 & \cellcolor{lightgreen}41.06 $\pm$ 4.83 *& 39.99 $\pm$ 5.33 & \cellcolor{lightgreen}43.96 $\pm$ 1.53 *& 42.37 $\pm$ 5.20 & \cellcolor{lightgreen}42.45 $\pm$ 3.57 *\\

        & \textbf{kk} & 46.24 $\pm$ 3.02 & \cellcolor{darkgreen}49.65 $\pm$ 2.75 **& 46.36 $\pm$ 3.47 & \cellcolor{darkgreen}51.25 $\pm$ 3.04 **& \cellcolor{darkgreen}49.94 $\pm$ 2.91 **& \cellcolor{darkgreen}50.61 $\pm$ 3.15 **& \cellcolor{darkgreen}49.51 $\pm$ 3.47 **& \cellcolor{lightgreen}48.38 $\pm$ 4.0 *& \cellcolor{darkgreen}51.49 $\pm$ 1.22 **& 42.57 $\pm$ 9.04 & 46.94 $\pm$ 2.84 & \cellcolor{darkgreen}51.45 $\pm$ 0.58 **\\

        & \textbf{ku} & 57.04 $\pm$ 3.98 & 57.70 $\pm$ 5.20 & \cellcolor{lightgreen}58.30 $\pm$ 3.25 *& \cellcolor{lightred}56.28 $\pm$ 3.71 $\dagger$& \cellcolor{lightgreen}58.90 $\pm$ 3.41 *& \cellcolor{lightgreen}57.97 $\pm$ 3.11 *& 56.99 $\pm$ 3.20 & 56.62 $\pm$ 3.40 & \cellcolor{lightgreen}58.60 $\pm$ 2.10 *& \cellcolor{lightgreen}57.67 $\pm$ 3.71 *& 54.46 $\pm$ 6.77 & 57.09 $\pm$ 2.68 \\
        
        & \textbf{ta} &  84.30 $\pm$ 1.87 & \cellcolor{lightgreen}86.96 $\pm$ 0.54 *& \cellcolor{darkgreen}86.37 $\pm$ 0.26 **& 85.94 $\pm$ 0.35 & \cellcolor{darkgreen}86.29 $\pm$ 0.40 **& \cellcolor{darkgreen}86.62 $\pm$ 0.14 **& \cellcolor{darkgreen}86.37 $\pm$ 0.57 **& \cellcolor{lightgreen}86.24 $\pm$ 0.62 *& \cellcolor{lightgreen}86.33 $\pm$ 0.77 *& 84.63 $\pm$ 0.79 & 84.36 $\pm$ 0.57 & \cellcolor{lightgreen}85.98 $\pm$ 0.80 *\\

        & \textbf{te} & 90.96 $\pm$ 1.38 & 90.07 $\pm$ 1.85 & 90.51 $\pm$ 1.21 & \cellcolor{lightred}90.80 $\pm$ 0.36 $\dagger$& 89.90 $\pm$ 1.54 & 90.21 $\pm$ 1.23 & 90.49 $\pm$ 1.68 & 90.62 $\pm$ 1.40 & 90.23 $\pm$ 1.73 & \cellcolor{darkred}89.99 $\pm$ 0.49 $\dagger\dagger$& \cellcolor{lightred}90.01 $\pm$ 0.63 $\dagger$& - \\
        
        & \textbf{vi} & 77.62 $\pm$ 0.54 & 77.45 $\pm$ 0.75 & 77.10 $\pm$ 0.90 & 77.93 $\pm$ 0.21 & 76.97 $\pm$ 0.87 & 77.88 $\pm$ 0.24 & 77.29 $\pm$ 0.25 & \cellcolor{lightred}76.00 $\pm$ 0.57 $\dagger$& 78.08 $\pm$ 0.76 & \cellcolor{lightred}77.17 $\pm$ 0.26 $\dagger$& \cellcolor{lightred}77.20 $\pm$ 0.45 $\dagger$& - \\
        
        \midrule

        \multirow{6}{*}{\rotatebox[origin=c]{90}{\texttt{BPE}}} & \textbf{be} & 88.43 $\pm$ 0.59 & \cellcolor{darkgreen}89.75 $\pm$ 0.42 **& \cellcolor{darkgreen}89.72 $\pm$ 0.38 **& \cellcolor{lightred}87.61 $\pm$ 0.80 $\dagger$& \cellcolor{darkgreen}86.06 $\pm$ 1.64 **& 88.73 $\pm$ 1.68 & \cellcolor{lightred}87.30 $\pm$ 1.59 $\dagger$& 88.41 $\pm$ 1.47 & \cellcolor{lightgreen}89.23 $\pm$ 1.53 *& 87.83 $\pm$ 0.74 & 87.81 $\pm$ 1.81 & 88.40 $\pm$ 1.66 \\

        & \textbf{bxr} & 33.11 $\pm$ 0.80 & \cellcolor{darkgreen}44.21 $\pm$ 7.06 **& \cellcolor{darkgreen}41.50 $\pm$ 7.81 **& \cellcolor{darkgreen}46.67 $\pm$ 7.55 **& \cellcolor{darkgreen}41.82 $\pm$ 7.37 **& \cellcolor{darkgreen}43.49 $\pm$ 7.72 **& \cellcolor{darkgreen}39.14 $\pm$ 5.37 **& \cellcolor{darkgreen}46.06 $\pm$ 7.43 **& \cellcolor{darkgreen}42.20 $\pm$ 6.91 **& 28.60 $\pm$ 9.77 & \cellcolor{darkgreen}40.47 $\pm$ 7.43 **& \cellcolor{darkgreen}44.21 $\pm$ 5.51 **\\

        & \textbf{kk} & 46.20 $\pm$ 0.68 & \cellcolor{darkgreen}50.36 $\pm$ 1.75 **& \cellcolor{lightgreen}46.83 $\pm$ 0.30 *& \cellcolor{darkgreen}50.29 $\pm$ 1.66 **& \cellcolor{lightgreen}48.64 $\pm$ 2.84 *& \cellcolor{darkgreen}47.14 $\pm$ 0.52 **& \cellcolor{darkgreen}48.33 $\pm$ 1.84 **& 47.28 $\pm$ 2.57 & \cellcolor{lightgreen}47.90 $\pm$ 2.40 *& 47.02 $\pm$ 6.51 & \cellcolor{darkgreen}49.00 $\pm$ 2.16 **& 50.75 $\pm$ 2.31 \\

        & \textbf{ta} &  82.76 $\pm$ 0.41 & 81.96 $\pm$ 1.00 & 82.48 $\pm$ 0.59 & 81.51 $\pm$ 1.22 & 82.77 $\pm$ 0.31 & 82.79 $\pm$ 0.35 & 82.22 $\pm$ 0.52 & 82.48 $\pm$ 0.49 & 82.26 $\pm$ 0.68 & 81.69 $\pm$ 0.56 & \cellcolor{lightred}80.97 $\pm$ 0.64 $\dagger$& 81.97 $\pm$ 1.00 \\

        & \textbf{te} & 88.27 $\pm$ 0.60 & \cellcolor{lightgreen}88.60 $\pm$ 0.23 *& \cellcolor{lightgreen}88.74 $\pm$ 0.63 *& \cellcolor{darkred}87.71 $\pm$ 0.63 $\dagger\dagger$& \cellcolor{darkgreen}88.93 $\pm$ 0.26 **& 88.21 $\pm$ 0.31 & \cellcolor{lightred}87.93 $\pm$ 0.67 $\dagger$& 88.27 $\pm$ 0.67 & 88.35 $\pm$ 0.34 & 87.38 $\pm$ 0.93 & 87.68 $\pm$ 0.36 & - \\

        & \textbf{vi} & 77.07 $\pm$ 0.46 & \cellcolor{darkgreen}77.83 $\pm$ 0.50 **& \cellcolor{darkgreen}77.77 $\pm$ 0.33 **& \cellcolor{darkgreen}77.83 $\pm$ 0.62 **& 77.42 $\pm$ 0.33 & 77.39 $\pm$ 0.54 & \cellcolor{lightgreen}77.46 $\pm$ 0.56 *& \cellcolor{lightgreen}77.53 $\pm$ 0.37 *& \cellcolor{darkgreen}77.72 $\pm$ 0.43 **& \cellcolor{darkred}75.17 $\pm$ 1.08 $\dagger\dagger$& \cellcolor{darkred}76.04 $\pm$ 0.41 $\dagger\dagger$& - \\

        \midrule
        \multirow{4}{*}{\rotatebox[origin=c]{90}{\texttt{mBERT}}} & \textbf{be} & 95.30 $\pm$ 0.65 & \cellcolor{lightred}94.50 $\pm$ 0.63 $\dagger$& \cellcolor{lightred}94.27 $\pm$ 0.68 $\dagger$& 94.55 $\pm$ 0.92 & 94.44 $\pm$ 1.39 & \cellcolor{darkgreen}95.70 $\pm$ 0.60 **& 95.18 $\pm$ 0.74 & 94.76 $\pm$ 1.16 & \cellcolor{lightgreen}95.56 $\pm$ 0.34 *& \cellcolor{lightgreen}95.64 $\pm$ 0.71 *& 95.11 $\pm$ 0.70 & 94.48 $\pm$ 1.02 \\

        & \textbf{kk} & 71.34 $\pm$ 2.28 & 70.75 $\pm$ 1.89 & \cellcolor{darkgreen}73.10 $\pm$ 1.65 **& \cellcolor{lightgreen}71.88 $\pm$ 2.16 *& \cellcolor{lightgreen}71.65 $\pm$ 1.51 *& 69.00 $\pm$ 4.01 & \cellcolor{darkgreen}73.11 $\pm$ 0.77 **& 70.90 $\pm$ 1.58 & \cellcolor{lightgreen}72.57 $\pm$ 1.92 *& 67.42 $\pm$ 2.20 & \cellcolor{lightgreen}72.00 $\pm$ 1.59 *& \cellcolor{darkgreen}72.45 $\pm$ 1.72 **\\

        & \textbf{te} & 89.92 $\pm$ 0.48 & 89.53 $\pm$ 0.95 & \cellcolor{lightgreen}90.12 $\pm$ 1.03 *& \cellcolor{lightred}88.43 $\pm$ 0.81 $\dagger$& 89.49 $\pm$ 1.17 & 88.84 $\pm$ 1.41 & 88.59 $\pm$ 0.64 & \cellcolor{lightred}89.40 $\pm$ 0.41 $\dagger$& \cellcolor{lightgreen}90.15 $\pm$ 1.31 *& \cellcolor{lightred}88.82 $\pm$ 1.01 $\dagger$& \cellcolor{lightred}89.01 $\pm$ 0.48 $\dagger$& - \\

        & \textbf{vi} & 82.41 $\pm$ 0.72 & 81.66 $\pm$ 0.97 & 82.75 $\pm$ 0.87 & 81.04 $\pm$ 1.37 & 82.45 $\pm$ 0.89 & 82.09 $\pm$ 1.07 & \cellcolor{lightgreen}82.48 $\pm$ 0.81 *& 82.20 $\pm$ 0.87 & \cellcolor{lightgreen}82.57 $\pm$ 0.65 *& 82.28 $\pm$ 0.80 & 82.33 $\pm$ 0.79 & - \\
        \bottomrule         
        \end{tabular}
        }
      \caption{ Part-of-speech tagging results on original (Org) and augmented datasets, where the results of \texttt{char} are given at the top, \texttt{BPE} in the middle and \texttt{mBERT} at the bottom.}
      \label{tab:postag_res_multi}
  \end{table}
  \vspace*{\fill}
  \end{landscape}

\subsection{Dependency Parsing}
\label{ssec:dep_res_sec}

  \secrev{Due to infeasibility of performing parameter search for all languages, we run a grid search on parameters for the languages with the lowest resources: Buryat, Kazakh and Kurmanji. For Belarusian, Tamil, Telugu and Vietnamese, we determine the best configurations for each language-method pair based on POS tagging results. To elaborate, we measure the average performance of the POS taggers for each parameter (e.g., Ratio) and value (e.g., 0.1) and fix the value to the one that provided the best average performance for the specific augmentation method. We use POS tagging as the basis of our dependency parsing experiments for the following reasons: (i) the two tasks are closely related and purely syntactic; and (ii) share the same datasets, hence easily comparable. We report the results for the best performing parameters for Buryat, Kazakh and Kurmanji; and average of multiple runs with fixed parameters for the other languages in Table~\ref{tab:depparsing_res_multi}.}{We use the transition-based parser \texttt{uuparser} and the \texttt{biaffine} parser based on mBERT from \citet{RustPVRG20} to conduct the dependency parsing experiments. The mean and the standard deviations for each language-dataset pair are given in Table~\ref{tab:depparsing_res_multi}.} 

  \begin{table*}[!ht]
    \centering
    \scalebox{0.55}{
    
    \begin{tabular}{clllllllllll}
    \toprule
    & & &\multicolumn{1}{c}{\textbf{Token-Level}} &\multicolumn{5}{c}{\textbf{Character-Level}} &\multicolumn{3}{c}{\textbf{Syntactic}} \\
    \midrule
    & &\textbf{Org} &\textbf{SR} &\textbf{CI} &\textbf{CSU} &\textbf{CSW} &\textbf{CD} &\textbf{CA} &\textbf{Crop} &\textbf{Rotate} &\textbf{Nonce} \\
    
    \toprule

    \multirow{7}{*}{\rotatebox[origin=c]{90}{\texttt{uuparser}}} & \textbf{be} & 52.85 $\pm$ 0.70 & \cellcolor{darkred}50.10 $\pm$ 0.67 $\dagger\dagger$& \cellcolor{darkgreen}54.05 $\pm$ 0.40 **& 53.39 $\pm$ 0.51 & \cellcolor{lightgreen}55.37 $\pm$ 1.48 *& \cellcolor{lightgreen}54.89 $\pm$ 0.87 *& \cellcolor{lightgreen}53.57 $\pm$ 0.97 *& 53.01 $\pm$ 1.01 & 53.03 $\pm$ 0.31 & 54.27 $\pm$ 0.68 \\

    & \textbf{bxr} & 11.73 $\pm$ 1.65 & \cellcolor{lightgreen}12.70 $\pm$ 1.84 *& 12.39 $\pm$ 1.31 & \cellcolor{darkgreen}13.04 $\pm$ 1.20 **& \cellcolor{darkgreen}13.83 $\pm$ 0.46 **& \cellcolor{darkgreen}13.44 $\pm$ 1.06 **& \cellcolor{darkgreen}14.06 $\pm$ 1.14 **& \cellcolor{darkgreen}11.99 $\pm$ 1.52 **& 11.43 $\pm$ 1.39 & \cellcolor{darkred}10.70 $\pm$ 0.77 $\dagger\dagger$\\

    & \textbf{kk} & 23.82 $\pm$ 0.92 & 23.82 $\pm$ 0.98 & 24.36 $\pm$ 0.59 & \cellcolor{darkgreen}24.80 $\pm$ 0.75 **& \cellcolor{lightgreen}24.87 $\pm$ 0.77 *& \cellcolor{lightgreen}24.79 $\pm$ 0.86 *& \cellcolor{lightgreen}24.62 $\pm$ 0.83 *& 24.09 $\pm$ 0.79 & 23.76 $\pm$ 0.32 & \cellcolor{lightgreen} 24.78 $\pm$ 1.07 *\\

    & \textbf{ku} & 21.09 $\pm$ 0.86 & \cellcolor{darkgreen}23.22 $\pm$ 0.60 **& \cellcolor{darkgreen}23.70 $\pm$ 0.92 **& \cellcolor{darkgreen}24.08 $\pm$ 0.74 **& \cellcolor{lightgreen}23.48 $\pm$ 0.87 *& \cellcolor{darkgreen}24.32 $\pm$ 0.55 **& \cellcolor{darkgreen}24.39 $\pm$ 0.12 **& \cellcolor{darkgreen}22.74 $\pm$ 0.35 **& \cellcolor{darkgreen}24.50 $\pm$ 0.44 **& \cellcolor{darkgreen}23.98 $\pm$ 0.31 **\\

    & \textbf{ta} & 55.52 $\pm$ 0.39 & 54.27 $\pm$ 1.99 & 55.39 $\pm$ 0.42 & 55.15 $\pm$ 0.71 & 54.89 $\pm$ 0.90 & 56.11 $\pm$ 0.77 & 56.52 $\pm$ 1.31 & \cellcolor{lightgreen}54.62 $\pm$ 0.65 *& \cellcolor{lightgreen}55.16 $\pm$ 0.18 *& \cellcolor{darkgreen}57.50 $\pm$ 0.43 **\\

    & \textbf{te} & 77.79 $\pm$ 0.85 & 76.94 $\pm$ 0.94 & 77.65 $\pm$ 0.94 & \cellcolor{lightgreen}78.83 $\pm$ 0.73 *& \cellcolor{lightgreen}77.27 $\pm$ 1.12 *& \cellcolor{darkgreen}78.11 $\pm$ 0.80 **& 78.87 $\pm$ 0.51 & \cellcolor{lightgreen}78.01 $\pm$ 0.86 *& \cellcolor{lightgreen}78.27 $\pm$ 1.05 *& - \\
 
    & \textbf{vi} & 55.01 $\pm$ 0.15 & \cellcolor{darkred}48.23 $\pm$ 0.27 $\dagger\dagger$& 53.80 $\pm$ 0.32 & \cellcolor{lightred}52.26 $\pm$ 0.47 $\dagger$& \cellcolor{darkred}52.19 $\pm$ 0.36 $\dagger\dagger$& \cellcolor{darkred}52.92 $\pm$ 0.23 $\dagger\dagger$& \cellcolor{lightred}53.15 $\pm$ 0.42 $\dagger$& 55.33 $\pm$ 0.40 & 54.87 $\pm$ 0.26  & - \\

    \midrule

    \multirow{7}{*}{\rotatebox[origin=c]{90}{\texttt{biaffine}}} & \textbf{be\^} & 78.17 $\pm$ 0.36 & 79.66 $\pm$ 0.28 & \cellcolor{darkgreen}80.47 $\pm$ 0.11 **& \cellcolor{darkgreen}80.04 $\pm$ 0.80 **& 78.79 $\pm$ 0.64 & \cellcolor{darkgreen}80.58 $\pm$ 0.49 **& \cellcolor{darkgreen}80.45 $\pm$ 0.36 **& \cellcolor{darkgreen}79.84 $\pm$ 0.30 **& \cellcolor{darkgreen}79.78 $\pm$ 0.35 **& \cellcolor{darkgreen}79.33 $\pm$ 0.27 **\\

    & \textbf{bxr} & 17.71 $\pm$ 0.52 & \cellcolor{darkgreen}18.73 $\pm$ 0.13 **& 17.91 $\pm$ 0.51 & 17.65 $\pm$ 0.44 & 17.73 $\pm$ 0.17 & 17.72 $\pm$ 0.87 & \cellcolor{lightgreen}17.81 $\pm$ 0.50 *& \cellcolor{lightred}16.74 $\pm$ 0.88 $\dagger$& 17.59 $\pm$ 0.36 & 17.12 $\pm$ 0.82 \\

    & \textbf{kk\^} & 34.23 $\pm$ 0.32 & \cellcolor{darkgreen}35.49 $\pm$ 0.75 **& \cellcolor{darkgreen}35.61 $\pm$ 0.52 **& \cellcolor{darkgreen}36.69 $\pm$ 0.76 **& \cellcolor{darkgreen}36.30 $\pm$0.59 **& \cellcolor{darkgreen}35.42 $\pm$ 0.49 **& \cellcolor{darkgreen}35.76 $\pm$ 0.42 **& \cellcolor{lightgreen}34.61 $\pm$ 0.75 *& 34.18 $\pm$ 0.42 & \cellcolor{darkgreen}38.43 $\pm$ 0.17 **\\

    & \textbf{ku} & 20.48 $\pm$ 0.35 & 20.64 $\pm$ 0.24 & \cellcolor{darkgreen}22.16 $\pm$ 0.47 **& 20.49 $\pm$ 0.33 & \cellcolor{darkgreen}22.56 $\pm$ 0.68 **& \cellcolor{darkgreen}22.71 $\pm$ 0.30 **& \cellcolor{darkgreen}23.13 $\pm$ 0.26 **& 20.53 $\pm$ 0.41 & \cellcolor{darkred}19.39 $\pm$ 0.50 $\dagger\dagger$& \cellcolor{darkgreen}22.82 $\pm$ 0.24 **\\

    & \textbf{ta\^} & 70.01 $\pm$ 0.62 & \cellcolor{darkgreen}70.99 $\pm$ 0.21 **& 70.27 $\pm$ 0.10 & \cellcolor{lightgreen}71.21 $\pm$ 1.08 *& 70.54 $\pm$ 0.49 & \cellcolor{darkgreen}71.13 $\pm$ 0.67 **& \cellcolor{darkgreen}71.47 $\pm$ 0.53 **& \cellcolor{darkgreen}70.89 $\pm$ 0.30 **& \cellcolor{lightgreen}71.11 $\pm$ 1.01 *& \cellcolor{darkgreen}71.24 $\pm$ 0.85 **\\

    & \textbf{te\^} & 84.60 $\pm$ 0.79 & 84.60 $\pm$ 0.80 & \cellcolor{lightred}84.28 $\pm$ 0.97 $\dagger$& \cellcolor{darkgreen}84.88 $\pm$ 0.54 **& \cellcolor{darkgreen}85.02 $\pm$ 0.32 **& \cellcolor{darkred}84.05 $\pm$ 0.78 $\dagger\dagger$& 84.74 $\pm$ 0.81 & \cellcolor{darkgreen}85.30 $\pm$ 0.57 **& \cellcolor{darkred}84.19 $\pm$ 0.59 $\dagger\dagger$& - \\

    & \textbf{vi\^} & 66.25 $\pm$ 0.84 & \cellcolor{darkred}62.40 $\pm$ 0.76 $\dagger\dagger$& \cellcolor{darkred}63.49 $\pm$ 0.80 $\dagger\dagger$& \cellcolor{darkred}63.87 $\pm$ 0.55 $\dagger\dagger$& \cellcolor{darkred}64.22 $\pm$ 2.07 $\dagger\dagger$& \cellcolor{darkred}64.54 $\pm$ 2.05 $\dagger\dagger$& \cellcolor{darkred}64.09 $\pm$ 1.25 $\dagger\dagger$& 66.11 $\pm$ 0.85 & 65.94 $\pm$ 0.91 & - \\

    \bottomrule 
    
    \end{tabular}
    }
    \caption{Dependency parsing LAS scores on original (Org) and augmented datasets, where the results of \texttt{uuparser} are given at the top, and \texttt{biaffine} parser at the bottom. \emph{language\^} denotes that \emph{language} is part of the multilingual BERT.}
    \label{tab:depparsing_res_multi}
  \end{table*} 

  \paragraph{ \texttt{uuparser}} Compared to POS tagging, the relative improvements over the baseline are substantially higher. However, unlike POS tagging there is no linear relation between low baseline scores and larger improvements. For instance, the relative improvement for Kazakh is only 3\%, despite having the second lowest baseline. This suggests that more factors such as dataset statistics and linguistic properties come into play for dependency parsing.
  \secrev{}{
  Except for Vietnamese, we observe significant gains for all languages by at least one of the augmentation methods---sometimes by \textit{any} technique as in Kurmanji. Similar to POS tagging, \textit{SR} provides a mixture of results, significantly reducing the scores for Belarus and Vietnamese, while improving the Kurmanji and Buryat parsers. Character-level augmentation techniques mostly improve the performance of Belarusian, Buryat, Kazakh, Kurmanji and Telugu parsers significantly. Unlike POS tagging, the improvements from syntactic methods are more emphasized for most languages with the exception of Vietnamese, Belarusian and Buryat-\textit{Nonce}.}  

  \secrev{}{\paragraph{\texttt{biaffine}} As discussed earlier, mBERT is trained on the top 100 languages with the largest Wikipedia, however, training is performed on a \textit{shared} vocabulary. Hence even if a language is not seen during training, the shared vocabulary might enable \textit{zero-shot} learning---especially if the model is trained with related languages. In Table~\ref{tab:depparsing_res_multi}, the languages without an * sign are not included to mBERT training, therefore the scores are the results of zero-shot learning. Compared to \texttt{uuparser}, all baselines (Org) are substantially higher as expected---except for Kurmanji. Despite such high scores, we observe a considerable amount of statistically significant improvements over the baseline with some exceptions. As with the \texttt{uuparser}, Vietnamese dependency parsing performance is significantly reduced by augmentation, suggesting that augmentation methods introduce too much noise for Vietnamese dependency parser. \textit{SR} improves the scores more consistently for Belarus, Kazakh and Tamil, while not being able to bring significant gains for the other languages. Character noise injection techniques---especially \textit{CD} and \textit{CA}---boost the performance of most parsers, again with the exception of Vietnamese and Telugu. Similar to \texttt{uuparser}, syntactic techniques result in higher scores compared to POS tagging. We observe significant gains from \textit{Nonce}, while \textit{Crop} also improves substantially in most cases---Buryat and Vietnamese are exceptions. Unlike \textit{Nonce} and \textit{Crop}, \textit{Rotate} gives mixed results. 
  }

  \paragraph{Summary and Discussion} The similarities \secrev{to POS tagging results}{among different parsers and POS taggers} are numerous, such as (i) the small number of augmentation techniques that were able to improve scores for \secrev{Telugu, Vietnamese and}{} Buryat; (ii) high performance of character noise methods \secrev{and the Nonce techniques}{} for most languages\secrev{and}{;} (iii) generally \secrev{low}{confusing} scores produced by the \textit{SR} and \textit{rotation}\secrev{}{, and (iv) mostly unimproved or worse results for Vietnamese and Telugu}. 
  Considering the synthetic nature of most languages, where characters, e.g., case markers, provide valuable information on the relationship between two tokens, high performance of character-level noise is rather expected. In other words, varying character sequences may help the network to strengthen the dependency relations. Unlike POS tagging, we also observe a performance boost from the syntactic augmenters: \textit{Crop} and \textit{Nonce}, sometimes leading to best results e.g., for Vietnamese and Kurmanji. This suggests that introducing more syntactic variation/noise, even in smaller amounts than character-level noise, helps in certain cases. Nevertheless, the performances of both techniques are comparable, and it is not possible to single out one technique that is guaranteed to improve the results. Additionally, we observe that not \textit{all} character-level methods increase the scores and there is no clear pattern to which character noise improves which language. \secrev{For instance, while \textit{CA} yields the best scores in Belarusian POS tagging, \textit{CSW} and \textit{CD} are the ones to increase the dependency parsing scores for Belarusian. Finally, for dependency parsing the largest improvements are almost always obtained by injecting character-level noise as in POS tagging; however syntactic augmentation almost always improves dependency parsing unlike POS tagging.}{Our results also suggest that a higher number of augmentation techniques are able to improve significantly over competitive baseline scores provided by \texttt{biaffine} compared to \texttt{mBERT}-based POS tagger. One reason may be that \texttt{mBERT} model already containing a substantial amount of low-level syntactic knowledge and augmentation techniques only adding noise to the fine-tuning process.} 

\subsection{Semantic Role Labeling}
\label{ssec:srl_res_sec}
  As discussed in Sec~\ref{ssec:tasks}, we \textit{simulate} a low-resource scenario by sampling 250, 500 and 1000 training sentences from the original training sets. We perform a grid search on augmentation parameters for the \#sampled=250 setting, and choose the parameters with the best average performance for other dataset settings. The results are given in Table~\ref{tab:srl_results}. 

  \begin{table*}[!htp]
    \centering
    \scalebox{0.55}{
    
      \begin{tabular}{lllllllllll}
      \toprule
      
      \multicolumn{9}{c}{\textbf{Catalan}} & \\
      \midrule
      \textbf{\#sample} &\textbf{Org} &\textbf{SR} &\textbf{CI} &\textbf{CSU} &\textbf{CSW} &\textbf{CD} &\textbf{CA} &\textbf{Crop} &\textbf{Rotate}  \\
      
      250 & 31.34 $\pm$ 0.99 & \cellcolor{darkgreen}37.29 $\pm$ 1.26 **& \cellcolor{darkgreen}38.50 $\pm$ 1.37 **& \cellcolor{darkgreen}37.74 $\pm$ 1.64 **& \cellcolor{darkgreen}39.79 $\pm$ 0.25 **& \cellcolor{darkgreen}37.45 $\pm$ 1.58 **& \cellcolor{darkgreen}38.28 $\pm$ 1.48 **& \cellcolor{darkred}26.59 $\pm$ 0.52 $\dagger\dagger$& \cellcolor{darkred}25.96 $\pm$ 0.62 $\dagger\dagger$\\

      500 & 44.53 $\pm$ 1.39 & 44.12 $\pm$ 1.78 & \cellcolor{darkgreen}49.75 $\pm$ 0.10 **& \cellcolor{lightgreen}47.77 $\pm$ 0.65 *& \cellcolor{lightgreen}46.81 $\pm$ 1.59 *& \cellcolor{lightgreen}47.47 $\pm$ 0.62 *& \cellcolor{darkgreen}49.46 $\pm$ 0.19 **& \cellcolor{lightred}44.30 $\pm$ 1.16 $\dagger$& 44.72 $\pm$ 0.36 \\
      
      1000 & 53.06 $\pm$ 1.33 & 53.49 $\pm$ 0.64 & 53.80 $\pm$ 0.67 & 54.83 $\pm$ 0.81 & \cellcolor{lightgreen}56.37 $\pm$ 0.29 *& \cellcolor{lightgreen}55.50 $\pm$ 0.17 *& 53.97 $\pm$ 0.96 & 53.57 $\pm$ 0.07 & 52.31 $\pm$ 0.96 \\
  
      \midrule
      \multicolumn{9}{c}{\textbf{Turkish}} & \\
      \midrule
      \textbf{\#sample} &\textbf{Org} &\textbf{SR} &\textbf{CI} &\textbf{CSU} &\textbf{CSW} &\textbf{CD} &\textbf{CA} &\textbf{Crop} &\textbf{Rotate}  \\
      
      250 & 31.28 $\pm$ 0.12 & 31.78 $\pm$ 1.20 & 31.62 $\pm$ 1.53 & 30.67 $\pm$ 2.19 & 30.14 $\pm$ 2.46 & \cellcolor{lightgreen}32.52 $\pm$ 0.89 *& 32.02 $\pm$ 1.53 & \cellcolor{lightgreen}32.42 $\pm$ 0.64 *& 30.37 $\pm$ 1.75 \\

      500 & 35.75 $\pm$ 1.38 & 35.95 $\pm$ 0.65 & 36.35 $\pm$ 0.68 & \cellcolor{lightgreen}38.27 $\pm$ 0.88 *& 37.30 $\pm$ 1.05 & 34.80 $\pm$ 1.96 & 36.23 $\pm$ 1.37 & \cellcolor{lightgreen}37.40 $\pm$ 1.07 *& 34.58 $\pm$ 1.89 \\

      1000 & 44.89 $\pm$ 0.80 & 42.41 $\pm$ 1.20 & 43.36 $\pm$ 0.67 & 41.78 $\pm$ 1.39 & 42.28 $\pm$ 1.31 & 41.42 $\pm$ 1.38 & \cellcolor{darkred}29.14 $\pm$ 0.47 $\dagger\dagger$& 44.83 $\pm$ 1.07 & 42.71 $\pm$ 0.76 \\
     
      \midrule
      \multicolumn{9}{c}{\textbf{Spanish}} & \\
      \midrule
      \textbf{\#sample} &\textbf{Org} &\textbf{SR} &\textbf{CI} &\textbf{CSU} &\textbf{CSW} &\textbf{CD} &\textbf{CA} &\textbf{Crop} &\textbf{Rotate}  \\
      250 & 31.23 $\pm$ 1.30 & \cellcolor{lightgreen}34.65 $\pm$ 1.02 *& 36.31 $\pm$ 2.28 & \cellcolor{darkgreen}37.91 $\pm$ 1.21 **& \cellcolor{lightgreen}36.80 $\pm$ 1.90 *& \cellcolor{lightgreen}36.76 $\pm$ 2.11 *& \cellcolor{darkgreen}37.34 $\pm$ 1.14 **& \cellcolor{darkred}26.04 $\pm$ 1.34 $\dagger\dagger$& \cellcolor{darkred}26.95 $\pm$ 2.22 $\dagger\dagger$\\

      500 & 44.16 $\pm$ 0.68 & 43.89 $\pm$ 0.65 & 44.80 $\pm$ 0.81 & 44.82 $\pm$ 1.10 & 43.75 $\pm$ 1.36 & \cellcolor{lightgreen}44.60 $\pm$ 0.72 *& \cellcolor{lightgreen}45.03 $\pm$ 0.91 *& \cellcolor{darkred}42.35 $\pm$ 0.42 $\dagger\dagger$& \cellcolor{darkred}41.13 $\pm$ 0.38 $\dagger\dagger$\\

      1000 & 50.98 $\pm$ 1.30 & 50.80 $\pm$ 1.21 & \cellcolor{darkgreen}53.03 $\pm$ 0.76 **& \cellcolor{darkgreen}52.53 $\pm$ 0.60 **& \cellcolor{lightgreen}52.10 $\pm$ 1.00 *& \cellcolor{darkgreen}52.76 $\pm$ 0.57 **& \cellcolor{darkgreen}54.12 $\pm$ 0.12 **& 51.32 $\pm$ 1.09 & \cellcolor{lightgreen}52.40 $\pm$ 0.34 *\\
     
      \midrule
      \multicolumn{9}{c}{\textbf{Czech}} & \\
      \midrule
      \textbf{\#sample} &\textbf{Org} &\textbf{SR} &\textbf{CI} &\textbf{CSU} &\textbf{CSW} &\textbf{CD} &\textbf{CA} &\textbf{Crop} &\textbf{Rotate}  \\
      
      250 & 35.63 $\pm$ 1.08 & 33.48 $\pm$ 1.58 & \cellcolor{lightgreen}37.17 $\pm$ 0.13 *& \cellcolor{lightgreen}37.56 $\pm$ 0.02 *& \cellcolor{lightred}31.56 $\pm$ 1.00 $\dagger$& 35.70 $\pm$ 0.80 & 36.14 $\pm$ 1.78 & \cellcolor{darkgreen}34.84 $\pm$ 1.55 **& \cellcolor{darkred}30.78 $\pm$ 0.38 $\dagger\dagger$\\

      500 & 44.96 $\pm$ 0.51 & \cellcolor{lightred}42.04 $\pm$ 1.08 $\dagger$& \cellcolor{lightgreen}46.44 $\pm$ 1.20 *& 46.13 $\pm$ 1.97 & 46.24 $\pm$ 1.60 & \cellcolor{darkgreen}47.92 $\pm$ 0.59 **& \cellcolor{lightgreen}47.42 $\pm$ 1.07 *& \cellcolor{lightgreen}46.38 $\pm$ 0.12 *& 44.83 $\pm$ 1.47 \\
      
      1000 & 50.29 $\pm$ 0.09 & \cellcolor{lightred}48.15 $\pm$ 0.31 $\dagger$& \cellcolor{lightred}48.66 $\pm$ 0.49 $\dagger$& \cellcolor{lightgreen}52.63 $\pm$ 0.97 *& 51.10 $\pm$ 1.23 & \cellcolor{lightred}47.85 $\pm$ 1.19 $\dagger$& \cellcolor{lightgreen}51.97 $\pm$ 0.54 *& 49.33 $\pm$ 1.36 & 48.45 $\pm$ 1.34 \\
     
      \midrule
      \multicolumn{9}{c}{\textbf{Finnish}} & \\
      \midrule
      \textbf{\#sample} &\textbf{Org} &\textbf{SR} &\textbf{CI} &\textbf{CSU} &\textbf{CSW} &\textbf{CD} &\textbf{CA} &\textbf{Crop} &\textbf{Rotate} \\
      
      250 & 18.80 $\pm$ 0.89 & 18.60 $\pm$ 2.34 & 17.18 $\pm$ 1.56 & 19.68 $\pm$ 1.16 & \cellcolor{darkgreen}25.71 $\pm$ 0.80 **& \cellcolor{darkgreen}28.87 $\pm$ 1.55 **& \cellcolor{darkgreen}27.96 $\pm$ 0.44 **& \cellcolor{darkgreen}30.83 $\pm$ 0.17 **& 20.07 $\pm$ 2.42 \\

      500 & 37.23 $\pm$ 0.35 & 34.10 $\pm$ 1.61 & \cellcolor{lightred}35.59 $\pm$ 0.87 $\dagger$& \cellcolor{lightred}36.88 $\pm$ 0.29 $\dagger$& \cellcolor{lightgreen}38.98 $\pm$ 0.03 *& \cellcolor{lightred}35.29 $\pm$ 0.99 $\dagger$& 35.32 $\pm$ 1.63 & \cellcolor{lightgreen}37.58 $\pm$ 0.93 *& 35.16 $\pm$ 1.08 \\

      1000 & 41.64 $\pm$ 0.98 & \cellcolor{lightred}40.15 $\pm$ 0.95 $\dagger$& 41.27 $\pm$ 0.84 & 41.41 $\pm$ 1.08 & 39.80 $\pm$ 0.81 & 40.57 $\pm$ 1.17 & 41.99 $\pm$ 0.04 & 41.35 $\pm$ 1.37 & \cellcolor{lightred}39.08 $\pm$ 0.34 $\dagger$\\
     
      \bottomrule
      \end{tabular}
    }
  \caption{ Semantic role labeling results on original (Org) and augmented datasets.}
  \label{tab:srl_results}
  \end{table*}

  As expectedly, the relative improvement over the baselines decreases, as the number of samples increases. For some languages like Finnish, the drop is dramatic, while for the majority, the decrease is exponential. The only exception is Czech. The reason is the fine-grained, language specific semantic annotation scheme that requires larger datasets. 
  Another noticeable pattern is the decreasing number of augmentation methods that improve the F1 scores with the increasing sample size. \secrev{For instance, while six of the methods increase the SRL performance in \#sample=250 setting for Spanish, only 2 provides improvement for \#sample=2000 setting. Similarly in Turkish and Finnish, SRL benefits almost from all augmentation methods for 250 samples, while only a couple of methods continue to provide gains in case of 500 samples.}{For instance, while six of the methods increase the SRL performance in \#sample=250 setting for Catalan, only two of them provides improvement for \#sample=1000 setting---which are also less significant.}

  We see one distinctive pattern for the languages Turkish, Czech and Finnish. Unlike Spanish and Catalan, the syntactic operation \textit{crop} improves the performances for almost all settings. We believe this is due to the rich case marking systems of these languages that enable generating almost natural sentences from subtrees. Furthermore, we observe that the \textit{rotation} operation introduces a high amount of noise that can not be regularized easily with the models. One reason for more consistent improvements with cropping operation is related the semantic role statistics. Most treebanks in this study are dominated by the core arguments, i.e., Arg0, Arg1. These core arguments are usually observed as subjects or objects. In addition, many predicates are encountered with missing arguments, i.e., it is more likely to see a predicate with only one of the arguments than containing all. We believe cropping introduces more variation in terms of core arguments compared to rotation, which provides more signal for the cases such as missing arguments. For Spanish and Catalan, the gains are almost always provided by character-level noise. \secrev{The gain is consistent in almost all numbers of samples.}{} On the contrary, both languages never benefit from the syntactic operations, as expected.  

\secrev{}{
\subsection{Summary of the Findings}
\label{ssec:summary_findings}
  We summarize the key findings from experiments from different perspectives: languages, downstream tasks, augmentation techniques and models.  
  \subsubsection{Languages}
    \begin{itemize}
      \item Most languages have seen significant improvements in at least one augmentation configuration independent from tasks, models and the dataset sizes.  
      \item Vietnamese has mostly witnessed drops in scores which were especially highlighted in dependency parsing. This suggests that the augmentation methods may be less effective for analytic languages. 
      \item We have not observed any significant difference between fusional and agglutinative languages by means of POS and DEP scores, however the differences between syntactic and non-syntactic augmentation techniques were pronounced for languages with different morphological properties.    
      \item The suitability of augmentation techniques have been found to be dependent on the language and subword unit pair. For instance Tamil POS tagging with \texttt{BPE} baseline could not be improved, where we have observed significant improvements over the Tamil POS tagging \texttt{char} baseline.   
      \item We have detected inconsistent results for the Telugu language for majority of the cases. Furthermore, we have not seen many configurations that led to substantial improvements. Telugu had the second largest number of significant declines in performance after Vietnamese. We believe this may be due to Telugu having one of the largest treebanks, and augmentation techniques adding unmeaningful noise to the already strong baseline.  
    \end{itemize}
  \subsubsection{Tasks}
    \begin{itemize}
        \item We have found many similarities between the results for POS and DEP. The most important ones are: (i) character-level augmentations providing significant improvements in most cases and (ii) inconsistent results from \textit{SR} and \textit{rotation} methods.
        \item The task that has been improved the most \textit{(i.e., statistically significant improvements with a large gap over the baseline)} was DEP, which has been followed by POS and SRL. In other words, the experimented augmentation benefited the task with the \textit{intermediate} complexity the most. 
        \item Strong baseline scores provided by exploiting large pretrained contextualized embeddings were more likely to be further improved for DEP (e.g., \texttt{biaffine}) than POS (e.g., \texttt{mBERT}). 
    \end{itemize}
  \subsubsection{Augmentation Techniques}
    \begin{itemize}
        \item The most consistent augmenters across tasks, models and languages have been found to be the character-level ones. 
        \item A satisfactory choice of augmentation techniques depends heavily on the input unit. For instance, token-level augmentation provide significant improvements for \texttt{BPE}, while character-level ones give higher scores for \texttt{char} and Word Piece of \texttt{mBERT}. 
        \item The performance of \textit{SR} has been detected as irregular. The reason may be that, it relies on external pretrained embeddings that may be of lower-quality for some of the languages. 
        \item Even if we have not found one single winner across character-level augmentation methods, the mixed character noise, a.k.a., \textit{CA}, has improved the POS, DEP and SRL tasks more consistently.  
        \item Among the syntactic augmenters, \textit{crop} and \textit{Nonce} have been found to be more reliable compared to \textit{rotate}---with some exceptions like the case for \texttt{BPE}. 
    \end{itemize}
  \subsubsection{Models}
      \begin{itemize}
        \item We have observed almost a regular improvement pattern among different models for DEP, such as a significant drop in scores for Vietnamese. Even there were some similarities among models for POS, such as syntactic augmenters achieving the lowest scores, a regular pattern was not visible. The reason might be the difference among the subword units in POS\footnote{All POS models use \textit{distinct} input types: character, BPE and WordPiece; while \texttt{uuparser} (using a combination of char and word) and \texttt{biaffine} (WordPiece) parsers are likely to share more vocabulary.}.
        \item Strong baseline scores provided by exploiting large pretrained contextualized embeddings were more likely to be further improved for DEP (e.g., \texttt{biaffine}) than POS (e.g., \texttt{mBERT}). This may be due to \texttt{mBERT} containing a substantial amount of low-level syntactic knowledge \textit{already} (e.g., POS), hence augmentation techniques only adding noise to the fine-tuning process.
    \end{itemize}
  }    

\section{Analysis}
\label{sec:analysis}
\secrev{}{In this section, we first discuss the parameter choice for augmentation techniques and conduct a case study on POS tagging to analyze their sensitivity to parameter choice and their performance range. Next, we study the performance of individual augmentation techniques on frequent and infrequent tokens and lexical types to reveal whether the performance improvement also corresponds to better generalization in out of domain settings.}

\subsection{Augmentation Parameters}
  
\begin{figure*}
  \centering
        \begin{subfigure}[b]{0.45\textwidth}
         \centering
         \includegraphics[width=1.\linewidth]{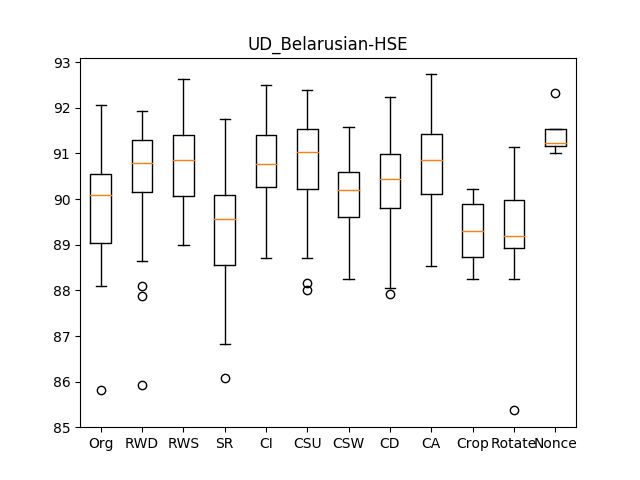}  
        \end{subfigure}
        ~
        \begin{subfigure}[b]{0.45\textwidth}
         \centering
         \includegraphics[width=1.\linewidth]{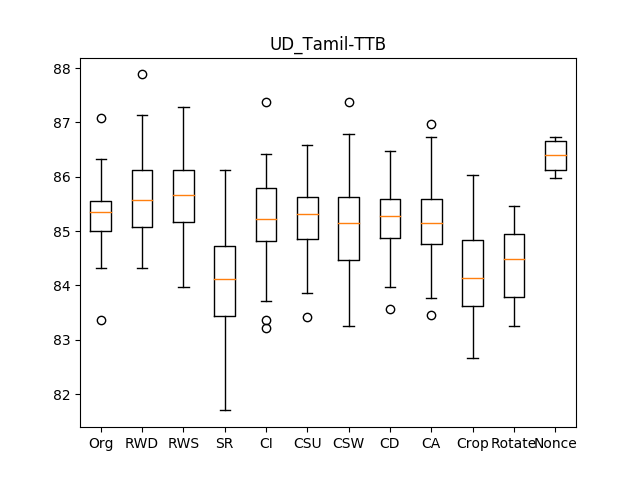}  
        \end{subfigure}

        \begin{subfigure}[b]{0.45\textwidth}
         \centering
         \includegraphics[width=1.\linewidth]{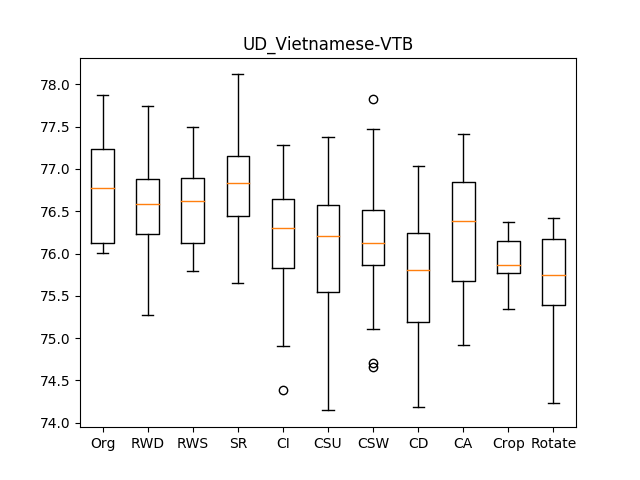}  
        \end{subfigure}
        ~
        \begin{subfigure}[b]{0.45\textwidth}
         \centering
         \includegraphics[width=1.\linewidth]{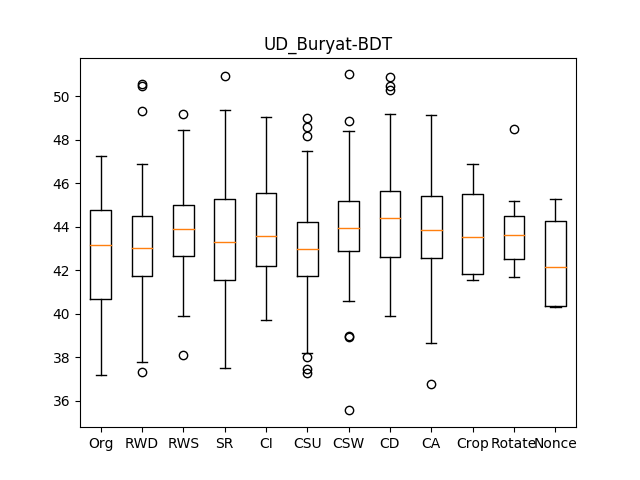}  
        \end{subfigure}

        \begin{subfigure}[b]{0.45\textwidth}
         \centering
         \includegraphics[width=1.\linewidth]{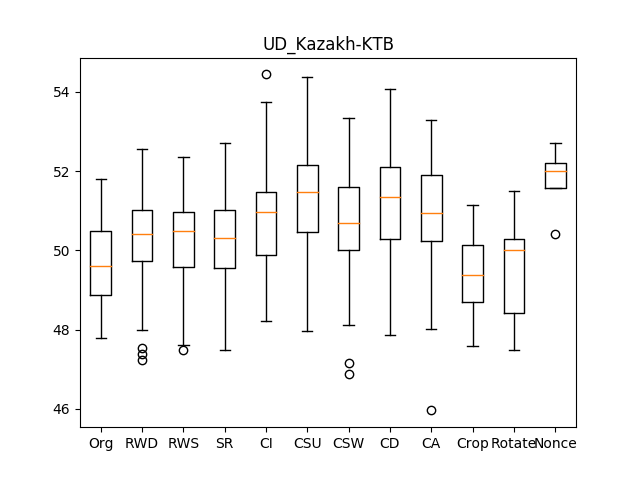}  
        \end{subfigure}
        ~
        \begin{subfigure}[b]{0.45\textwidth}
         \centering
         \includegraphics[width=1.\linewidth]{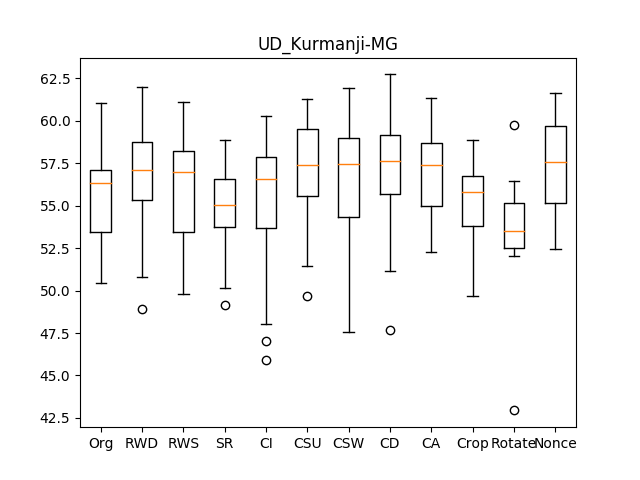}  
        \end{subfigure}

      \caption{ Box plots for augmentation model performances on POS Tagging. Org box refers to the baseline on the original dataset, while other techniques are defined in Sec.~\ref{sec:augmain}.}  
  \label{fig:box_plot_pos}
  \end{figure*}

  Augmentation methods are parameterized and their performance may be sensitive to changes in parameter values. However in a real-world low-case scenario (mostly) without any development set, parameter tuning may not always be possible. Therefore we create augmented datasets using all combinations of augmentation parameters described in Sec.~\ref{sec:augmain} (e.g., $4x4x2=32$ augmented datasets with RWD method for each treebank). To analyze the behavior of each augmentation technique, we perform a grid search on the parameters 
  and draw the box plots for each augmentation method on Belarusian, Tamil, Vietnamese, Buryat, Kazakh and Kurmanji POS tagging shown in Fig.~\ref{fig:box_plot_pos}. The box plots ensure that the augmentation techniques can be compared with respect to their performance range and their sensitivity to parameters rather than their best performance.

  \paragraph{Belarusian} The median lines of RWD, RWS, CI, CSU, CA and the Nonce lie outside of the Org (baseline) box. This suggests that these techniques are statistically likely to perform higher than the baseline. Out of these methods, Nonce and RWD performances are found to be less dispersed than the others, i.e., are more prone to parameter changes. The min-max range of Nonce is quite smaller than others, signaling the reliability of the method 
  There are only a few outliers to most techniques, suggesting that the results mostly follow a distribution. 

  \paragraph{Tamil} For Tamil only RWS and Nonce median lines lie above the baseline box. Similar to Belarusian, the box length and the min-max range of Nonce is smaller than RWS; hinting the reliability of the model. Unlike Belarusian, SR, crop and rotate lie under the box, meaning that they are likely to yield worse results than the baseline. Similarly the number of outliers are limited and the maximum improvement over the best baseline model is around $0.7\%$. 

  \paragraph{Vietnamese, Hungarian} For Vietnamese, there are no methods that are significantly better than the baseline, but only worse: CD, crop and rotate. Although the maximum value of SR surpasses the best baseline model, the box plots reveal that this is statistically unlikely. Similar pattern is observed for Hungarian, despite being from a different language family. 

  \paragraph{Buryat, Telugu} While the training dataset of Buryat is the smallest of all, even modest changes in parameters may lead to outliers. Interestingly, we found none of the techniques to be significantly better or worse than the baseline according to the median lines. However, the outliers provide a performance boost around $8\%$ over the best baseline. Even the plot is not shown for convenience, we noticed that all augmentation techniques, except from SR, provide neither a significant drop or increase in the scores for Telugu, similar to Buryat. 


  \paragraph{Kazakh, Kurmanji} For both languages, the median lines above the baseline box are the character-level methods and the syntactic method Nonce. In Kazakh, similar to Belarusian and Tamil, Nonce has the smallest box and the min-max range. For Kurmanji, however, CA has the smallest box and the min-max range instead of Nonce. Kazakh and Kurmanji, as the languages with the lowest resources, benefit significantly more from character-level augmentation compared to other languages. This may be due to higher ratio of out-of-vocabulary words for these treebanks. In other words, association of a character set to POS labels is more important than association of tokens with POS labels. Apparently, character-level noise helps to strengthen such links.

\subsection{Frequency Analysis}
\begin{figure*}
      \centering
          \begin{subfigure}[b]{0.45\textwidth}
           \centering
           \includegraphics[width=1.\linewidth]{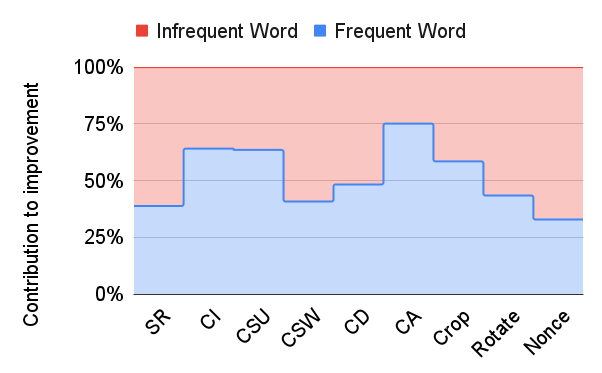}  
          \end{subfigure}
          ~
          \begin{subfigure}[b]{0.45\textwidth}
           \centering
           \includegraphics[width=1.\linewidth]{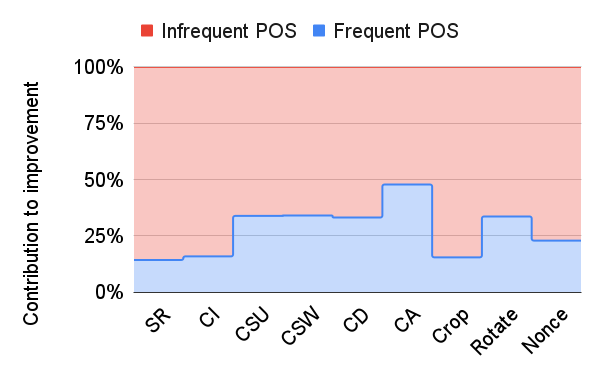}  
          \end{subfigure}
        \caption{ Individual contributions of correct labeling of frequent and infrequent tokens and POS tags to the overall parser performance. Shown separately for each augmentation technique.}  
    \label{fig:freq}
  \end{figure*}

  \secrev{}{Besides improving the downstream task scores, one of the important goals of an augmentation method is to increase generalization capability of the model. In order to evaluate the extent of this skill, we measure the individual performances on frequent and infrequent tokens along with word types. We perform a case study on dependency parsing since it stands between POS and SRL by means of complexity. The results are given in Fig.~\ref{fig:freq}.}
  
  \secrev{}{\paragraph{Frequent vs Infrequent Tokens} First, we define the token as \textit{frequent} if it is among the top 10\% in the token frequency list extracted from the combination of training and development set. Next, we create a vocabulary of frequent tokens and then measure the LAS scores individually for frequent and infrequent tokens for each language and augmentation technique. Finally, we average the scores over all languages to ease the presentation of the results. The results show that only \textit{SR}, \textit{CSW}, \textit{Rotate} and \textit{Nonce} improve the scores of infrequent tokens than frequent ones. \textit{SR} and \textit{Nonce} are likely to replace frequent tokens with infrequent ones, since their objective is to replace words with another. However, \textit{CSW} and \textit{Rotate} are not designed to replace tokens. We believe character switching sometimes coincidentally resulted with rare tokens, and \textit{Rotate} operation has randomly chosen the subtrees with rare tokens to augment. Interestingly, there is no one-to-one correspondence between the techniques that improved the dependency scores the most and the techniques that improved labeling of infrequent tokens the most. This may be due to building the vocabulary over tiny training sets and identifying the frequent/rare tokens accordingly. Therefore we analyze a more general property: POS tags.} 

 \paragraph{Frequent vs Infrequent POS tags} \secrev{}{We perform a similar analysis for the token class---using the gold POS tags as the class. Since the number of unique POS tags are much lower than unique tokens, we identify the top 50\% of the POS tags as frequent. We use the same calculation technique as above. The results show that all techniques improve the performance on rare token classes more than the frequent ones, however there still does not exist a direct correlation between the best performing technique and the improvement over infrequent POS tags. This suggests that the improvement cannot simply be explained by frequency analysis, i.e., the method that focuses on improving rare tokens or token classes is not guaranteed to improve the overall score. This is because the parser's performance is a more complicated multivariate variable that rely on many other factors apparent in dataset statistics.}    

\section{Conclusion}
   Neural models have emerged as the standard model for a wide range of NLP tasks including part-of-speech tagging, dependency parsing and semantic role labeling---the task of assigning semantic role labels to predicate-argument pairs. These models were shown to provide state-of-the-art scores in the presence on large training datasets, however still fall behind traditional statistical techniques in genuinely low-resource scenarios. One method that is commonly used to overcome the low dataset size problem is enhancing the original dataset synthetically, which is referred to as data augmentation. Recently, variety of text augmentation techniques have been proposed such as replacing words with their synonyms or related words, injecting spelling errors, generating grammatically correct but meaningless sentences, simplifying sentences and many more. 

   Despite the richness of augmentation literature, previous works mostly explore text classification and machine translation tasks on high-resource languages, and report single scores. In this study, on the contrary, we provide a more detailed performance analysis of the existing augmentation techniques on a diverse set of languages and traditional sequence tagging tasks which are more sensitive to noise. First, we compile a rich set of augmentation methods of different categories that can be applied to our tasks and languages, namely as character-level, token-level and syntactic-level. Then we systematically compare them on the following downstream tasks: part-of-speech tagging, dependency parsing and semantic role labeling. We conduct experiments on truly low-resource languages (when possible) such as Buryat, Kazakh and Kurmanji; and simulate a low-resource setting on a diverse set of languages such as Catalan, Turkish, Finnish and Czech when data is not available.  

   We find that the easy-to-apply techniques such as injecting character-level noise or generating grammatical, meaningless sentences provide gains across languages more consistently for majority of the cases. Compared to POS tagging, we have observed \secrev{larger relative improvements}{more significant improvements} over the baseline for dependency parsing. Although the augmentation patterns for dependency parsing are found similar to POS tagging, a few differences, such as larger gains from syntactic methods, are noted. Again the largest improvements in dependency parsers are mostly obtained by injecting character-level noises. For SRL, we show that the improvement from augmentation decreases with more training samples, along with the number of augmentation techniques that increases the scores. We observe that languages with fusional morphology, almost always benefit from character-level noise the most; but always suffer from the syntactic operations \textit{crop} and \textit{rotate}. On the contrary, agglutinative languages such as Turkish and Finnish benefit from cropping while character-level augmentation may provide inconsistent improvements. \secrev{}{We show that augmentation techniques can provide consistent and statistically significant improvements for all languages except from Vietnamese and Telugu. We find that the improvements do not solely depend on the task architecture, i.e., augmentation methods can further improve on the strong \texttt{biaffine} parser as well as the weaker \texttt{uuparser}.}

\section{Acknowledgements}

We first thank to anonymous reviewers that helped us improve the paper. We would like to thank Clara Vania, Benjamin Heinzerling, Jonas Pfeiffer and Phillip Rust for the valuable discussions and providing early access to their implementations. We finally thank Celal Şahin, Necla İşgüder and Osman İşgüder for their invaluable support during the writing of this paper. 

\starttwocolumn
\bibliography{main}

\begin{thebibliography}{69}
\expandafter\ifx\csname natexlab\endcsname\relax\def\natexlab#1{#1}\fi

\bibitem[{Anaby{-}Tavor et~al.(2020)Anaby{-}Tavor, Carmeli, Goldbraich, Kantor,
  Kour, Shlomov, Tepper, and Zwerdling}]{anaby2020not}
Anaby{-}Tavor, Ateret, Boaz Carmeli, Esther Goldbraich, Amir Kantor, George
  Kour, Segev Shlomov, Naama Tepper, and Naama Zwerdling. 2020.
\newblock Do not have enough data? deep learning to the rescue!
\newblock In \emph{The Thirty-Fourth {AAAI} Conference on Artificial
  Intelligence, {AAAI} 2020, The Thirty-Second Innovative Applications of
  Artificial Intelligence Conference, {IAAI} 2020, The Tenth {AAAI} Symposium
  on Educational Advances in Artificial Intelligence, {EAAI} 2020, New York,
  NY, USA, February 7-12, 2020}, pages 7383--7390.

\bibitem[{Andreas(2020)}]{andreas-2020-good}
Andreas, Jacob. 2020.
\newblock Good-enough compositional data augmentation.
\newblock In \emph{Proceedings of the 58th Annual Meeting of the Association
  for Computational Linguistics}, pages 7556--7566, Association for
  Computational Linguistics, Online.

\bibitem[{Belinkov and Bisk(2018)}]{BelinkovB18}
Belinkov, Yonatan and Yonatan Bisk. 2018.
\newblock Synthetic and natural noise both break neural machine translation.
\newblock In \emph{6th International Conference on Learning Representations,
  {ICLR} 2018, Vancouver, BC, Canada, April 30 - May 3, 2018, Conference Track
  Proceedings}, pages 1--13.

\bibitem[{Chen, Ji, and Evans(2020)}]{chen-etal-2020-finding}
Chen, Hannah, Yangfeng Ji, and David Evans. 2020.
\newblock Finding friends and flipping frenemies: Automatic paraphrase dataset
  augmentation using graph theory.
\newblock In \emph{Findings of the Association for Computational Linguistics:
  EMNLP 2020}, pages 4741--4751, Association for Computational Linguistics,
  Online.

\bibitem[{Chen et~al.(2021)Chen, Tam, Raffel, Bansal, and Yang}]{ChenJE20}
Chen, Jiaao, Derek Tam, Colin Raffel, Mohit Bansal, and Diyi Yang. 2021.
\newblock An empirical survey of data augmentation for limited data learning in
  {NLP}.
\newblock \emph{CoRR}, abs/2106.07499:1--19.

\bibitem[{Chen et~al.(2020)Chen, Wang, Tian, Yang, and Yang}]{ChenWTYY20}
Chen, Jiaao, Zhenghui Wang, Ran Tian, Zichao Yang, and Diyi Yang. 2020.
\newblock Local additivity based data augmentation for semi-supervised {NER}.
\newblock In \emph{Proceedings of the 2020 Conference on Empirical Methods in
  Natural Language Processing, {EMNLP} 2020, Online, November 16-20, 2020},
  pages 1241--1251.

\bibitem[{Chen, Yang, and Yang(2020)}]{chen2020mixtext}
Chen, Jiaao, Zichao Yang, and Diyi Yang. 2020.
\newblock {M}ix{T}ext: Linguistically-informed interpolation of hidden space
  for semi-supervised text classification.
\newblock In \emph{Proceedings of the 58th Annual Meeting of the Association
  for Computational Linguistics}, pages 2147--2157, Association for
  Computational Linguistics, Online.

\bibitem[{Devlin et~al.(2019)Devlin, Chang, Lee, and Toutanova}]{DevlinCLT19}
Devlin, Jacob, Ming{-}Wei Chang, Kenton Lee, and Kristina Toutanova. 2019.
\newblock {BERT:} pre-training of deep bidirectional transformers for language
  understanding.
\newblock In \emph{Proceedings of the 2019 Conference of the North American
  Chapter of the Association for Computational Linguistics: Human Language
  Technologies, {NAACL-HLT} 2019, Minneapolis, MN, USA, June 2-7, 2019, Volume
  1 (Long and Short Papers)}, pages 4171--4186.

\bibitem[{Ding et~al.(2020)Ding, Liu, Bing, Kruengkrai, Nguyen, Joty, Si, and
  Miao}]{ding-etal-2020-daga}
Ding, Bosheng, Linlin Liu, Lidong Bing, Canasai Kruengkrai, Thien~Hai Nguyen,
  Shafiq Joty, Luo Si, and Chunyan Miao. 2020.
\newblock {DAGA}: Data augmentation with a generation approach
  for{L}ow-resource tagging tasks.
\newblock In \emph{Proceedings of the 2020 Conference on Empirical Methods in
  Natural Language Processing (EMNLP)}, pages 6045--6057, Association for
  Computational Linguistics, Online.

\bibitem[{Fadaee, Bisazza, and Monz(2017)}]{FadaeeBM17a}
Fadaee, Marzieh, Arianna Bisazza, and Christof Monz. 2017.
\newblock Data augmentation for low-resource neural machine translation.
\newblock In \emph{Proceedings of the 55th Annual Meeting of the Association
  for Computational Linguistics, {ACL} 2017, Vancouver, Canada, July 30 -
  August 4, Volume 2: Short Papers}, pages 567--573.

\bibitem[{Feng et~al.(2020)Feng, Gangal, Kang, Mitamura, and
  Hovy}]{feng2020genaug}
Feng, Steven~Y., Varun Gangal, Dongyeop Kang, Teruko Mitamura, and Eduard Hovy.
  2020.
\newblock {G}en{A}ug: Data augmentation for finetuning text generators.
\newblock In \emph{Proceedings of Deep Learning Inside Out (DeeLIO): The First
  Workshop on Knowledge Extraction and Integration for Deep Learning
  Architectures}, pages 29--42, Association for Computational Linguistics,
  Online.

\bibitem[{Feng et~al.(2021)Feng, Gangal, Wei, Chandar, Vosoughi, Mitamura, and
  Hovy}]{FengGWCVMH21}
Feng, Steven~Y., Varun Gangal, Jason Wei, Sarath Chandar, Soroush Vosoughi,
  Teruko Mitamura, and Eduard~H. Hovy. 2021.
\newblock A survey of data augmentation approaches for {NLP}.
\newblock In \emph{Findings of the Association for Computational Linguistics:
  {ACL/IJCNLP} 2021, Online Event, August 1-6, 2021}, pages 968--988.

\bibitem[{Feng, Li, and Hoey(2019)}]{feng-etal-2019-keep}
Feng, Steven~Y., Aaron~W. Li, and Jesse Hoey. 2019.
\newblock Keep calm and switch on! {P}reserving sentiment and fluency in
  semantic text exchange.
\newblock In \emph{Proceedings of the 2019 Conference on Empirical Methods in
  Natural Language Processing and the 9th International Joint Conference on
  Natural Language Processing (EMNLP-IJCNLP)}, pages 2701--2711, Association
  for Computational Linguistics, Hong Kong, China.

\bibitem[{Futrell, Mahowald, and Gibson(2015)}]{wordOrder}
Futrell, Richard, Kyle Mahowald, and Edward Gibson. 2015.
\newblock Quantifying word order freedom in dependency corpora.
\newblock In \emph{Proceedings of the Third International Conference on
  Dependency Linguistics (Depling 2015)}, pages 91--100, Uppsala University,
  Uppsala, Sweden.

\bibitem[{Gao et~al.(2019)Gao, Zhu, Wu, Xia, Qin, Cheng, Zhou, and
  Liu}]{gao-etal-2019-soft}
Gao, Fei, Jinhua Zhu, Lijun Wu, Yingce Xia, Tao Qin, Xueqi Cheng, Wengang Zhou,
  and Tie-Yan Liu. 2019.
\newblock Soft contextual data augmentation for neural machine translation.
\newblock In \emph{Proceedings of the 57th Annual Meeting of the Association
  for Computational Linguistics}, pages 5539--5544, Association for
  Computational Linguistics, Florence, Italy.

\bibitem[{Glavas and Vulic(2021)}]{GlavasV21}
Glavas, Goran and Ivan Vulic. 2021.
\newblock Is supervised syntactic parsing beneficial for language understanding
  tasks? an empirical investigation.
\newblock In \emph{Proceedings of the 16th Conference of the European Chapter
  of the Association for Computational Linguistics: Main Volume, {EACL} 2021,
  Online, April 19 - 23, 2021}, pages 3090--3104.

\bibitem[{Grave et~al.(2018)Grave, Bojanowski, Gupta, Joulin, and
  Mikolov}]{grave2018learning}
Grave, Edouard, Piotr Bojanowski, Prakhar Gupta, Armand Joulin, and Tomas
  Mikolov. 2018.
\newblock Learning word vectors for 157 languages.
\newblock In \emph{Proceedings of the International Conference on Language
  Resources and Evaluation (LREC 2018)}, pages 3483--3487.

\bibitem[{Grundkiewicz, Junczys-Dowmunt, and
  Heafield(2019)}]{grundkiewicz-etal-2019-neural}
Grundkiewicz, Roman, Marcin Junczys-Dowmunt, and Kenneth Heafield. 2019.
\newblock Neural grammatical error correction systems with unsupervised
  pre-training on synthetic data.
\newblock In \emph{Proceedings of the Fourteenth Workshop on Innovative Use of
  NLP for Building Educational Applications}, pages 252--263, Association for
  Computational Linguistics, Florence, Italy.

\bibitem[{Gulordava et~al.(2018)Gulordava, Bojanowski, Grave, Linzen, and
  Baroni}]{GulordavaBGLB18}
Gulordava, Kristina, Piotr Bojanowski, Edouard Grave, Tal Linzen, and Marco
  Baroni. 2018.
\newblock Colorless green recurrent networks dream hierarchically.
\newblock In \emph{Proceedings of the 2018 Conference of the North American
  Chapter of the Association for Computational Linguistics: Human Language
  Technologies, {NAACL-HLT} 2018, New Orleans, Louisiana, USA, June 1-6, 2018,
  Volume 1 (Long Papers)}, pages 1195--1205.

\bibitem[{Guo, Kim, and Rush(2020)}]{GuoKR20}
Guo, Demi, Yoon Kim, and Alexander~M. Rush. 2020.
\newblock Sequence-level mixed sample data augmentation.
\newblock In \emph{Proceedings of the 2020 Conference on Empirical Methods in
  Natural Language Processing, {EMNLP} 2020, Online, November 16-20, 2020},
  pages 5547--5552.

\bibitem[{Guo(2020)}]{Guo20}
Guo, Hongyu. 2020.
\newblock Nonlinear mixup: Out-of-manifold data augmentation for text
  classification.
\newblock In \emph{The Thirty-Fourth {AAAI} Conference on Artificial
  Intelligence, {AAAI} 2020, The Thirty-Second Innovative Applications of
  Artificial Intelligence Conference, {IAAI} 2020, The Tenth {AAAI} Symposium
  on Educational Advances in Artificial Intelligence, {EAAI} 2020, New York,
  NY, USA, February 7-12, 2020}, pages 4044--4051, {AAAI} Press.

\bibitem[{Guo, Mao, and
  Zhang(2019{\natexlab{a}})}]{DBLP:journals/corr/abs-1905-08941}
Guo, Hongyu, Yongyi Mao, and Richong Zhang. 2019{\natexlab{a}}.
\newblock Augmenting data with mixup for sentence classification: An empirical
  study.
\newblock \emph{CoRR}, abs/1905.08941:1--7.

\bibitem[{Guo, Mao, and Zhang(2019{\natexlab{b}})}]{GuoMZ19}
Guo, Hongyu, Yongyi Mao, and Richong Zhang. 2019{\natexlab{b}}.
\newblock Mixup as locally linear out-of-manifold regularization.
\newblock In \emph{The Thirty-Third {AAAI} Conference on Artificial
  Intelligence, {AAAI} 2019, The Thirty-First Innovative Applications of
  Artificial Intelligence Conference, {IAAI} 2019, The Ninth {AAAI} Symposium
  on Educational Advances in Artificial Intelligence, {EAAI} 2019, Honolulu,
  Hawaii, USA, January 27 - February 1, 2019}, pages 3714--3722, {AAAI} Press.

\bibitem[{Han et~al.(2020)Han, Zhang, Jiang, and Tu}]{HanZJT20}
Han, Wenjuan, Liwen Zhang, Yong Jiang, and Kewei Tu. 2020.
\newblock Adversarial attack and defense of structured prediction models.
\newblock In \emph{Proceedings of the 2020 Conference on Empirical Methods in
  Natural Language Processing, {EMNLP} 2020, Online, November 16-20, 2020},
  pages 2327--2338.

\bibitem[{Haverinen et~al.(2015)Haverinen, Kanerva, Kohonen, Missila, Ojala,
  Viljanen, Laippala, and Ginter}]{Haverinen2015}
Haverinen, Katri, Jenna Kanerva, Samuel Kohonen, Anna Missila, Stina Ojala,
  Timo Viljanen, Veronika Laippala, and Filip Ginter. 2015.
\newblock {The Finnish Proposition Bank}.
\newblock \emph{Language Resources and Evaluation}, 49(4):907--926.

\bibitem[{Hedderich et~al.(2021)Hedderich, Lange, Adel, Str{\"{o}}tgen, and
  Klakow}]{HedderichLASK21}
Hedderich, Michael~A., Lukas Lange, Heike Adel, Jannik Str{\"{o}}tgen, and
  Dietrich Klakow. 2021.
\newblock A survey on recent approaches for natural language processing in
  low-resource scenarios.
\newblock In \emph{Proceedings of the 2021 Conference of the North American
  Chapter of the Association for Computational Linguistics: Human Language
  Technologies, {NAACL-HLT} 2021, Online, June 6-11, 2021}, pages 2545--2568.

\bibitem[{Heinzerling and Strube(2018)}]{Heinzerling018}
Heinzerling, Benjamin and Michael Strube. 2018.
\newblock Bpemb: Tokenization-free pre-trained subword embeddings in 275
  languages.
\newblock In \emph{Proceedings of the Eleventh International Conference on
  Language Resources and Evaluation, {LREC} 2018, Miyazaki, Japan, May 7-12,
  2018}, pages 2989--2993.

\bibitem[{Heinzerling and Strube(2019)}]{HeinzerlingS19multi}
Heinzerling, Benjamin and Michael Strube. 2019.
\newblock Sequence tagging with contextual and non-contextual subword
  representations: {A} multilingual evaluation.
\newblock In \emph{Proceedings of the 57th Conference of the Association for
  Computational Linguistics, {ACL} 2019, Florence, Italy, July 28- August 2,
  2019, Volume 1: Long Papers}, pages 273--291.

\bibitem[{Jindal et~al.(2020{\natexlab{a}})Jindal, Ghosh~Chowdhury, Didolkar,
  Jin, Sawhney, and Shah}]{jindal-etal-2020-augmenting}
Jindal, Amit, Arijit Ghosh~Chowdhury, Aniket Didolkar, Di~Jin, Ramit Sawhney,
  and Rajiv~Ratn Shah. 2020{\natexlab{a}}.
\newblock Augmenting {NLP} models using latent feature interpolations.
\newblock In \emph{Proceedings of the 28th International Conference on
  Computational Linguistics}, pages 6931--6936, International Committee on
  Computational Linguistics, Barcelona, Spain (Online).

\bibitem[{Jindal et~al.(2020{\natexlab{b}})Jindal, Ranganatha, Didolkar,
  Chowdhury, Jin, Sawhney, and Shah}]{jindalspeechmix}
Jindal, Amit, Narayanan~Elavathur Ranganatha, Aniket Didolkar, Arijit~Ghosh
  Chowdhury, Di~Jin, Ramit Sawhney, and Rajiv~Ratn Shah. 2020{\natexlab{b}}.
\newblock Speechmix - augmenting deep sound recognition using hidden space
  interpolations.
\newblock In \emph{INTERSPEECH}, pages 861--865.

\bibitem[{Karpukhin et~al.(2019)Karpukhin, Levy, Eisenstein, and
  Ghazvininejad}]{KarpukhinLEG19}
Karpukhin, Vladimir, Omer Levy, Jacob Eisenstein, and Marjan Ghazvininejad.
  2019.
\newblock Training on synthetic noise improves robustness to natural noise in
  machine translation.
\newblock In \emph{Proceedings of the 5th Workshop on Noisy User-generated
  Text, W-NUT@EMNLP 2019, Hong Kong, China, November 4, 2019}, pages 42--47.

\bibitem[{Kiperwasser and Goldberg(2016)}]{KiperwasserG16}
Kiperwasser, Eliyahu and Yoav Goldberg. 2016.
\newblock Simple and accurate dependency parsing using bidirectional {LSTM}
  feature representations.
\newblock \emph{{TACL}}, 4:313--327.

\bibitem[{Kobayashi(2018)}]{Kobayashi18}
Kobayashi, Sosuke. 2018.
\newblock Contextual augmentation: Data augmentation by words with paradigmatic
  relations.
\newblock In \emph{Proceedings of the 2018 Conference of the North {A}merican
  Chapter of the Association for Computational Linguistics: Human Language
  Technologies, Volume 2 (Short Papers)}, pages 452--457, Association for
  Computational Linguistics, New Orleans, Louisiana.

\bibitem[{Kolomiyets, Bethard, and Moens(2011)}]{KolomiyetsBM11}
Kolomiyets, Oleksandr, Steven Bethard, and Marie{-}Francine Moens. 2011.
\newblock Model-portability experiments for textual temporal analysis.
\newblock In \emph{The 49th Annual Meeting of the Association for Computational
  Linguistics: Human Language Technologies, Proceedings of the Conference,
  19-24 June, 2011, Portland, Oregon, {USA} - Short Papers}, pages 271--276.

\bibitem[{Kumar, Choudhary, and Cho(2020)}]{kumar2020}
Kumar, Varun, Ashutosh Choudhary, and Eunah Cho. 2020.
\newblock Data augmentation using pre-trained transformer models.
\newblock In \emph{Proceedings of the 2nd Workshop on Life-long Learning for
  Spoken Language Systems}, pages 18--26.

\bibitem[{de~Lhoneux, Stymne, and Nivre(2017)}]{delhoneux17arc}
de~Lhoneux, Miryam, Sara Stymne, and Joakim Nivre. 2017.
\newblock Arc-hybrid non-projective dependency parsing with a static-dynamic
  oracle.
\newblock In \emph{Proceedings of the 15th International Conference on Parsing
  Technologies, {IWPT} 2017, Pisa, Italy, September 20-22, 2017}, pages
  99--104.

\bibitem[{Ling et~al.(2015)Ling, Dyer, Black, Trancoso, Fermandez, Amir,
  Marujo, and Lu{\'{\i}}s}]{LingDBTFAML15}
Ling, Wang, Chris Dyer, Alan~W. Black, Isabel Trancoso, Ramon Fermandez, Silvio
  Amir, Lu{\'{\i}}s Marujo, and Tiago Lu{\'{\i}}s. 2015.
\newblock Finding function in form: Compositional character models for open
  vocabulary word representation.
\newblock In \emph{Proceedings of the 2015 Conference on Empirical Methods in
  Natural Language Processing, {EMNLP} 2015, Lisbon, Portugal, September 17-21,
  2015}, pages 1520--1530.

\bibitem[{Louvan and Magnini(2020)}]{louvan2020simple}
Louvan, Samuel and Bernardo Magnini. 2020.
\newblock Simple is better! lightweight data augmentation for low resource slot
  filling and intent classification.
\newblock In \emph{Proceedings of the 34th Pacific Asia Conference on Language,
  Information and Computation}, pages 167--177, Association for Computational
  Linguistics, Hanoi, Vietnam.

\bibitem[{Martha, Dan, and Paul(2005)}]{martha2005proposition}
Martha, Palmer, Gildea Dan, and Kingsbury Paul. 2005.
\newblock The proposition bank: a corpus annotated with semantic roles.
\newblock \emph{Computational Linguistics Journal}, 31(1):10--1162.

\bibitem[{Nguyen et~al.(2020)Nguyen, Joty, Wu, and Aw}]{nguyen2020data}
Nguyen, Xuan-Phi, Shafiq Joty, Kui Wu, and Ai~Ti Aw. 2020.
\newblock Data diversification: A simple strategy for neural machine
  translation.
\newblock In \emph{Advances in Neural Information Processing Systems},
  volume~33, pages 10018--10029, Curran Associates, Inc.

\bibitem[{Pennington, Socher, and Manning(2014)}]{PenningtonSM14}
Pennington, Jeffrey, Richard Socher, and Christopher~D. Manning. 2014.
\newblock Glove: Global vectors for word representation.
\newblock In \emph{Proceedings of the 2014 Conference on Empirical Methods in
  Natural Language Processing, {EMNLP} 2014, October 25-29, 2014, Doha, Qatar,
  {A} meeting of SIGDAT, a Special Interest Group of the {ACL}}, pages
  1532--1543.

\bibitem[{Rosa and Marecek(2018)}]{RosaM18}
Rosa, Rudolf and David Marecek. 2018.
\newblock {CUNI} x-ling: Parsing under-resourced languages in conll 2018 {UD}
  shared task.
\newblock In \emph{Proceedings of the CoNLL 2018 Shared Task: Multilingual
  Parsing from Raw Text to Universal Dependencies, Brussels, Belgium, October
  31 - November 1, 2018}, pages 187--196.

\bibitem[{Rust et~al.(2021)Rust, Pfeiffer, Vulic, Ruder, and
  Gurevych}]{RustPVRG20}
Rust, Phillip, Jonas Pfeiffer, Ivan Vulic, Sebastian Ruder, and Iryna Gurevych.
  2021.
\newblock How good is your tokenizer? on the monolingual performance of
  multilingual language models.
\newblock In \emph{Proceedings of the 59th Annual Meeting of the Association
  for Computational Linguistics and the 11th International Joint Conference on
  Natural Language Processing, {ACL/IJCNLP} 2021, (Volume 1: Long Papers),
  Virtual Event, August 1-6, 2021}, pages 3118--3135.

\bibitem[{{\c{S}}ahin(2016)}]{csahin2016verb}
{\c{S}}ahin, G{\"o}zde~G{\"u}l. 2016.
\newblock Verb sense annotation for turkish propbank via crowdsourcing.
\newblock In \emph{International Conference on Intelligent Text Processing and
  Computational Linguistics}, pages 496--506, Springer.

\bibitem[{Sahin and Adali(2018)}]{csahinannotation}
Sahin, G{\"{o}}zde~G{\"{u}}l and Esref Adali. 2018.
\newblock Annotation of semantic roles for the turkish proposition bank.
\newblock \emph{Language Resources and Evaluation}, 52(3):673--706.

\bibitem[{{\c{S}}ahin and Steedman(2018)}]{SahinS18}
{\c{S}}ahin, G{\"o}zde~G{\"u}l and Mark Steedman. 2018.
\newblock Data augmentation via dependency tree morphing for low-resource
  languages.
\newblock In \emph{Proceedings of the 2018 Conference on Empirical Methods in
  Natural Language Processing}, pages 5004--5009, Association for Computational
  Linguistics, Brussels, Belgium.

\bibitem[{Sennrich, Haddow, and
  Birch(2016{\natexlab{a}})}]{sennrich-etal-2016-improving}
Sennrich, Rico, Barry Haddow, and Alexandra Birch. 2016{\natexlab{a}}.
\newblock Improving {N}eural {M}achine {T}ranslation {M}odels with
  {M}onolingual {D}ata.
\newblock In \emph{Proceedings of the 54th Annual Meeting of the Association
  for Computational Linguistics (Volume 1: Long Papers)}, pages 86--96,
  Association for Computational Linguistics, Berlin, Germany.

\bibitem[{Sennrich, Haddow, and Birch(2016{\natexlab{b}})}]{SennrichHB16a}
Sennrich, Rico, Barry Haddow, and Alexandra Birch. 2016{\natexlab{b}}.
\newblock Neural machine translation of rare words with subword units.
\newblock In \emph{Proceedings of the 54th Annual Meeting of the Association
  for Computational Linguistics (Volume 1: Long Papers)}, pages 1715--1725,
  Association for Computational Linguistics, Berlin, Germany.

\bibitem[{Singh et~al.(2019)Singh, McCann, Keskar, Xiong, and
  Socher}]{singh2019xlda}
Singh, Jasdeep, Bryan McCann, Nitish~Shirish Keskar, Caiming Xiong, and Richard
  Socher. 2019.
\newblock Xlda: Cross-lingual data augmentation for natural language inference
  and question answering.
\newblock \emph{arXiv preprint arXiv:1905.11471}, pages 1--10.

\bibitem[{Sulubacak and Eryi{\u{g}}it(2018)}]{sulubacak2018implementing}
Sulubacak, Umut and G{\"u}l{\c{s}}en Eryi{\u{g}}it. 2018.
\newblock Implementing universal dependency, morphology, and multiword
  expression annotation standards for turkish language processing.
\newblock \emph{Turkish Journal of Electrical Engineering \& Computer
  Sciences}, 26(3):1662--1672.

\bibitem[{Sulubacak, Eryi{\u{g}}it, and Pamay(2016)}]{sulubacak2016imst}
Sulubacak, Umut, G{\"u}l{\c{s}}en Eryi{\u{g}}it, and Tu{\u{g}}ba Pamay. 2016.
\newblock Imst: A revisited turkish dependency treebank.
\newblock In \emph{1st International Conference on Turkic Computational
  Linguistics}, pages 1--6, EGE UNIVERSITY PRESS.

\bibitem[{Tenney, Das, and Pavlick(2019)}]{TenneyDP19}
Tenney, Ian, Dipanjan Das, and Ellie Pavlick. 2019.
\newblock {BERT} rediscovers the classical {NLP} pipeline.
\newblock In \emph{Proceedings of the 57th Conference of the Association for
  Computational Linguistics, {ACL} 2019, Florence, Italy, July 28- August 2,
  2019, Volume 1: Long Papers}, pages 4593--4601.

\bibitem[{Vaibhav et~al.(2019)Vaibhav, Singh, Stewart, and
  Neubig}]{vaibhav-etal-2019-improving}
Vaibhav, Vaibhav, Sumeet Singh, Craig Stewart, and Graham Neubig. 2019.
\newblock Improving {R}obustness of {M}achine {T}ranslation with {S}ynthetic
  {N}oise.
\newblock In \emph{Proceedings of the 2019 Conference of the North {A}merican
  Chapter of the Association for Computational Linguistics: Human Language
  Technologies, Volume 1 (Long and Short Papers)}, pages 1916--1920,
  Association for Computational Linguistics, Minneapolis, Minnesota.

\bibitem[{Vania et~al.(2019)Vania, Kementchedjhieva, S{\o}gaard, and
  Lopez}]{vaniaKSL19}
Vania, Clara, Yova Kementchedjhieva, Anders S{\o}gaard, and Adam Lopez. 2019.
\newblock A systematic comparison of methods for low-resource dependency
  parsing on genuinely low-resource languages.
\newblock In \emph{Proceedings of the 2019 Conference on Empirical Methods in
  Natural Language Processing and the 9th International Joint Conference on
  Natural Language Processing, {EMNLP-IJCNLP} 2019, Hong Kong, China, November
  3-7, 2019}, pages 1105--1116.

\bibitem[{Vaswani et~al.(2017)Vaswani, Shazeer, Parmar, Uszkoreit, Jones,
  Gomez, Kaiser, and Polosukhin}]{VaswaniSPUJGKP17}
Vaswani, Ashish, Noam Shazeer, Niki Parmar, Jakob Uszkoreit, Llion Jones,
  Aidan~N. Gomez, Lukasz Kaiser, and Illia Polosukhin. 2017.
\newblock Attention is all you need.
\newblock In \emph{Advances in Neural Information Processing Systems 30: Annual
  Conference on Neural Information Processing Systems 2017, December 4-9, 2017,
  Long Beach, CA, {USA}}, pages 5998--6008.

\bibitem[{Vickrey and Koller(2008)}]{VickreyK08}
Vickrey, David and Daphne Koller. 2008.
\newblock Sentence simplification for semantic role labeling.
\newblock In \emph{{ACL} 2008, Proceedings of the 46th Annual Meeting of the
  Association for Computational Linguistics, June 15-20, 2008, Columbus, Ohio,
  {USA}}, pages 344--352.

\bibitem[{Wang and Yang(2015)}]{wang-yang-2015-thats}
Wang, William~Yang and Diyi Yang. 2015.
\newblock That{'}s so annoying!!!: A lexical and frame-semantic embedding based
  data augmentation approach to automatic categorization of annoying behaviors
  using {\#}petpeeve tweets.
\newblock In \emph{Proceedings of the 2015 Conference on Empirical Methods in
  Natural Language Processing}, pages 2557--2563, Association for Computational
  Linguistics, Lisbon, Portugal.

\bibitem[{Wang et~al.(2018)Wang, Pham, Dai, and Neubig}]{wang2018switchout}
Wang, Xinyi, Hieu Pham, Zihang Dai, and Graham Neubig. 2018.
\newblock {S}witch{O}ut: an {E}fficient {D}ata {A}ugmentation {A}lgorithm for
  {N}eural {M}achine {T}ranslation.
\newblock In \emph{Proceedings of the 2018 Conference on Empirical Methods in
  Natural Language Processing}, pages 856--861, Association for Computational
  Linguistics, Brussels, Belgium.

\bibitem[{Wei and Zou(2019)}]{wei-zou-2019-eda}
Wei, Jason and Kai Zou. 2019.
\newblock {EDA}: Easy data augmentation techniques for boosting performance on
  text classification tasks.
\newblock In \emph{Proceedings of the 2019 Conference on Empirical Methods in
  Natural Language Processing and the 9th International Joint Conference on
  Natural Language Processing (EMNLP-IJCNLP)}, pages 6382--6388, Association
  for Computational Linguistics, Hong Kong, China.

\bibitem[{Wieting and Gimpel(2017)}]{wieting-gimpel-2017-revisiting}
Wieting, John and Kevin Gimpel. 2017.
\newblock Revisiting {R}ecurrent {N}etworks for {P}araphrastic {S}entence
  {E}mbeddings.
\newblock In \emph{Proceedings of the 55th Annual Meeting of the Association
  for Computational Linguistics (Volume 1: Long Papers)}, pages 2078--2088,
  Association for Computational Linguistics, Vancouver, Canada.

\bibitem[{Wu et~al.(2019)Wu, Lv, Zang, Han, and Hu}]{WuLZHH19}
Wu, Xing, Shangwen Lv, Liangjun Zang, Jizhong Han, and Songlin Hu. 2019.
\newblock Conditional {BERT} contextual augmentation.
\newblock In \emph{Computational Science - {ICCS} 2019 - 19th International
  Conference, Faro, Portugal, June 12-14, 2019, Proceedings, Part {IV}}, pages
  84--95.

\bibitem[{Yoo et~al.(2021)Yoo, Park, Kang, Lee, and Park}]{abs-2104-08826}
Yoo, Kang~Min, Dongju Park, Jaewook Kang, Sang{-}Woo Lee, and Woomyeong Park.
  2021.
\newblock Gpt3mix: Leveraging large-scale language models for text
  augmentation.
\newblock \emph{CoRR}, abs/2104.08826:1--11.

\bibitem[{Zeman et~al.(2018)Zeman, Hajic, Popel, Potthast, Straka, Ginter,
  Nivre, and Petrov}]{ZemanHPPSGNP18}
Zeman, Daniel, Jan Hajic, Martin Popel, Martin Potthast, Milan Straka, Filip
  Ginter, Joakim Nivre, and Slav Petrov. 2018.
\newblock Conll 2018 shared task: Multilingual parsing from raw text to
  universal dependencies.
\newblock In \emph{Proceedings of the CoNLL 2018 Shared Task: Multilingual
  Parsing from Raw Text to Universal Dependencies, Brussels, Belgium, October
  31 - November 1, 2018}, pages 1--21.

\bibitem[{Zeman, Nivre, and Abrams(2020)}]{ud26}
Zeman, Daniel, Joakim Nivre, and Mitchell et.~al. Abrams. 2020.
\newblock Universal dependencies 2.6.
\newblock {LINDAT}/{CLARIAH}-{CZ} digital library at the Institute of Formal
  and Applied Linguistics ({{\'U}FAL}), Faculty of Mathematics and Physics,
  Charles University.

\bibitem[{Zhang et~al.(2018)Zhang, Ciss{\'{e}}, Dauphin, and
  Lopez{-}Paz}]{ZhangCDL18}
Zhang, Hongyi, Moustapha Ciss{\'{e}}, Yann~N. Dauphin, and David Lopez{-}Paz.
  2018.
\newblock Mixup: Beyond empirical risk minimization.
\newblock pages 1--13.

\bibitem[{Zhang, Yu, and Zhang(2020)}]{zhang-etal-2020-seqmix}
Zhang, Rongzhi, Yue Yu, and Chao Zhang. 2020.
\newblock {S}eq{M}ix: {A}ugmenting {A}ctive {S}equence {L}abeling via
  {S}equence {M}ixup.
\newblock In \emph{Proceedings of the 2020 Conference on Empirical Methods in
  Natural Language Processing (EMNLP)}, pages 8566--8579, Association for
  Computational Linguistics, Online.

\bibitem[{Zhang et~al.(2020)Zhang, Kishore, Wu, Weinberger, and
  Artzi}]{ZhangKWWA20}
Zhang, Tianyi, Varsha Kishore, Felix Wu, Kilian~Q. Weinberger, and Yoav Artzi.
  2020.
\newblock Bertscore: Evaluating text generation with {BERT}.
\newblock In \emph{8th International Conference on Learning Representations,
  {ICLR} 2020, Addis Ababa, Ethiopia, April 26-30, 2020}, pages 1--43.

\bibitem[{Zhang, Zhao, and LeCun(2015)}]{ZhangL15}
Zhang, Xiang, Junbo~Jake Zhao, and Yann LeCun. 2015.
\newblock Character-level convolutional networks for text classification.
\newblock In \emph{Advances in Neural Information Processing Systems 28: Annual
  Conference on Neural Information Processing Systems 2015, December 7-12,
  2015, Montreal, Quebec, Canada}, pages 649--657.

\bibitem[{Zheng et~al.(2020)Zheng, Zeng, Zhou, Hsieh, Cheng, and
  Huang}]{ZhengZZHCH20}
Zheng, Xiaoqing, Jiehang Zeng, Yi~Zhou, Cho{-}Jui Hsieh, Minhao Cheng, and
  Xuanjing Huang. 2020.
\newblock Evaluating and enhancing the robustness of neural network-based
  dependency parsing models with adversarial examples.
\newblock In \emph{Proceedings of the 58th Annual Meeting of the Association
  for Computational Linguistics, {ACL} 2020, Online, July 5-10, 2020}, pages
  6600--6610.

\end{thebibliography}

\clearpage

\section{Appendix}

\appendixsection{SRL Preprocessing}
\paragraph{Turkish:} We use \texttt{Modifier}, \texttt{Subject} and \texttt{Object} dependency labels as LOI and merge the predicate tokens linked with \texttt{MWE} (Multi Word Expression), and \texttt{Deriv} (Derivation) while performing augmentation. In order to comply with the requirements of the SRL model, we first merge the inflectional groups of the words that are split via derivational boundaries; and then use character sequences of the full word in the SRL model.
\paragraph{Finnish:} This dataset uses Universal Dependency (UD) formalism, hence we use the same LOIs as in the original study~\cite{SahinS18} for augmentation. These are namely \texttt{nsubj}, \texttt{dobj}, \texttt{iobj}, \texttt{obj}, \texttt{obl} and \texttt{nmod} for LOIs and \texttt{case}, \texttt{fixed}, \texttt{flat}, \texttt{cop} and \texttt{compound} for multi-word predicates. In addition, to standardize the input format to our SRL model, the custom semantic layer annotation used in Finnish PropBank, has been converted to the same CoNLL-09 format. 
\paragraph{Spanish, Catalan:} Both datasets use the same dependency annotation scheme: \texttt{suj} for subject, \texttt{cd} for direct and \texttt{ci} for indirect object relations. Furthermore, we shorten the organization names to abbreviations \textit{(e.g., Confederación\_Francesa to CF)} since such long sequences cause memory problems for the SRL model. 
\paragraph{Czech:}\texttt{Sb}, \texttt{Obj} and \texttt{Atr} dependency labels are used as LOI and the predicate tokens with \texttt{Pred} dependency are merged. 
        


\appendixsection{UAS Scores}
The UAS scores for dependency parsing experiments are given in Tables ~\ref{tab:depparsing_uas} and \ref{tab:depparsing_uas_no_color}. 

  \begin{table*}[!ht]
    \centering
    \scalebox{0.55}{
    
    \begin{tabular}{clllllllllll}
    \toprule
    & & &\multicolumn{1}{c}{\textbf{Token-Level}} &\multicolumn{5}{c}{\textbf{Character-Level}} &\multicolumn{3}{c}{\textbf{Syntactic}} \\
    \midrule
    & &\textbf{Org} &\textbf{SR} &\textbf{CI} &\textbf{CSU} &\textbf{CSW} &\textbf{CD} &\textbf{CA} &\textbf{Crop} &\textbf{Rotate} &\textbf{Nonce} \\
    
    \toprule

    \multirow{7}{*}{\rotatebox[origin=c]{90}{\texttt{uuparser}}} & \textbf{be} & 61.78 $\pm$ 0.33 & \cellcolor{darkred}56.79 $\pm$ 1.18 $\dagger\dagger$& \cellcolor{darkgreen}63.29 $\pm$ 0.39 **& 61.64 $\pm$ 0.84 & \cellcolor{lightgreen}62.53 $\pm$ 1.24 *& 61.00 $\pm$ 1.02 & 61.62 $\pm$ 0.59 & 61.13 $\pm$ 0.46 & 61.84 $\pm$ 2.64 & 62.92 $\pm$ 1.39 \\

    & \textbf{bxr} & 28.14 $\pm$ 3.80 & \cellcolor{lightgreen}29.04 $\pm$ 3.79 *& 29.46 $\pm$ 3.46 & \cellcolor{darkgreen}30.25 $\pm$ 2.79 **& \cellcolor{darkgreen}29.93 $\pm$ 0.90 **& \cellcolor{darkgreen}30.24 $\pm$ 2.78 **& \cellcolor{darkgreen}31.59 $\pm$ 1.97 **& \cellcolor{darkgreen}29.80 $\pm$ 3.08 **& 27.82 $\pm$ 3.92 & \cellcolor{darkred}27.07 $\pm$ 1.90 $\dagger\dagger$\\

    & \textbf{kk} & 44.96 $\pm$ 1.51 & 45.09 $\pm$ 1.80 & 45.34 $\pm$ 0.87 & \cellcolor{darkgreen}46.61 $\pm$ 1.82 **& \cellcolor{lightgreen}45.60 $\pm$ 1.28 *& \cellcolor{lightgreen}45.62 $\pm$ 1.51 *& \cellcolor{lightgreen}45.20 $\pm$ 1.49 *& 44.45 $\pm$ 1.02 & 43.01 $\pm$ 1.48 & 44.75 $\pm$ 1.62 \\

    & \textbf{ku} & 32.71 $\pm$ 1.10 & \cellcolor{darkgreen}35.02 $\pm$ 1.06 **& \cellcolor{darkgreen}35.85 $\pm$ 0.22 **& \cellcolor{darkgreen}35.98 $\pm$ 0.47 **& \cellcolor{lightgreen}35.08 $\pm$ 1.08 *& \cellcolor{darkgreen}37.59 $\pm$ 0.50 **& \cellcolor{darkgreen}36.54 $\pm$ 0.24 **& \cellcolor{darkgreen}34.35 $\pm$ 0.35 **& \cellcolor{darkgreen}38.01 $\pm$ 0.54 **& \cellcolor{darkgreen}37.49 $\pm$ 0.11 **\\

    & \textbf{ta} & 64.01 $\pm$ 2.02 & 63.21 $\pm$ 2.13 & 64.04 $\pm$ 2.34 & 63.74 $\pm$ 0.86 & 63.00 $\pm$ 1.61 & 64.62 $\pm$ 1.71 & 63.78 $\pm$ 1.12 & 64.07 $\pm$ 1.12 & \cellcolor{lightgreen}65.11 $\pm$ 0.28 *& \cellcolor{lightgreen}65.69 $\pm$ 0.49 *\\

    & \textbf{te} & 87.80 $\pm$ 0.67 & 88.19 $\pm$ 0.62 & 87.75 $\pm$ 0.99 & 88.33 $\pm$ 0.98 & 88.10 $\pm$ 0.77 & \cellcolor{lightgreen}88.66 $\pm$ 0.73 *& 88.26 $\pm$ 0.64 & 88.06 $\pm$ 0.89 & 88.09 $\pm$ 0.94 & - \\
 
    & \textbf{vi} & 64.89 $\pm$ 0.33 & \cellcolor{darkred}59.10 $\pm$ 0.54 $\dagger\dagger$& \cellcolor{lightred}62.79 $\pm$ 0.49 $\dagger$& \cellcolor{lightred}62.79 $\pm$ 0.22 $\dagger$& \cellcolor{darkred}62.17 $\pm$ 0.31 $\dagger\dagger$& \cellcolor{darkred}61.87 $\pm$ 0.53 $\dagger\dagger$& \cellcolor{darkred}61.91 $\pm$ 0.81 $\dagger\dagger$& 65.70 $\pm$ 0.98 & 64.84 $\pm$ 0.76  & - \\

    \midrule

    \multirow{7}{*}{\rotatebox[origin=c]{90}{\texttt{biaffine}}} & \textbf{be\^} & 82.81 $\pm$ 0.45 & \cellcolor{lightgreen}85.03 $\pm$ 0.92 *& \cellcolor{darkgreen}85.29 $\pm$ 0.39 **& \cellcolor{lightgreen}84.69 $\pm$ 1.12 *& 84.01 $\pm$ 0.65 & \cellcolor{darkgreen}85.07 $\pm$ 0.72 **& \cellcolor{darkgreen}85.74 $\pm$ 0.39 **& \cellcolor{darkgreen}84.25 $\pm$ 0.75 **& 83.72 $\pm$ 1.10 & 83.76 $\pm$ 0.45 \\

    & \textbf{bxr} & 29.67 $\pm$ 0.66 & \cellcolor{darkgreen}31.24 $\pm$ 0.42 **& 30.81 $\pm$ 0.28 & 30.45 $\pm$ 0.33 & 29.89 $\pm$ 0.46 & 30.03 $\pm$ 0.93 & \cellcolor{lightgreen}30.53 $\pm$ 0.68 *& \cellcolor{lightred}29.33 $\pm$ 0.42 $\dagger$& 29.85 $\pm$ 0.71 & 29.16 $\pm$ 0.82 \\

    & \textbf{kk\^} & 50.85 $\pm$ 0.51 & \cellcolor{darkgreen}52.50 $\pm$ 0.67 **& \cellcolor{darkgreen}52.80 $\pm$ 0.56 **& \cellcolor{darkgreen}54.30 $\pm$ 0.83 **& \cellcolor{darkgreen}54.21 $\pm$0.38 **& \cellcolor{darkgreen}53.27 $\pm$ 0.67 **& \cellcolor{darkgreen}53.29 $\pm$ 0.35 **& 50.44 $\pm$ 0.35 & \cellcolor{lightred}48.92 $\pm$ 0.52 $\dagger$& \cellcolor{darkgreen}54.38 $\pm$ 0.75 **\\

    & \textbf{ku} & 30.48 $\pm$ 0.48 & 30.23 $\pm$ 0.15 & \cellcolor{darkgreen} 32.27 $\pm$ 0.59 **& 30.32 $\pm$ 0.25 & \cellcolor{darkgreen}32.11 $\pm$ 0.56 **& \cellcolor{darkgreen}32.55 $\pm$ 0.46 **& \cellcolor{darkgreen}33.46 $\pm$ 0.66 **& 30.33 $\pm$ 0.70 & \cellcolor{darkred}28.78 $\pm$ 0.52 $\dagger\dagger$& \cellcolor{darkgreen}33.08 $\pm$ 0.37 **\\

    & \textbf{ta\^} & 77.39 $\pm$ 0.43 & \cellcolor{darkgreen}79.11 $\pm$ 0.47 **& 77.81 $\pm$ 0.26 & \cellcolor{lightgreen}78.65 $\pm$ 1.18 *& 77.86 $\pm$ 0.60 & \cellcolor{darkgreen}78.77 $\pm$ 0.63 **& \cellcolor{darkgreen}78.85 $\pm$ 0.36 **& \cellcolor{darkgreen}78.23 $\pm$ 0.48 **& \cellcolor{lightgreen}78.45 $\pm$ 0.91 *& 77.88 $\pm$ 0.85 \\

    & \textbf{te\^} & 91.57 $\pm$ 0.57 & 91.82 $\pm$ 0.82 & \cellcolor{lightred}91.35 $\pm$ 0.78 $\dagger$& 91.96 $\pm$ 0.43 & \cellcolor{lightgreen}92.09 $\pm$ 0.62 *&91.40 $\pm$ 0.40 & \cellcolor{lightgreen}92.23 $\pm$ 0.81 *& \cellcolor{darkgreen}92.51 $\pm$ 0.32 **& 91.75 $\pm$ 0.21 & - \\

    & \textbf{vi\^} & 74.06 $\pm$ 0.21 & \cellcolor{darkred}69.24 $\pm$ 0.79 $\dagger\dagger$& \cellcolor{darkred}71.90 $\pm$ 0.26 $\dagger\dagger$& \cellcolor{darkred}72.23 $\pm$ 0.91 $\dagger\dagger$& \cellcolor{darkred}72.30 $\pm$ 0.87 $\dagger\dagger$& \cellcolor{darkred}72.40 $\pm$ 0.32 $\dagger\dagger$& \cellcolor{darkred}72.61 $\pm$ 0.42 $\dagger\dagger$& \cellcolor{darkred}73.74 $\pm$ 0.33 $\dagger\dagger$& 73.78 $\pm$ 0.5 & - \\

    \bottomrule 
    
    \end{tabular}
    }
    \caption{ Dependency parsing UAS scores on original (Org) and augmented datasets, where the results of \texttt{uuparser} are given at the top, and \texttt{biaffine} parser at the bottom. \emph{language\^} denotes that \emph{language} is part of the multilingual BERT.}
    \label{tab:depparsing_uas}
  \end{table*}

  \begin{table*}[!ht]
    \centering
    \scalebox{0.55}{
    
    \begin{tabular}{clllllllllll}
    \toprule
    & & &\multicolumn{1}{c}{\textbf{Token-Level}} &\multicolumn{5}{c}{\textbf{Character-Level}} &\multicolumn{3}{c}{\textbf{Syntactic}} \\
    \midrule
    & &\textbf{Org} &\textbf{SR} &\textbf{CI} &\textbf{CSU} &\textbf{CSW} &\textbf{CD} &\textbf{CA} &\textbf{Crop} &\textbf{Rotate} &\textbf{Nonce} \\
    
    \toprule

    \multirow{7}{*}{\rotatebox[origin=c]{90}{\texttt{uuparser}}} & \textbf{be} & 61.78 $\pm$ 0.33 & 56.79 $\pm$ 1.18 $\dagger\dagger$& 63.29 $\pm$ 0.39 **& 61.64 $\pm$ 0.84 & 62.53 $\pm$ 1.24 *& 61.00 $\pm$ 1.02 & 61.62 $\pm$ 0.59 & 61.13 $\pm$ 0.46 & 61.84 $\pm$ 2.64 & 62.92 $\pm$ 1.39 \\

    & \textbf{bxr} & 28.14 $\pm$ 3.80 & 29.04 $\pm$ 3.79 *& 29.46 $\pm$ 3.46 & 30.25 $\pm$ 2.79 **& 29.93 $\pm$ 0.90 **& 30.24 $\pm$ 2.78 **& 31.59 $\pm$ 1.97 **& 29.80 $\pm$ 3.08 **& 27.82 $\pm$ 3.92 & 27.07 $\pm$ 1.90 $\dagger\dagger$\\

    & \textbf{kk} & 44.96 $\pm$ 1.51 & 45.09 $\pm$ 1.80 & 45.34 $\pm$ 0.87 & 46.61 $\pm$ 1.82 **& 45.60 $\pm$ 1.28 *& 45.62 $\pm$ 1.51 *& 45.20 $\pm$ 1.49 *& 44.45 $\pm$ 1.02 & 43.01 $\pm$ 1.48 & 44.75 $\pm$ 1.62 \\

    & \textbf{ku} & 32.71 $\pm$ 1.10 & 35.02 $\pm$ 1.06 **& 35.85 $\pm$ 0.22 **& 35.98 $\pm$ 0.47 **& 35.08 $\pm$ 1.08 *& 37.59 $\pm$ 0.50 **& 36.54 $\pm$ 0.24 **& 34.35 $\pm$ 0.35 **& 38.01 $\pm$ 0.54 **& 37.49 $\pm$ 0.11 **\\

    & \textbf{ta} & 64.01 $\pm$ 2.02 & 63.21 $\pm$ 2.13 & 64.04 $\pm$ 2.34 & 63.74 $\pm$ 0.86 & 63.00 $\pm$ 1.61 & 64.62 $\pm$ 1.71 & 63.78 $\pm$ 1.12 & 64.07 $\pm$ 1.12 & 65.11 $\pm$ 0.28 *& 65.69 $\pm$ 0.49 *\\

    & \textbf{te} & 87.80 $\pm$ 0.67 & 88.19 $\pm$ 0.62 & 87.75 $\pm$ 0.99 & 88.33 $\pm$ 0.98 & 88.10 $\pm$ 0.77 & 88.66 $\pm$ 0.73 *& 88.26 $\pm$ 0.64 & 88.06 $\pm$ 0.89 & 88.09 $\pm$ 0.94 & - \\
 
    & \textbf{vi} & 64.89 $\pm$ 0.33 & 59.10 $\pm$ 0.54 $\dagger\dagger$& 62.79 $\pm$ 0.49 $\dagger$& 62.79 $\pm$ 0.22 $\dagger$& 62.17 $\pm$ 0.31 $\dagger\dagger$& 61.87 $\pm$ 0.53 $\dagger\dagger$& 61.91 $\pm$ 0.81 $\dagger\dagger$& 65.70 $\pm$ 0.98 & 64.84 $\pm$ 0.76  & - \\

    \midrule

    \multirow{7}{*}{\rotatebox[origin=c]{90}{\texttt{biaffine}}} & \textbf{be\^} & 82.81 $\pm$ 0.45 & 85.03 $\pm$ 0.92 *& 85.29 $\pm$ 0.39 **& 84.69 $\pm$ 1.12 *& 84.01 $\pm$ 0.65 & 85.07 $\pm$ 0.72 **& 85.74 $\pm$ 0.39 **& 84.25 $\pm$ 0.75 **& 83.72 $\pm$ 1.10 & 83.76 $\pm$ 0.45 \\

    & \textbf{bxr} & 29.67 $\pm$ 0.66 & 31.24 $\pm$ 0.42 **& 30.81 $\pm$ 0.28 & 30.45 $\pm$ 0.33 & 29.89 $\pm$ 0.46 & 30.03 $\pm$ 0.93 & 30.53 $\pm$ 0.68 *& 29.33 $\pm$ 0.42 $\dagger$& 29.85 $\pm$ 0.71 & 29.16 $\pm$ 0.82 \\

    & \textbf{kk\^} & 50.85 $\pm$ 0.51 & 52.50 $\pm$ 0.67 **& 52.80 $\pm$ 0.56 **& 54.30 $\pm$ 0.83 **& 54.21 $\pm$0.38 **& 53.27 $\pm$ 0.67 **& 53.29 $\pm$ 0.35 **& 50.44 $\pm$ 0.35 & 48.92 $\pm$ 0.52 $\dagger$& 54.38 $\pm$ 0.75 **\\

    & \textbf{ku} & 30.48 $\pm$ 0.48 & 30.23 $\pm$ 0.15 &  32.27 $\pm$ 0.59 **& 30.32 $\pm$ 0.25 & 32.11 $\pm$ 0.56 **& 32.55 $\pm$ 0.46 **& 33.46 $\pm$ 0.66 **& 30.33 $\pm$ 0.70 & 28.78 $\pm$ 0.52 $\dagger\dagger$& 33.08 $\pm$ 0.37 **\\

    & \textbf{ta\^} & 77.39 $\pm$ 0.43 & 79.11 $\pm$ 0.47 **& 77.81 $\pm$ 0.26 & 78.65 $\pm$ 1.18 *& 77.86 $\pm$ 0.60 & 78.77 $\pm$ 0.63 **& 78.85 $\pm$ 0.36 **& 78.23 $\pm$ 0.48 **& 78.45 $\pm$ 0.91 *& 77.88 $\pm$ 0.85 \\

    & \textbf{te\^} & 91.57 $\pm$ 0.57 & 91.82 $\pm$ 0.82 & 91.35 $\pm$ 0.78 $\dagger$& 91.96 $\pm$ 0.43 & 92.09 $\pm$ 0.62 *&91.40 $\pm$ 0.40 & 92.23 $\pm$ 0.81 *& 92.51 $\pm$ 0.32 **& 91.75 $\pm$ 0.21 & - \\

    & \textbf{vi\^} & 74.06 $\pm$ 0.21 & 69.24 $\pm$ 0.79 $\dagger\dagger$& 71.90 $\pm$ 0.26 $\dagger\dagger$& 72.23 $\pm$ 0.91 $\dagger\dagger$& 72.30 $\pm$ 0.87 $\dagger\dagger$& 72.40 $\pm$ 0.32 $\dagger\dagger$& 72.61 $\pm$ 0.42 $\dagger\dagger$& 73.74 $\pm$ 0.33 $\dagger\dagger$& 73.78 $\pm$ 0.5 & - \\

    \bottomrule 
    
    \end{tabular}
    }
    \caption{ Dependency parsing UAS scores on original (Org) and augmented datasets, where the results of \texttt{uuparser} are given at the top, and \texttt{biaffine} parser at the bottom. \emph{language\^} denotes that \emph{language} is part of the multilingual BERT. Significance denoted with symbols only.}
    \label{tab:depparsing_uas_no_color}
  \end{table*} 

\appendixsection{Result Tables without Colors}
\thirdrev{}{We visualize the results presented in Sec.~\ref{sec:results} using no colors. POS Tagging, Dependency Parsing and Semantic Role Labeling results are given in Table~\ref{tab:postag_res_no_color}, Table~\ref{tab:depparsing_res_no_color} and Table~\ref{tab:srl_results_no_color} accordingly.}

  \begin{table*}
    \centering
    \scalebox{0.55}{
    
    \begin{tabular}{clllllllllll}
    \toprule
    & & &\multicolumn{1}{c}{\textbf{Token-Level}} &\multicolumn{5}{c}{\textbf{Character-Level}} &\multicolumn{3}{c}{\textbf{Syntactic}} \\
    \midrule
    & &\textbf{Org} &\textbf{SR} &\textbf{CI} &\textbf{CSU} &\textbf{CSW} &\textbf{CD} &\textbf{CA} &\textbf{Crop} &\textbf{Rotate} &\textbf{Nonce} \\
    
    \toprule

    \multirow{7}{*}{\rotatebox[origin=c]{90}{\texttt{uuparser}}} & \textbf{be} & 52.85 $\pm$ 0.70 & 50.10 $\pm$ 0.67 $\dagger\dagger$& 54.05 $\pm$ 0.40 **& 53.39 $\pm$ 0.51 & 55.37 $\pm$ 1.48 *& 54.89 $\pm$ 0.87 *& 53.57 $\pm$ 0.97 *& 53.01 $\pm$ 1.01 & 53.03 $\pm$ 0.31 & 54.27 $\pm$ 0.68 \\

    & \textbf{bxr} & 11.73 $\pm$ 1.65 & 12.70 $\pm$ 1.84 *& 12.39 $\pm$ 1.31 & 13.04 $\pm$ 1.20 **& 13.83 $\pm$ 0.46 **& 13.44 $\pm$ 1.06 **& 14.06 $\pm$ 1.14 **& 11.99 $\pm$ 1.52 **& 11.43 $\pm$ 1.39 & 10.70 $\pm$ 0.77 $\dagger\dagger$\\

    & \textbf{kk} & 23.82 $\pm$ 0.92 & 23.82 $\pm$ 0.98 & 24.36 $\pm$ 0.59 & 24.80 $\pm$ 0.75 **& 24.87 $\pm$ 0.77 *& 24.79 $\pm$ 0.86 *& 24.62 $\pm$ 0.83 *& 24.09 $\pm$ 0.79 & 23.76 $\pm$ 0.32 &  24.78 $\pm$ 1.07 *\\

    & \textbf{ku} & 21.09 $\pm$ 0.86 & 23.22 $\pm$ 0.60 **& 23.70 $\pm$ 0.92 **& 24.08 $\pm$ 0.74 **& 23.48 $\pm$ 0.87 *& 24.32 $\pm$ 0.55 **& 24.39 $\pm$ 0.12 **& 22.74 $\pm$ 0.35 **& 24.50 $\pm$ 0.44 **& 23.98 $\pm$ 0.31 **\\

    & \textbf{ta} & 55.52 $\pm$ 0.39 & 54.27 $\pm$ 1.99 & 55.39 $\pm$ 0.42 & 55.15 $\pm$ 0.71 & 54.89 $\pm$ 0.90 & 56.11 $\pm$ 0.77 & 56.52 $\pm$ 1.31 & 54.62 $\pm$ 0.65 *& 55.16 $\pm$ 0.18 *& 57.50 $\pm$ 0.43 **\\

    & \textbf{te} & 77.79 $\pm$ 0.85 & 76.94 $\pm$ 0.94 & 77.65 $\pm$ 0.94 & 78.83 $\pm$ 0.73 *& 77.27 $\pm$ 1.12 *& 78.11 $\pm$ 0.80 **& 78.87 $\pm$ 0.51 & 78.01 $\pm$ 0.86 *& 78.27 $\pm$ 1.05 *& - \\
 
    & \textbf{vi} & 55.01 $\pm$ 0.15 & 48.23 $\pm$ 0.27 $\dagger\dagger$& 53.80 $\pm$ 0.32 & 52.26 $\pm$ 0.47 $\dagger$& 52.19 $\pm$ 0.36 $\dagger\dagger$& 52.92 $\pm$ 0.23 $\dagger\dagger$& 53.15 $\pm$ 0.42 $\dagger$& 55.33 $\pm$ 0.40 & 54.87 $\pm$ 0.26  & - \\

    \midrule

    \multirow{7}{*}{\rotatebox[origin=c]{90}{\texttt{biaffine}}} & \textbf{be\^} & 78.17 $\pm$ 0.36 & 79.66 $\pm$ 0.28 & 80.47 $\pm$ 0.11 **& 80.04 $\pm$ 0.80 **& 78.79 $\pm$ 0.64 & 80.58 $\pm$ 0.49 **& 80.45 $\pm$ 0.36 **& 79.84 $\pm$ 0.30 **& 79.78 $\pm$ 0.35 **& 79.33 $\pm$ 0.27 **\\

    & \textbf{bxr} & 17.71 $\pm$ 0.52 & 18.73 $\pm$ 0.13 **& 17.91 $\pm$ 0.51 & 17.65 $\pm$ 0.44 & 17.73 $\pm$ 0.17 & 17.72 $\pm$ 0.87 & 17.81 $\pm$ 0.50 *& 16.74 $\pm$ 0.88 $\dagger$& 17.59 $\pm$ 0.36 & 17.12 $\pm$ 0.82 \\

    & \textbf{kk\^} & 34.23 $\pm$ 0.32 & 35.49 $\pm$ 0.75 **& 35.61 $\pm$ 0.52 **& 36.69 $\pm$ 0.76 **& 36.30 $\pm$0.59 **& 35.42 $\pm$ 0.49 **& 35.76 $\pm$ 0.42 **& 34.61 $\pm$ 0.75 *& 34.18 $\pm$ 0.42 & 38.43 $\pm$ 0.17 **\\

    & \textbf{ku} & 20.48 $\pm$ 0.35 & 20.64 $\pm$ 0.24 & 22.16 $\pm$ 0.47 **& 20.49 $\pm$ 0.33 & 22.56 $\pm$ 0.68 **& 22.71 $\pm$ 0.30 **& 23.13 $\pm$ 0.26 **& 20.53 $\pm$ 0.41 & 19.39 $\pm$ 0.50 $\dagger\dagger$& 22.82 $\pm$ 0.24 **\\

    & \textbf{ta\^} & 70.01 $\pm$ 0.62 & 70.99 $\pm$ 0.21 **& 70.27 $\pm$ 0.10 & 71.21 $\pm$ 1.08 *& 70.54 $\pm$ 0.49 & 71.13 $\pm$ 0.67 **& 71.47 $\pm$ 0.53 **& 70.89 $\pm$ 0.30 **& 71.11 $\pm$ 1.01 *& 71.24 $\pm$ 0.85 **\\

    & \textbf{te\^} & 84.60 $\pm$ 0.79 & 84.60 $\pm$ 0.80 & 84.28 $\pm$ 0.97 $\dagger$& 84.88 $\pm$ 0.54 **& 85.02 $\pm$ 0.32 **& 84.05 $\pm$ 0.78 $\dagger\dagger$& 84.74 $\pm$ 0.81 & 85.30 $\pm$ 0.57 **& 84.19 $\pm$ 0.59 $\dagger\dagger$& - \\

    & \textbf{vi\^} & 66.25 $\pm$ 0.84 & 62.40 $\pm$ 0.76 $\dagger\dagger$& 63.49 $\pm$ 0.80 $\dagger\dagger$& 63.87 $\pm$ 0.55 $\dagger\dagger$& 64.22 $\pm$ 2.07 $\dagger\dagger$& 64.54 $\pm$ 2.05 $\dagger\dagger$& 64.09 $\pm$ 1.25 $\dagger\dagger$& 66.11 $\pm$ 0.85 & 65.94 $\pm$ 0.91 & - \\

    \bottomrule 
    
    \end{tabular}
    }
    \caption{ Dependency parsing LAS scores on original (Org) and augmented datasets, where the results of \texttt{uuparser} are given at the top, and \texttt{biaffine} parser at the bottom. \emph{language\^} denotes that \emph{language} is part of the multilingual BERT. Significance denoted with symbols only.}
    \label{tab:depparsing_res_no_color}
  \end{table*}

    \begin{table*}
    \centering
    \scalebox{0.55}{
    
      \begin{tabular}{lllllllllll}
      \toprule
      
      \multicolumn{9}{c}{\textbf{Catalan}} & \\
      \midrule
      \textbf{\#sample} &\textbf{Org} &\textbf{SR} &\textbf{CI} &\textbf{CSU} &\textbf{CSW} &\textbf{CD} &\textbf{CA} &\textbf{Crop} &\textbf{Rotate}  \\
      
      250 & 31.34 $\pm$ 0.99 & 37.29 $\pm$ 1.26 **& 38.50 $\pm$ 1.37 **& 37.74 $\pm$ 1.64 **& 39.79 $\pm$ 0.25 **& 37.45 $\pm$ 1.58 **& 38.28 $\pm$ 1.48 **& 26.59 $\pm$ 0.52 $\dagger\dagger$& 25.96 $\pm$ 0.62 $\dagger\dagger$\\

      500 & 44.53 $\pm$ 1.39 & 44.12 $\pm$ 1.78 & 49.75 $\pm$ 0.10 **& 47.77 $\pm$ 0.65 *& 46.81 $\pm$ 1.59 *& 47.47 $\pm$ 0.62 *& 49.46 $\pm$ 0.19 **& 44.30 $\pm$ 1.16 $\dagger$& 44.72 $\pm$ 0.36 \\
      
      1000 & 53.06 $\pm$ 1.33 & 53.49 $\pm$ 0.64 & 53.80 $\pm$ 0.67 & 54.83 $\pm$ 0.81 & 56.37 $\pm$ 0.29 *& 55.50 $\pm$ 0.17 *& 53.97 $\pm$ 0.96 & 53.57 $\pm$ 0.07 & 52.31 $\pm$ 0.96 \\
  
      \midrule
      \multicolumn{9}{c}{\textbf{Turkish}} & \\
      \midrule
      \textbf{\#sample} &\textbf{Org} &\textbf{SR} &\textbf{CI} &\textbf{CSU} &\textbf{CSW} &\textbf{CD} &\textbf{CA} &\textbf{Crop} &\textbf{Rotate}  \\
      
      250 & 31.28 $\pm$ 0.12 & 31.78 $\pm$ 1.20 & 31.62 $\pm$ 1.53 & 30.67 $\pm$ 2.19 & 30.14 $\pm$ 2.46 & 32.52 $\pm$ 0.89 *& 32.02 $\pm$ 1.53 & 32.42 $\pm$ 0.64 *& 30.37 $\pm$ 1.75 \\

      500 & 35.75 $\pm$ 1.38 & 35.95 $\pm$ 0.65 & 36.35 $\pm$ 0.68 & 38.27 $\pm$ 0.88 *& 37.30 $\pm$ 1.05 & 34.80 $\pm$ 1.96 & 36.23 $\pm$ 1.37 & 37.40 $\pm$ 1.07 *& 34.58 $\pm$ 1.89 \\

      1000 & 44.89 $\pm$ 0.80 & 42.41 $\pm$ 1.20 & 43.36 $\pm$ 0.67 & 41.78 $\pm$ 1.39 & 42.28 $\pm$ 1.31 & 41.42 $\pm$ 1.38 & 29.14 $\pm$ 0.47 $\dagger\dagger$& 44.83 $\pm$ 1.07 & 42.71 $\pm$ 0.76 \\
     
      \midrule
      \multicolumn{9}{c}{\textbf{Spanish}} & \\
      \midrule
      \textbf{\#sample} &\textbf{Org} &\textbf{SR} &\textbf{CI} &\textbf{CSU} &\textbf{CSW} &\textbf{CD} &\textbf{CA} &\textbf{Crop} &\textbf{Rotate}  \\
      250 & 31.23 $\pm$ 1.30 & 34.65 $\pm$ 1.02 *& 36.31 $\pm$ 2.28 & 37.91 $\pm$ 1.21 **& 36.80 $\pm$ 1.90 *& 36.76 $\pm$ 2.11 *& 37.34 $\pm$ 1.14 **& 26.04 $\pm$ 1.34 $\dagger\dagger$& 26.95 $\pm$ 2.22 $\dagger\dagger$\\

      500 & 44.16 $\pm$ 0.68 & 43.89 $\pm$ 0.65 & 44.80 $\pm$ 0.81 & 44.82 $\pm$ 1.10 & 43.75 $\pm$ 1.36 & 44.60 $\pm$ 0.72 *& 45.03 $\pm$ 0.91 *& 42.35 $\pm$ 0.42 $\dagger\dagger$& 41.13 $\pm$ 0.38 $\dagger\dagger$\\

      1000 & 50.98 $\pm$ 1.30 & 50.80 $\pm$ 1.21 & 53.03 $\pm$ 0.76 **& 52.53 $\pm$ 0.60 **& 52.10 $\pm$ 1.00 *& 52.76 $\pm$ 0.57 **& 54.12 $\pm$ 0.12 **& 51.32 $\pm$ 1.09 & 52.40 $\pm$ 0.34 *\\
     
      \midrule
      \multicolumn{9}{c}{\textbf{Czech}} & \\
      \midrule
      \textbf{\#sample} &\textbf{Org} &\textbf{SR} &\textbf{CI} &\textbf{CSU} &\textbf{CSW} &\textbf{CD} &\textbf{CA} &\textbf{Crop} &\textbf{Rotate}  \\
      
      250 & 35.63 $\pm$ 1.08 & 33.48 $\pm$ 1.58 & 37.17 $\pm$ 0.13 *& 37.56 $\pm$ 0.02 *& 31.56 $\pm$ 1.00 $\dagger$& 35.70 $\pm$ 0.80 & 36.14 $\pm$ 1.78 & 34.84 $\pm$ 1.55 **& 30.78 $\pm$ 0.38 $\dagger\dagger$\\

      500 & 44.96 $\pm$ 0.51 & 42.04 $\pm$ 1.08 $\dagger$& 46.44 $\pm$ 1.20 *& 46.13 $\pm$ 1.97 & 46.24 $\pm$ 1.60 & 47.92 $\pm$ 0.59 **& 47.42 $\pm$ 1.07 *& 46.38 $\pm$ 0.12 *& 44.83 $\pm$ 1.47 \\
      
      1000 & 50.29 $\pm$ 0.09 & 48.15 $\pm$ 0.31 $\dagger$& 48.66 $\pm$ 0.49 $\dagger$& 52.63 $\pm$ 0.97 *& 51.10 $\pm$ 1.23 & 47.85 $\pm$ 1.19 $\dagger$& 51.97 $\pm$ 0.54 *& 49.33 $\pm$ 1.36 & 48.45 $\pm$ 1.34 \\
     
      \midrule
      \multicolumn{9}{c}{\textbf{Finnish}} & \\
      \midrule
      \textbf{\#sample} &\textbf{Org} &\textbf{SR} &\textbf{CI} &\textbf{CSU} &\textbf{CSW} &\textbf{CD} &\textbf{CA} &\textbf{Crop} &\textbf{Rotate} \\
      
      250 & 18.80 $\pm$ 0.89 & 18.60 $\pm$ 2.34 & 17.18 $\pm$ 1.56 & 19.68 $\pm$ 1.16 & 25.71 $\pm$ 0.80 **& 28.87 $\pm$ 1.55 **& 27.96 $\pm$ 0.44 **& 30.83 $\pm$ 0.17 **& 20.07 $\pm$ 2.42 \\

      500 & 37.23 $\pm$ 0.35 & 34.10 $\pm$ 1.61 & 35.59 $\pm$ 0.87 $\dagger$& 36.88 $\pm$ 0.29 $\dagger$& 38.98 $\pm$ 0.03 *& 35.29 $\pm$ 0.99 $\dagger$& 35.32 $\pm$ 1.63 & 37.58 $\pm$ 0.93 *& 35.16 $\pm$ 1.08 \\

      1000 & 41.64 $\pm$ 0.98 & 40.15 $\pm$ 0.95 $\dagger$& 41.27 $\pm$ 0.84 & 41.41 $\pm$ 1.08 & 39.80 $\pm$ 0.81 & 40.57 $\pm$ 1.17 & 41.99 $\pm$ 0.04 & 41.35 $\pm$ 1.37 & 39.08 $\pm$ 0.34 $\dagger$\\
     
      \bottomrule
      \end{tabular}
    }
  \caption{ Semantic role labeling results on original (Org) and augmented datasets. Significance denoted with symbols only.}
  \label{tab:srl_results_no_color}
  \end{table*}

  \begin{landscape}
    \topskip0pt
    \vspace*{\fill}
    \begin{table}[htb!]
      \centering
      \setlength{\tabcolsep}{2pt}
      \scalebox{0.68}{
      
      \begin{tabular}{clllllllllllll}
        \toprule
        & & &\multicolumn{3}{c}{\textbf{Token-Level}} &\multicolumn{5}{c}{\textbf{Character-Level}} &\multicolumn{3}{c}{\textbf{Syntactic}} \\
        \midrule
        & &\textbf{Org} &\textbf{RWD} &\textbf{RWS} &\textbf{SR} &\textbf{CI} &\textbf{CSU} &\textbf{CSW} &\textbf{CD} &\textbf{CA} &\textbf{Crop} &\textbf{Rotate} &\textbf{Nonce} \\
        \toprule

        \multirow{7}{*}{\rotatebox[origin=c]{90}{\texttt{char}}} & \textbf{be} & 90.29 $\pm$ 1.23 & 90.41 $\pm$ 1.68 & 91.38 $\pm$ 1.36 *& 88.57 $\pm$ 2.48 $\dagger$& 91.00 $\pm$ 0.43 **& 91.66 $\pm$ 0.39 **& 90.24 $\pm$ 1.17 & 91.60 $\pm$ 0.79 **& 91.50 $\pm$ 0.63 **& 88.75 $\pm$ 2.30 $\dagger$& 90.06 $\pm$ 1.12 & 89.68 $\pm$ 1.26 \\
        
        & \textbf{bxr} & 39.29 $\pm$ 4.16 & 43.77 $\pm$ 6.37 & 41.48 $\pm$ 6.13 & 46.44 $\pm$ 3.16 **& 41.36 $\pm$ 5.60 *& 41.75 $\pm$ 5.60 & 41.15 $\pm$ 6.19 & 41.06 $\pm$ 4.83 *& 39.99 $\pm$ 5.33 & 43.96 $\pm$ 1.53 *& 42.37 $\pm$ 5.20 & 42.45 $\pm$ 3.57 *\\

        & \textbf{kk} & 46.24 $\pm$ 3.02 & 49.65 $\pm$ 2.75 **& 46.36 $\pm$ 3.47 & 51.25 $\pm$ 3.04 **& 49.94 $\pm$ 2.91 **& 50.61 $\pm$ 3.15 **& 49.51 $\pm$ 3.47 **& 48.38 $\pm$ 4.0 *& 51.49 $\pm$ 1.22 **& 42.57 $\pm$ 9.04 & 46.94 $\pm$ 2.84 & 51.45 $\pm$ 0.58 **\\

        & \textbf{ku} & 57.04 $\pm$ 3.98 & 57.70 $\pm$ 5.20 & 58.30 $\pm$ 3.25 *& 56.28 $\pm$ 3.71 $\dagger$& 58.90 $\pm$ 3.41 *& 57.97 $\pm$ 3.11 *& 56.99 $\pm$ 3.20 & 56.62 $\pm$ 3.40 & 58.60 $\pm$ 2.10 *& 57.67 $\pm$ 3.71 *& 54.46 $\pm$ 6.77 & 57.09 $\pm$ 2.68 \\
        
        & \textbf{ta} &  84.30 $\pm$ 1.87 & 86.96 $\pm$ 0.54 *& 86.37 $\pm$ 0.26 **& 85.94 $\pm$ 0.35 & 86.29 $\pm$ 0.40 **& 86.62 $\pm$ 0.14 **& 86.37 $\pm$ 0.57 **& 86.24 $\pm$ 0.62 *& 86.33 $\pm$ 0.77 *& 84.63 $\pm$ 0.79 & 84.36 $\pm$ 0.57 & 85.98 $\pm$ 0.80 *\\

        & \textbf{te} & 90.96 $\pm$ 1.38 & 90.07 $\pm$ 1.85 & 90.51 $\pm$ 1.21 & 90.80 $\pm$ 0.36 $\dagger$& 89.90 $\pm$ 1.54 & 90.21 $\pm$ 1.23 & 90.49 $\pm$ 1.68 & 90.62 $\pm$ 1.40 & 90.23 $\pm$ 1.73 & 89.99 $\pm$ 0.49 $\dagger\dagger$& 90.01 $\pm$ 0.63 $\dagger$& - \\
        
        & \textbf{vi} & 77.62 $\pm$ 0.54 & 77.45 $\pm$ 0.75 & 77.10 $\pm$ 0.90 & 77.93 $\pm$ 0.21 & 76.97 $\pm$ 0.87 & 77.88 $\pm$ 0.24 & 77.29 $\pm$ 0.25 & 76.00 $\pm$ 0.57 $\dagger$& 78.08 $\pm$ 0.76 & 77.17 $\pm$ 0.26 $\dagger$& 77.20 $\pm$ 0.45 $\dagger$& - \\
        
        \midrule

        \multirow{6}{*}{\rotatebox[origin=c]{90}{\texttt{BPE}}} & \textbf{be} & 88.43 $\pm$ 0.59 & 89.75 $\pm$ 0.42 **& 89.72 $\pm$ 0.38 **& 87.61 $\pm$ 0.80 $\dagger$& 86.06 $\pm$ 1.64 **& 88.73 $\pm$ 1.68 & 87.30 $\pm$ 1.59 $\dagger$& 88.41 $\pm$ 1.47 & 89.23 $\pm$ 1.53 *& 87.83 $\pm$ 0.74 & 87.81 $\pm$ 1.81 & 88.40 $\pm$ 1.66 \\

        & \textbf{bxr} & 33.11 $\pm$ 0.80 & 44.21 $\pm$ 7.06 **& 41.50 $\pm$ 7.81 **& 46.67 $\pm$ 7.55 **& 41.82 $\pm$ 7.37 **& 43.49 $\pm$ 7.72 **& 39.14 $\pm$ 5.37 **& 46.06 $\pm$ 7.43 **& 42.20 $\pm$ 6.91 **& 28.60 $\pm$ 9.77 & 40.47 $\pm$ 7.43 **& 44.21 $\pm$ 5.51 **\\

        & \textbf{kk} & 46.20 $\pm$ 0.68 & 50.36 $\pm$ 1.75 **& 46.83 $\pm$ 0.30 *& 50.29 $\pm$ 1.66 **& 48.64 $\pm$ 2.84 *& 47.14 $\pm$ 0.52 **& 48.33 $\pm$ 1.84 **& 47.28 $\pm$ 2.57 & 47.90 $\pm$ 2.40 *& 47.02 $\pm$ 6.51 & 49.00 $\pm$ 2.16 **& 50.75 $\pm$ 2.31 \\

        & \textbf{ta} &  82.76 $\pm$ 0.41 & 81.96 $\pm$ 1.00 & 82.48 $\pm$ 0.59 & 81.51 $\pm$ 1.22 & 82.77 $\pm$ 0.31 & 82.79 $\pm$ 0.35 & 82.22 $\pm$ 0.52 & 82.48 $\pm$ 0.49 & 82.26 $\pm$ 0.68 & 81.69 $\pm$ 0.56 & 80.97 $\pm$ 0.64 $\dagger$& 81.97 $\pm$ 1.00 \\

        & \textbf{te} & 88.27 $\pm$ 0.60 & 88.60 $\pm$ 0.23 *& 88.74 $\pm$ 0.63 *& 87.71 $\pm$ 0.63 $\dagger\dagger$& 88.93 $\pm$ 0.26 **& 88.21 $\pm$ 0.31 & 87.93 $\pm$ 0.67 $\dagger$& 88.27 $\pm$ 0.67 & 88.35 $\pm$ 0.34 & 87.38 $\pm$ 0.93 & 87.68 $\pm$ 0.36 & - \\

        & \textbf{vi} & 77.07 $\pm$ 0.46 & 77.83 $\pm$ 0.50 **& 77.77 $\pm$ 0.33 **& 77.83 $\pm$ 0.62 **& 77.42 $\pm$ 0.33 & 77.39 $\pm$ 0.54 & 77.46 $\pm$ 0.56 *& 77.53 $\pm$ 0.37 *& 77.72 $\pm$ 0.43 **& 75.17 $\pm$ 1.08 $\dagger\dagger$& 76.04 $\pm$ 0.41 $\dagger\dagger$& - \\

        \midrule
        \multirow{4}{*}{\rotatebox[origin=c]{90}{\texttt{mBERT}}} & \textbf{be} & 95.30 $\pm$ 0.65 & 94.50 $\pm$ 0.63 $\dagger$& 94.27 $\pm$ 0.68 $\dagger$& 94.55 $\pm$ 0.92 & 94.44 $\pm$ 1.39 & 95.70 $\pm$ 0.60 **& 95.18 $\pm$ 0.74 & 94.76 $\pm$ 1.16 & 95.56 $\pm$ 0.34 *& 95.64 $\pm$ 0.71 *& 95.11 $\pm$ 0.70 & 94.48 $\pm$ 1.02 \\

        & \textbf{kk} & 71.34 $\pm$ 2.28 & 70.75 $\pm$ 1.89 & 73.10 $\pm$ 1.65 **& 71.88 $\pm$ 2.16 *& 71.65 $\pm$ 1.51 *& 69.00 $\pm$ 4.01 & 73.11 $\pm$ 0.77 **& 70.90 $\pm$ 1.58 & 72.57 $\pm$ 1.92 *& 67.42 $\pm$ 2.20 & 72.00 $\pm$ 1.59 *& 72.45 $\pm$ 1.72 **\\

        & \textbf{te} & 89.92 $\pm$ 0.48 & 89.53 $\pm$ 0.95 & 90.12 $\pm$ 1.03 *& 88.43 $\pm$ 0.81 $\dagger$& 89.49 $\pm$ 1.17 & 88.84 $\pm$ 1.41 & 88.59 $\pm$ 0.64 & 89.40 $\pm$ 0.41 $\dagger$& 90.15 $\pm$ 1.31 *& 88.82 $\pm$ 1.01 $\dagger$& 89.01 $\pm$ 0.48 $\dagger$& - \\

        & \textbf{vi} & 82.41 $\pm$ 0.72 & 81.66 $\pm$ 0.97 & 82.75 $\pm$ 0.87 & 81.04 $\pm$ 1.37 & 82.45 $\pm$ 0.89 & 82.09 $\pm$ 1.07 & 82.48 $\pm$ 0.81 *& 82.20 $\pm$ 0.87 & 82.57 $\pm$ 0.65 *& 82.28 $\pm$ 0.80 & 82.33 $\pm$ 0.79 & - \\
        \bottomrule         
        \end{tabular}
        }
      \caption{ Part-of-speech tagging results on original (Org) and augmented datasets, where the results of \texttt{char} are given at the top, \texttt{BPE} in the middle and \texttt{mBERT} at the bottom. Significance denoted with symbols only.}
      \label{tab:postag_res_no_color}
  \end{table}
  \vspace*{\fill}
  \end{landscape}

\end{document}